\pdfoutput=1

\documentclass[11pt]{article}

\usepackage{EMNLP2023}

\usepackage{times}
\usepackage{latexsym}

\usepackage[T1]{fontenc}

\usepackage[utf8]{inputenc}

\usepackage{microtype}

\usepackage{inconsolata}

\usepackage{hyperref}       
\usepackage{url}            
\usepackage{booktabs}       
\usepackage{amsfonts}       
\usepackage{nicefrac}       
\usepackage{graphicx}
\usepackage{algorithm}
\usepackage{tabularx}
\usepackage{makecell}
\usepackage{array}
\usepackage{enumitem}
\usepackage[noend]{algpseudocode}
\usepackage{listings}
\usepackage{xcolor}

\usepackage{pgfplots}
\usepackage{pgfplotstable}
\pgfplotsset{compat=1.17}

\lstdefinestyle{python}{
    language=Python,
    basicstyle=\ttfamily\footnotesize,
    commentstyle=\color{gray}\itshape,
    keywordstyle=\color{blue}\bfseries,
    stringstyle=\color{red},
    identifierstyle=\color{black},
    numberstyle=\tiny\color{gray},
    backgroundcolor=\color{white},
    showspaces=false,
    showstringspaces=false,
    breaklines=true,
    breakindent=0pt,
    frame=single,
    tabsize=4,
    captionpos=b,
}
  
\title{TaskGen: A Task-Based, Memory-Infused Agentic Framework\\using StrictJSON}

\author{
  John Chong Min Tan \\
  Simbian AI\\
  National University of Singapore \\
  \texttt{john.chong@simbian.ai}
  \And
  Prince Saroj, Bharat Runwal, Hardik Maheshwari,\\
  \textbf{Alankrit Chona, Ambuj Kumar} \\
  Simbian AI \\
  \texttt{prince@simbian.ai}
  \AND
  Brian Lim Yi Sheng\\
  Singapore-ETH Centre \\
  \texttt{brian.lim@sec.ethz.ch}
  \And
  Richard Cottrill\\
  \texttt{richard\_c@tpg.com.au}
  \And
  Mehul Motani \\
  National University of Singapore\\
  \texttt{motani@nus.edu.sg}
}

\begin{document}
\maketitle

\begin{abstract}
TaskGen is an open-sourced agentic framework which uses an Agent to solve an arbitrary task by breaking them down into subtasks. Each subtask is mapped to an Equipped Function or another Agent to execute. In order to reduce verbosity (and hence token usage), TaskGen uses StrictJSON that ensures JSON output from the Large Language Model (LLM), along with additional features such as type checking and iterative error correction. Key to the philosophy of TaskGen is the management of information/memory on a need-to-know basis. We empirically evaluate TaskGen on various environments such as 40x40 dynamic maze navigation with changing obstacle locations (\textbf{100\%} solve rate), TextWorld escape room solving with dense rewards and detailed goals (\textbf{96\%} solve rate), web browsing (\textbf{69\%} of actions successful), solving the MATH dataset (\textbf{71\%} solve rate over 100 Level-5 problems), Retrieval Augmented Generation on NaturalQuestions dataset (F1 score of \textbf{47.03\%}).
\end{abstract}

\section{Introduction}
\begin{figure}[t]
\begin{center}
\includegraphics[width = 0.5\textwidth]{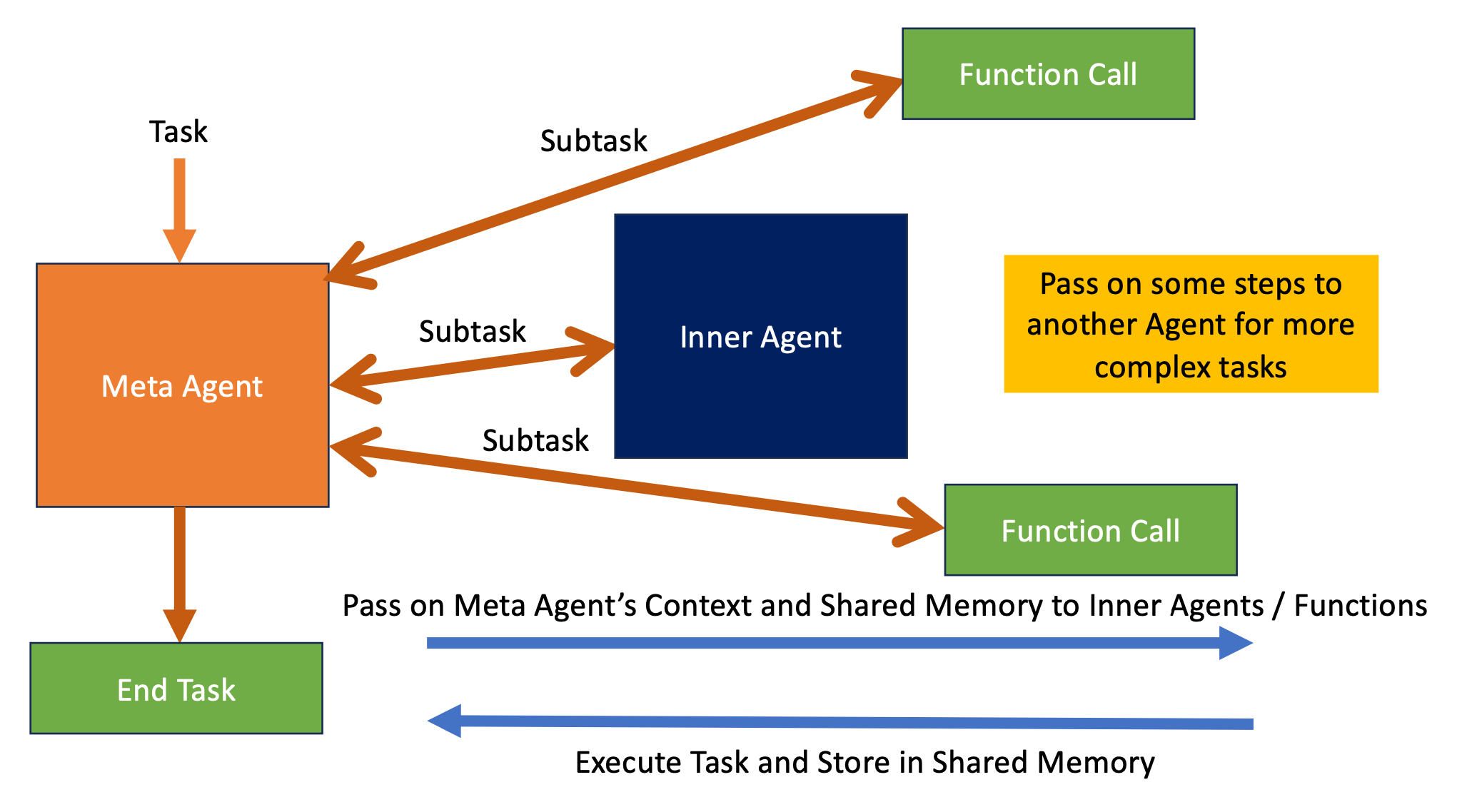}
\end{center}
\caption{An Overview of TaskGen}
\label{fig: TaskGen overview}
\end{figure}

TaskGen (\href{https://github.com/simbianai/taskgen}{https://github.com/simbianai/taskgen}) is an open-sourced agentic framework which breaks down a task into subtasks, each of which are mapped to an Equipped Function or another Agent to execute. The Agents and Equipped Functions operate independently, but share context on a need-to-know basis using Shared Memory (see Fig. \ref{fig: TaskGen overview}).

TaskGen is designed to be less verbose, and hence incurs lower processing latency and costs with potentially improved accuracy, than most existing agentic frameworks which output free text such as AutoGPT \cite{yang2023auto}, BabyAGI \cite{nakajima2023babyagi}, MetaGPT \cite{hong2023metagpt}, AutoGen \cite{wu2023autogen}, ChatDev \cite{qian2023communicative}, CrewAI \cite{joao2023crewai}, LangChain/LangGraph \cite{langgraph}.

\textbf{Our Contributions.} We propose a new open-sourced agentic framework named TaskGen:
\begin{enumerate}[leftmargin=*,nosep,wide=0pt]
\item TaskGen breaks a complex task down into bite-sized subtasks, each of which are mapped to an Equipped Function or Inner Agent to execute. 

\item In contrast to free-form text output in agentic frameworks, TaskGen uses a concise JSON output for each part of the process. Specifically, it uses StrictJSON \cite{tan2023strictjson}, which is an LLM output parser for JSON format with type checking, and helps ensure concise and extractable output which can be used for downstream tasks easily.

\item TaskGen has Shared Memory amongst various components on a need-to-know basis. This Shared Memory can come in the form of 1) \textbf{Subtasks Completed}, a list of past Equipped Functions inputs and outputs, or 2) \textbf{Shared Variables}, which stores important information that may also be of the form of long text or non-text modalities.

\item TaskGen utilises \textbf{Global Context} to inform the Agent of important information that may be dynamically changing. This allows the Agent to react to dynamic environments as the task progresses, or as the Agent switches tasks.

\item Lastly, as memory is key to learning and decision making, TaskGen implements memory of various abstraction spaces in the Agent's \textbf{Memory Bank}, which can be used to augment the prompt to the Agent via Retrieval Augmented Generation (RAG) \cite{lewis2020retrieval} based on semantic similarity to the task. These memories are learnable via experience and can be used to influence future behaviour.
\end{enumerate}

\section{Motivation}

\begin{figure}[H]
\centering
\includegraphics[width=\linewidth]{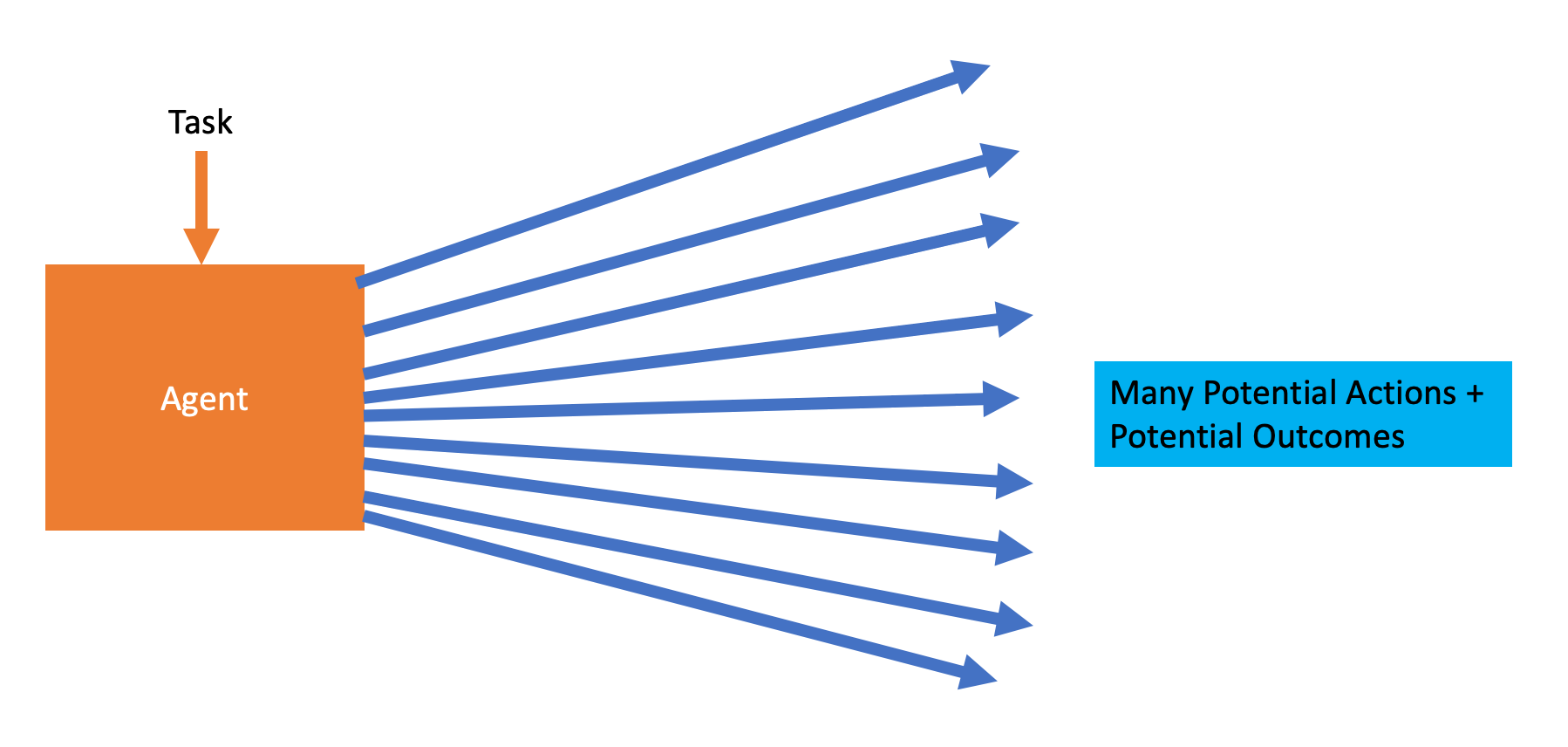}
\caption{Intractable action space when solving an arbitrary task}
\label{fig:taskgen_part1}
\end{figure}

We strive to create an Agent that can solve arbitrary tasks in arbitrary environments. However, when solving an arbitrary task, we could potentially do many actions, and there are many potential outcomes possible, as shown in Fig. \ref{fig:taskgen_part1}. This is intractable for any Agent to manage and we need to limit the scope of what the Agent can do for more robust Agents.

\begin{figure}[H]
\centering
\includegraphics[width=\linewidth]{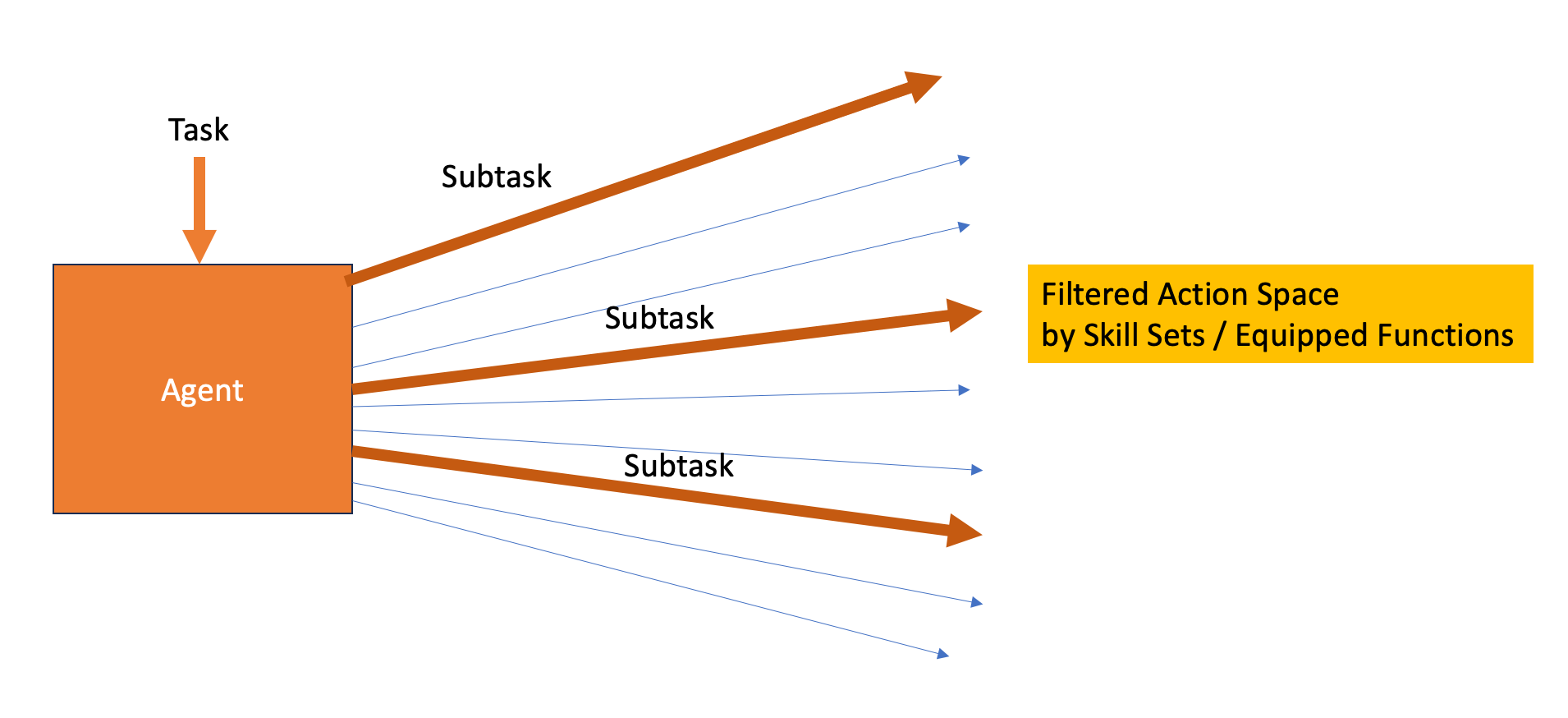}
\caption{Constraining action space by Equipped Functions}
\label{fig:taskgen_part2}
\end{figure}

Hence, we should limit the scope of the Agent by giving it only relevant Equipped Functions. This will help filter the vast action space into something tractable. Moreover, based on the Equipped Functions provided, we can break down a potentially complicated task into bite-sized subtasks, each of which can be solved entirely by one Equipped Function. This is shown in Fig. \ref{fig:taskgen_part2}.

\begin{figure}[H]
\centering
\includegraphics[width=\linewidth]{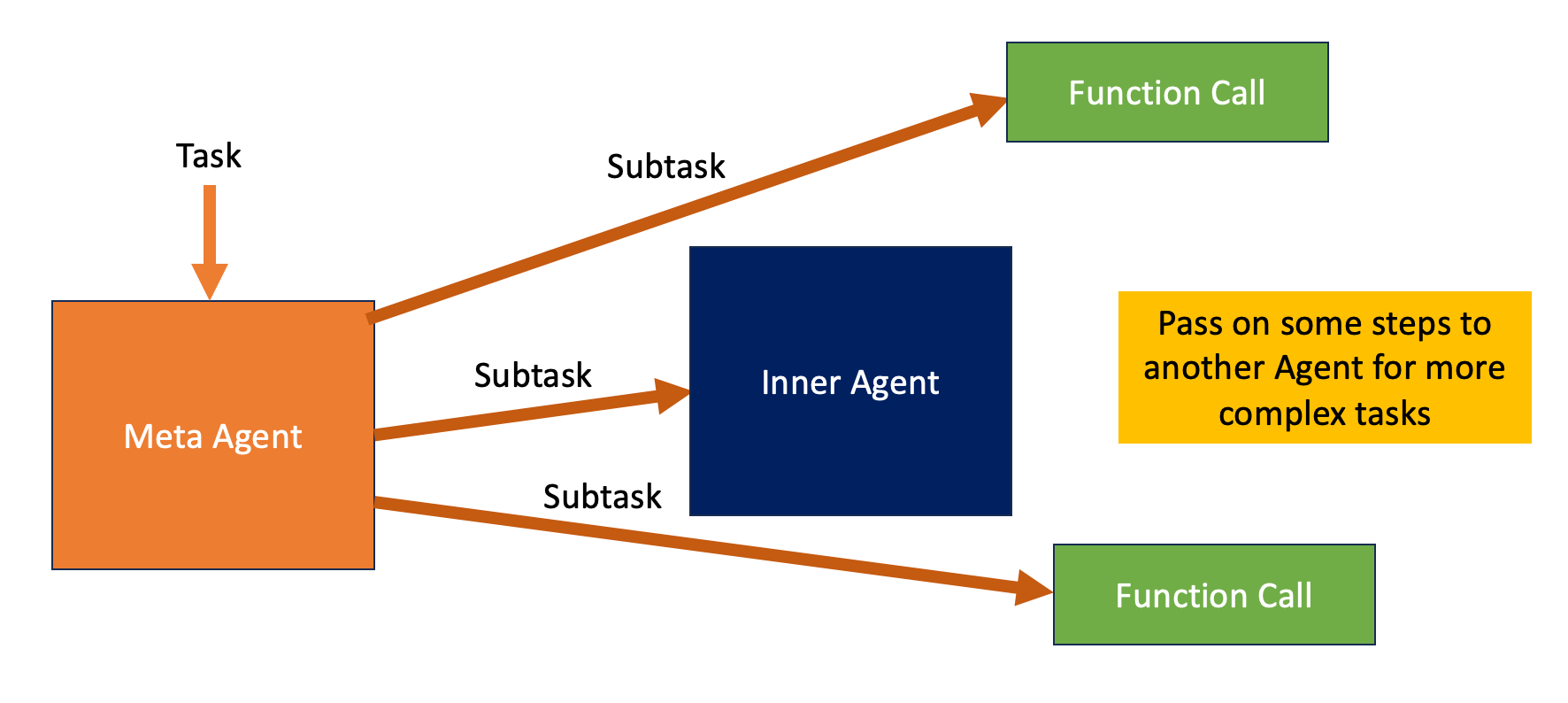}
\caption{Inner Agents assigned as Equipped Functions to a Meta Agent helps increase processing capability}
\label{fig:taskgen_part3}
\end{figure}

In fact, for more complex tasks, we can even let another Agent be the Equipped Function. This Agent will henceforth be referred to as Inner Agent. This is similar to how a manager offloads tasks to each worker, each of whom have their own experiences and skills to do the task. By having intelligent Inner Agents as the Equipped Function, the top-level agent (Meta Agent) will have greater processing capability. This is shown in Fig. \ref{fig:taskgen_part3}.

\textbf{Infusing Shared Awareness.} Each Equipped Function or Inner Agent would now be able to perform a subset of the entire task independently. However, they will need some shared context, as 1) the outcome of the subtask may influence other subtasks down the line, or 2) they may need input from earlier subtasks in order to perform their subtask. To solve this problem, we implement a Shared Memory amongst the Meta Agent, Equipped Function and Inner Agents. Notably, we have two types of Shared Memory, 1) \textbf{Subtasks Completed} and 2) \textbf{Shared Variables}. This is shown in Fig. \ref{fig: TaskGen overview}.

\section{TaskGen Overall Design Philosophy}
TaskGen has three key design philosophies.

Firstly, the output of each Agent or Equipped Function is made to be as concise as possible for minimal token use. This is done using StrictJSON. By ensuring a structured JSON output format with type checking, StrictJSON reduces verbosity typically associated with free-form text output in LLMs. This cuts down on latency and costs, and improves reliability of extracting output fields needed for downstream components. For a more in-depth run-through of StrictJSON, refer to Appendix \ref{appendix: strictjson}.

Secondly, we map each subtask to exactly one Equipped Function or Inner Agent, so as to guarantee executability of the subtask. Unlike AutoGPT \cite{yang2023auto}, we
ensure that there are no infinite loops when executing subtasks. This is done via the following design guidelines:
\begin{enumerate}[leftmargin=*,nosep,wide=0pt]
    \item An Agent can only call an Equipped Function or Inner Agent that is not above it in the hierarchy.
    \item Each Agent gets context relevant to its own processing abstraction space and are assigned Equipped Functions and Inner Agents suitable for that space.
\end{enumerate}

Lastly, information is only shared between Agents and Equipped Functions on a need-to-know basis. We have a shared pool of information in Shared Memory, but we only expose those that are relevant to each Agent / Equipped Function. This helps to reduce context length and minimise the cognitive load on each part of the system.

\section{The Core of TaskGen}

\subsection{Agent Definition}
At the core of TaskGen is the definition of an Agent, which consists of the following components:
\begin{enumerate}[leftmargin=*,nosep,wide=0pt]
\item \textbf{Agent Name}: Name of the Agent
\item \textbf{Agent Description}: Description of the Agent
\item \textbf{Equipped Functions}: List of Equipped Functions and Inner Agents available to solve subtasks 
\item \textbf{Assigned Task}: Agent's assigned task
\item \textbf{Subtasks Completed}: Python dictionary of past subtasks that Agent has done, which detail the Equipped Function's name and input parameters and their corresponding output
\item \textbf{Shared Variables}: Python dictionary containing variables that will be shared between Equipped Functions and Agents
\item \textbf{Global Context}: Additional context to the Agent that can reference persistent states, such as those in \textbf{Shared Variables}
\item \textbf{Memory Bank}: Python dictionary containing various abstraction spaces of memory that will be retrieved via \textit{top-k} retrieval via similarity to Assigned Task
\end{enumerate}

\subsection{Imbuing Agentic Capabilities with Equipped Functions}
By default, an Agent comes pre-built with a \textbf{use\_llm} function, which uses an LLM with the Agent Name and Agent Description as context to perform a task, and an \textbf{end\_task} function to end the current task. Additionally, we can assign Equipped Functions or Inner Agents to the Agent to imbue it additional capabilities.

Equipped Functions come in two forms:

\begin{enumerate}[leftmargin=*,nosep,wide=0pt]
    \item \textbf{Internal Functions} use an LLM to do processing of input-output relations. They are useful for tasks that are difficult for traditional rule-based approaches to handle well, such as sentiment analysis and summarisation. 
    \item \textbf{External Functions} utilise any Python function to do processing to get output, which makes it very easy for TaskGen to utilise functions from other agentic frameworks such as LangChain or CrewAI. They are suitable for tasks that can be called via fixed functions, or APIs, which guarantee reliability while imbuing additional functions to the LLM. As an aside, if we need a hybrid approach of rule-based fixed processes with flexibility of LLMs, an LLM can also be called within the External Function.
\end{enumerate}

\subsection{Choosing the Next Subtask}
The core ability of an Agent is the ability to choose the correct next subtask to fulfil the Assigned Task. This is a non-trivial problem as it requires understanding of the Assigned Task, Agent Name, Agent Description, Subtasks Completed, relevant Memory, Equipped Functions and Inner Agents in order to make an informed decision.

In order to increase robustness in choosing the right Equipped Function and corresponding input parameters, we split it up into two steps. 

\textbf{Step 1: Decide on subtask and corresponding Equipped Function / Inner Agent.}
The first step simply takes the available information to the Agent and does a Chain-of-Thought (CoT) \cite{wei2022chain} prompting to elicit reasoning via thoughts, leading to more accurate selection of subtask and the corresponding Equipped Function / Inner Agent in the following format:
\begin{enumerate}[leftmargin=*,nosep,wide=0pt]
    \item \textbf{Observation}: Reflect on what has been done in Subtasks Completed for Assigned Task
    \item \textbf{Thoughts}: Brainstorm how to complete remainder of Assigned Task only given Observation 
    \item \textbf{Current Subtask}: What to do now in detail with all context provided that can be done by one Equipped Function for Assigned Task
    \item \textbf{Equipped Function Name}: Name of Equipped Function to use for Current Subtask
\end{enumerate}

\textbf{Step 2: Decide on input parameters to Equipped Function / Inner Agent.}
Instead of providing the entire list of Equipped Functions / Inner Agents as per Step 1, we only give this step information of the exact Equipped Function / Inner Agent we have decided in Step 1, so as to encourage greater output specificity. We then generate the input parameters of the Equipped Function / Inner Agent given the Current Subtask and Equipped Function details (Equipped Function Name, Equipped Function Description, Equipped Function Input Parameter Description and type), and uses StrictJSON to ensure that the input parameters meet the type that is stated for in the Equipped Function. This ensures robustness and reliability for the input parameters.

\section{Using TaskGen}
Using TaskGen is extremely simple and is designed for any new user to learn it within 5 minutes. The steps needed are detailed as follows:
\begin{enumerate}[leftmargin=*,nosep,wide=0pt]
    \item \textbf{Install TaskGen.} "\texttt{pip install taskgen-ai}"
    \item \textbf{Define LLM.} This takes in a user prompt and system prompt as Python strings, and returns a Python string for the LLM generated response
    "\texttt{def llm(user\_prompt: str, system\_prompt: str) -> str}"
    \item \textbf{Define Agent.} Simply define an Agent class with the Agent Name, Agent Description\\
    "\texttt{agent = Agent(name, description, llm = llm)}"
    \item \textbf{Equip Functions.} Equip the Agent with Equipped Functions or Inner Agents to broaden the Agent's capabilities.
    "\texttt{agent.assign\_functions([fn\_1, fn\_2])}"
    \item \textbf{Run Agent.} Run the Agent with a task
    "\texttt{agent.run(task)}"
    \item \textbf{Query Agent.} Query the Agent about Subtasks Completed \\"\texttt{agent.reply\_user(query)}"
\end{enumerate}

For an in-depth tutorial on how to use TaskGen, refer to Appendix \ref{appendix: taskgen}.

\section{Benefits of TaskGen}

The key philosophy of TaskGen is to be concise. This helps greatly with the performance of the overall system, as numerous studies \cite{xiong2023effective, ding2024longrope} have shown that an increase in context length generally leads to poorer performance on tasks referencing the context.

\subsection{JSON is more concise than free text}

\begin{figure}[H]
\centering
\includegraphics[width=\linewidth]{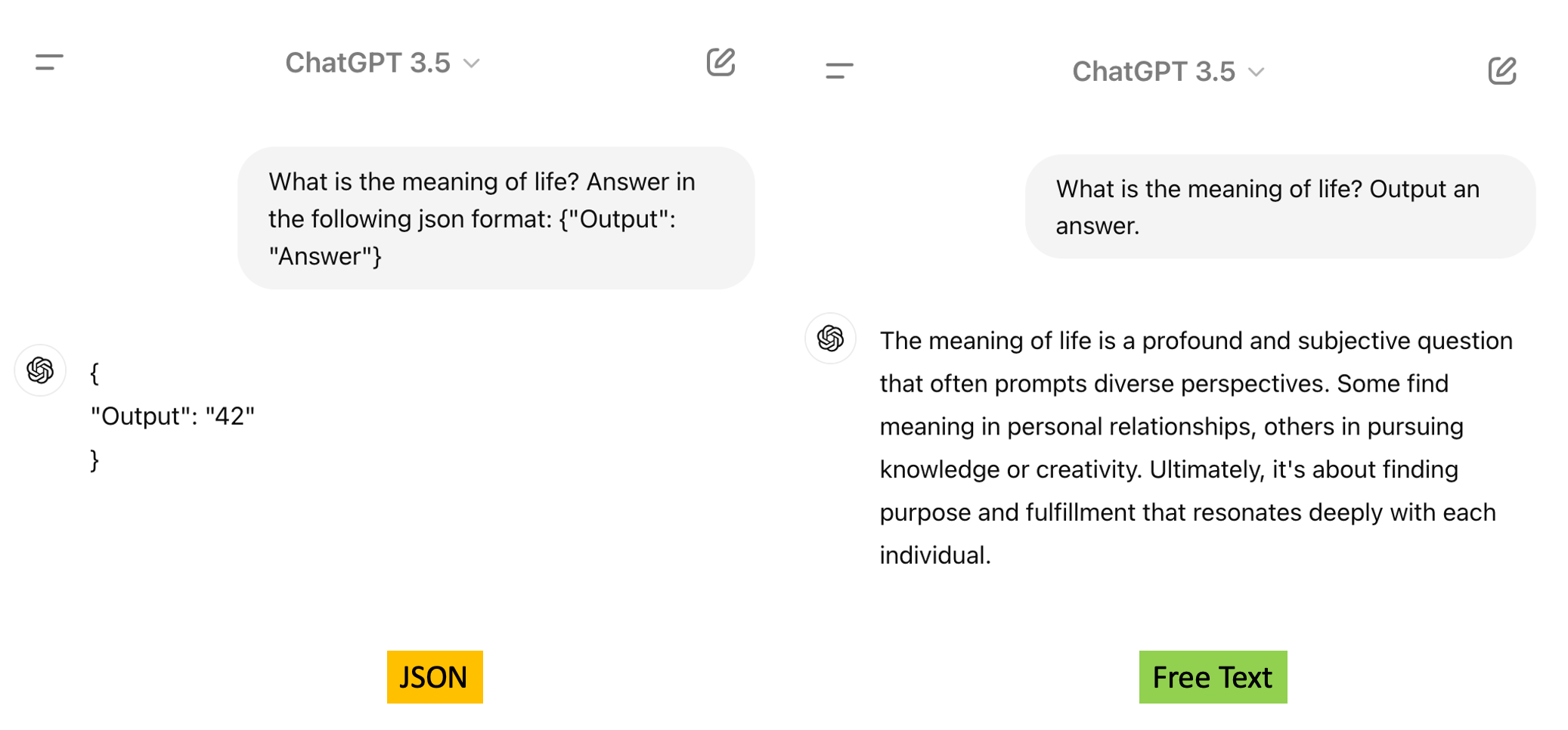}
\caption{More concise output using JSON as compared to Free Text using \texttt{gpt-3.5-turbo} on 12 Jul 2024}
\label{fig:json_vs_text}
\end{figure}

Given a similar input prompt, asking the LLM to output in a JSON format generally gives much less verbose output as compared to free text. An example can be seen from from Fig. \ref{fig:json_vs_text} for a prompt about the meaning of life. This is likely because the pre-training data of JSON on the web is more concise without much explanation, and the value of the field is very correlated to the key of the field. This means that we can use a JSON format to constrain the generation of the LLM to give the desired fields which we are interested in.

\subsection{StrictJSON is more concise than JSON}
\begin{figure}[H]
\centering
\includegraphics[width=\linewidth]{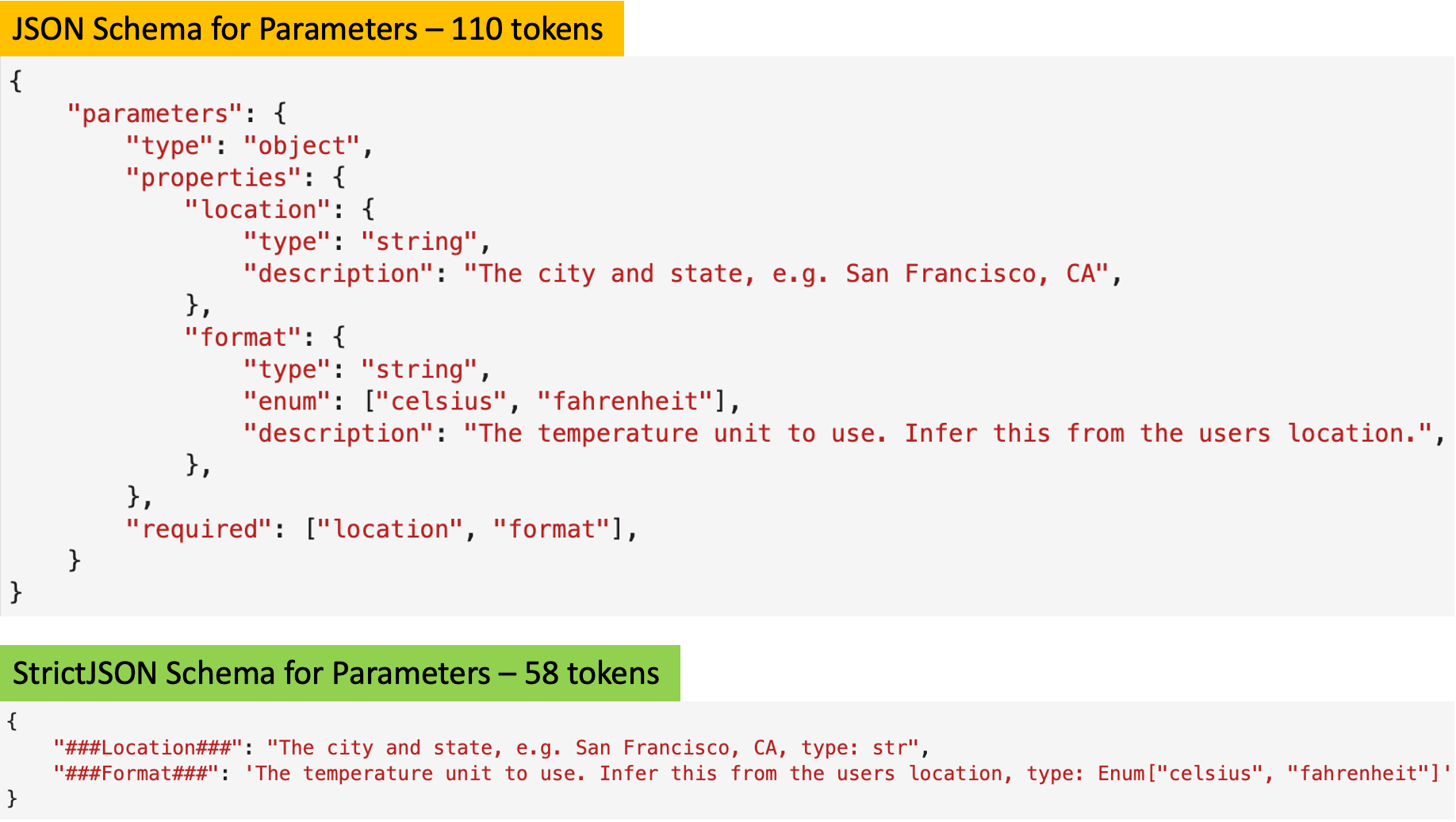}
\caption{StrictJSON Schema (bottom) is much less verbose than JSON Schema (top). Token count is computed using \texttt{gpt-3.5-turbo} tokeniser.}
\label{fig:json_schema}
\end{figure}

TaskGen steers clear away from the typical JSON schema approach to define functions, which are used in many agentic frameworks adopting Pydantic as the JSON parser. This is because the JSON schema format is extremely verbose, and TaskGen using the StrictJSON schema is able to express the entire JSON schema of a function with much fewer tokens. As can be seen in Fig. \ref{fig:json_schema}, in order to express two parameters, the StrictJSON Schema uses 58 tokens compared to JSON Schema of 110 tokens, or about 53\% the amount of tokens. The token savings are significant, and would be even more so with a lot more parameters.

\subsection{Modular and robust components}
TaskGen utilises a modular approach, where for each part of the system, be it Equipped Function or Inner Agent, we give it only the required context to do the task. This results in shorter context for LLM prompts, leading to better performance.

Moreover, as we move from one subtask to the next, we split the process into multiple smaller chunks as required. For instance, when deciding what to do for the next subtask, we choose the Equipped Function / Inner Agent as one chunk, and choose the input parameters as another chunk. This again helps with reducing context length and cognitive load on each part of the process, and we can error check better at each part of the process.

\subsection{Shared Memory}
\begin{figure}[h]
  \centering
  \includegraphics[width=0.5\textwidth]{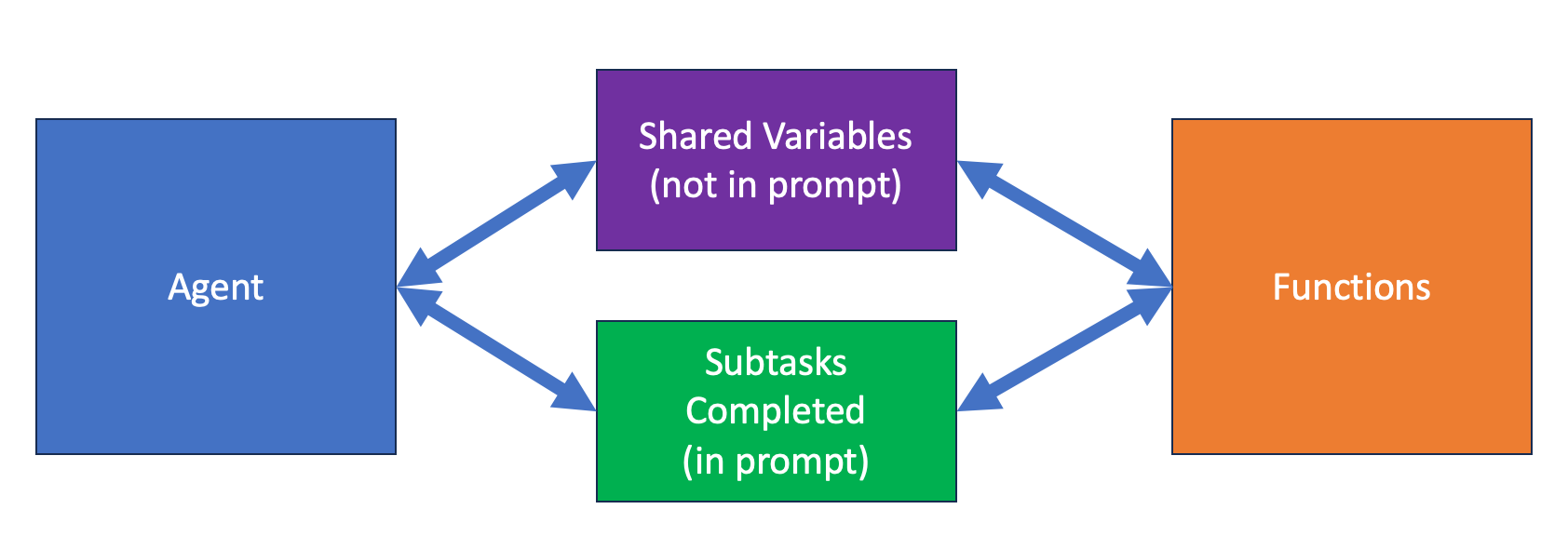}
  \caption{Two types of Shared Memory: Subtasks Completed and Shared Variables}
  \label{fig:types_of_shared_memory}
\end{figure}

One of the key design philosophy of TaskGen is to share information only on a need-to-know basis. To that end, we utilise Shared Memory (see Fig. \ref{fig:types_of_shared_memory}) to share information between the Agent and Equipped Function / Inner Agents.

There are two kinds of Shared Memory:
\begin{enumerate}[leftmargin=*,nosep,wide=0pt]
    \item \textbf{Subtasks Completed.} This is a Python dictionary which stores the outcome of each subtask. The dictionary key is the name of the Equipped Function / Inner Agent and its input parameters, the value is the function output. This past history of function inputs and outputs will be made known to all LLM-based components of the system to help with shared awareness. Do note that this differs from the traditional ReAct framework \cite{yao2022react} in that we do not store the earlier Thoughts. We notice empirically that just having the \textbf{Subtasks Completed} in the form of function inputs and outputs is enough for the LLM to understand past history to make an informed decision, and at the same time results in reduced context length.
    \item \textbf{Shared Variables.} This is a Python dictionary which stores Python variables. These Python variables will be made available to the Agent and all Equipped Functions / Inner Agents upon request. The exact names and values of these \textbf{Shared Variables} will not be in the prompt to LLM calls by default, meaning that this information will not increase context length unless explicitly referred to. As such, we are able to store lengthy text output as well as filenames for various other modalities for suitable pre-processing when needed later on. The Equipped Functions / Inner Agents are also allowed to modify these \textbf{Shared Variables}, and as such can directly update the 
    Shared Memory whenever needed.
\end{enumerate}

\subsection{Global Context}
\textbf{Global Context} augments the default LLM prompt for the Agent. We use \textbf{Global Context} to expose certain persistent variables, typically stored in \textbf{Shared Variables}, which we want to carry through the task / carry across tasks. This is very useful for letting the Agent know the current state in a dynamically changing environment. \textbf{Global Context} can also contain more specific instructions for the LLM beyond the defaults in TaskGen. 


\subsection{Memory Bank}
The \textbf{Memory Bank} contains all the important information that an Agent might need to know for an arbitrary task. We posit that a generic problem solver will need to contain memory at \textbf{multiple forms of abstraction}. For instance, when given a piece of text, we can store the 1) summary of it, 2) extracted entities and relationships in a knowledge graph, 3) entire text. These information will be useful when we are doing 1) generic question and answer, 2) causal reasoning, 3) specific question and answer respectively. If we just store information at one form of abstraction only (e.g. summary), some tasks will be significantly harder or impossible (e.g. find out specific details in text).

\textbf{Task-Augmented Prompt.} When given a task, we extract out the relevant memories using RAG or other semantic matching algorithms. This will be used to augment the LLM prompt when selecting the next subtask and using the \texttt{use\_llm} function. 


\textbf{Equipped Function Filtering by Task.} Furthermore, when given a task, not all Equipped Functions/Inner Agents are relevant, so we can filter them by semantic similarity to the task. This will help improve LLM performance provided that the correct functions are kept.



\subsection{Other Notable Features}
\textbf{Conversable Agent.} TaskGen provides a wrapper for a two-person chat interface with the Agent, where the Agent can use its Equipped Functions to perform actions and then reply the User.

\textbf{Code Generator.} TaskGen has an in-built code generator and code corrector, which can also be used to perform actions with Python code, similar to CodeAct. \cite{wang2024executable}

\textbf{Asynchronous Mode.} TaskGen has asynchronous equivalents of \texttt{strict\_json}, \texttt{Function} and \texttt{Agent} classes for faster asynchronous processing.

\textbf{Community Contributions.} TaskGen has a community space where users can easily upload and download Agents (see Appendix \ref{appendix: community contribs}).

\section{Evaluation}
We evaluate TaskGen on various environments to showcase its versatility: dynamic maze navigation (see Appendix \ref{appendix: maze}), escape room solving in TextWorld (see Appendix \ref{appendix: text-world}), web browsing (see Appendix \ref{appendix: web-browsing-agent}), MATH dataset (see Appendix \ref{appendix: MATH}), RAG-based Question Answering (QA) on NaturalQuestions dataset (see Appendix \ref{appendix:naturalquestions_rag}).

\section{Results}
Overall, TaskGen works well for generic environments. The summarised results for each environment are as follows:
\begin{enumerate}[leftmargin=*,nosep,wide=0pt]
\item \textbf{Dynamic Maze Navigation.} We implement a 40x40 maze with obstacles that change halfway during the Agent's learning, similar to Learning, Fast and Slow \cite{tan2023learning}. TaskGen with \textbf{Global Context} and an external StrictJSON Planner manages to solve \textbf{100\%} of the episodes on the first try, even after environment changes.
\item \textbf{Escape Room Solving in TextWorld.} We used TaskGen as a generic interactive fiction player to solve TextWorld \cite{cote2019textworld} challenges. Where dense rewards and detailed goals were provided, TaskGen achieved a \textbf{96\%} solve rate, outperforming a neural-network agent's \cite{cote2024textworld} solve rate of \textbf{88\%}. Where commands are not provided and needed to be derived by the agent, TaskGen achieved an \textbf{88\%} solve rate, outperforming the baseline LLM's \textbf{57\%}.
\item \textbf{Web-Browsing Agents.} We designed a series of tasks requiring agents to navigate and extract information from the web, simulating real-world scenarios where users need to find specific information across various websites. Tasks included searching for academic studies, gathering news headlines, summarising market trends, and exploring educational resources. The agent demonstrated varying levels of success across different tasks, with \textbf{69\%} of actions being completed successfully.
\item \textbf{MATH Dataset.} We randomly selected 20 problems from the test set of 5 categories (Algebra, Pre-Algebra, Intermediate Algebra, Number Theory, and Counting and Probability) of the MATH dataset \cite{Hendrycks2021MeasuringMP}. Our experiments (see Appendix \ref{appendix: MATH}) showed that the TaskGen Agent with Equipped Functions achieved an average accuracy of \textbf{71\%} on challenging Level-5 problems, compared to \textbf{44\%} accuracy for the Agent without these functions. This demonstrates that imbuing an Agent with code generation and debugging capabilities significantly improves problem-solving. 

\item \textbf{RAG-based QA on NaturalQuestions.} On the \href{https://github.com/google-research-datasets/natural-questions}{Natural Questions dataset} \cite{kwiatkowski2019natural}, TaskGen with Equipped Functions for dynamic retrieval and answering (we term this Interactive Retrieval) outperformed the baseline LLM with RAG across all metrics (see Appendix \ref{appendix:naturalquestions_rag}). Compared to the baseline LLM, Interactive Retrieval achieved an F1 Score of \textbf{47.03\% (+5.49\%)}, precision of \textbf{40.75\% (+7.43\%)}, and recall of \textbf{55.59\% (+0.42\%)}, demonstrating TaskGen's effectiveness in dynamically refining context for more accurate question answering.

\end{enumerate}

\section{Conclusion and Future Work}
TaskGen is already used in production at \href{https://simbian.ai/}{Simbian AI}, and we would like to share its benefits with others. TaskGen's approach of not using conversation, but instead focusing directly on solving the task is a marked improvement over most existing agentic frameworks. TaskGen will continue to be actively developed over the coming years. The future work includes: 1) better planning abilities using state-based graphs, parallel searching, 2) multiple memory abstraction spaces such as vector databases and knowledge graphs, 3) reflection as a way to consolidate experiences and use for future decision making, 4) extended multi-modal support and 5) multiple agents with different skills and biases collaborating with one another. 

\textbf{Towards Hybrid Workflows.} As demonstrated by systems like AGENTless \cite{xia2024agentless}, full end-to-end agentic workflows may not always provide the best performance, as we may want to fix parts of the processes without Agents if we already know what needs to be done. This mixture of fixed processes and flexible agentic process selection will form the core tenet of future agentic systems. While not featured in native TaskGen, such hybrid systems can be implemented by using StrictJSON or fixed rules for dynamic routing over TaskGen Agents. We will explore more of such approaches and incorporate key elements into TaskGen.

\section{Build Together With Us}
TaskGen is an actively developing framework, and we would love to seek your inputs / contributions / feedback. Build together with us via our \textbf{GitHub} (\href{https://github.com/simbianai/taskgen}{https://github.com/simbianai/taskgen}), and join the discussion group at \textbf{Discord} (\href{https://discord.com/invite/bzp87AHJy5}{https://discord.com/invite/bzp87AHJy5}).

\section{Experiment Details}
For more information on the following experiments in the Appendix, do contact the following:
\begin{enumerate}[leftmargin=*,nosep,wide=0pt]
\item Community Contributions (see Appendix \ref{appendix: community contribs}) \\- \textbf{Hardik} (\textit{hardik121998@gmail.com})
\item Dynamic Maze Navigation (see Appendix \ref{appendix: maze}) \\- \textbf{John Tan Chong Min}
\item TextWorld (see Appendix \ref{appendix: text-world}) \\- \textbf{Richard Cottrill}
\item Web-Browsing Agents (see Appendix \ref{appendix: web-browsing-agent}) \\- \textbf{Brian Lim Yi Sheng}
\item MATH Dataset (see Appendix \ref{appendix: MATH}) \\- \textbf{Bharat Runwal} (\textit{bharat.runwal@simbian.ai})
\item NaturalQuestions QA (see Appendix \ref{appendix:naturalquestions_rag}) \\- \textbf{Prince Saroj} 
\end{enumerate}

\section*{Limitations}
The experiments conducted in this paper are not extensive for all available LLMs. We mainly use OpenAI's "gpt-4o" and "gpt-3.5-turbo". That said, we have also empirically tested and verified, though not shown here, that TaskGen works with other LLMs such as OpenAI's "gpt-4o-mini", Llama-3 8B and Claude-3 Haiku.

\section*{Acknowledgements}
The research is supported by \href{https://simbian.ai/}{Simbian AI}, where it is used in the core products. This research is also supported by the National Research Foundation, Singapore under its AI Singapore Programme (AISG
Award No: AISG2-PhD-2021-01-003[T]) and by A*STAR,
CISCO Systems (USA) Pte. Ltd and National University of
Singapore under its Cisco-NUS Accelerated Digital Economy
Corporate Laboratory (Award I21001E0002).

\bibliography{emnlp_submission}
\bibliographystyle{acl_natbib}

\appendix

\onecolumn
\setlength{\parindent}{0pt}

\newpage
\noindent {\bf APPENDIX}
~\newline

The Appendix contains the following sections:
\begin{enumerate}[label=\Alph*]
    \item StrictJSON Details
    \item TaskGen Details
    \item Community Contributions to TaskGen
    \item Dynamic Maze Navigation
    \item Escape Room Solving in TextWorld
    \item Web-Browsing Agents
    \item MATH Dataset
    \item RAG-based Question Answering on NaturalQuestions Dataset
\end{enumerate}

The code for the experiments in the Appendix can be found at \href{https://github.com/simbianai/taskgen}{https://github.com/simbianai/taskgen}.

~\newline

\newpage
\section{StrictJSON Details}
\renewcommand{\thefigure}{A\arabic{figure}}
\renewcommand{\thetable}{A\arabic{table}}
\setcounter{figure}{0}
\setcounter{table}{0}
\label{appendix: strictjson}

StrictJSON is a library created in order to parse LLM output into a structured JSON format, and is used for all LLM calls in TaskGen. This enables efficient extraction of LLM output based on the JSON keys and enables interfacing the LLM as part of a larger system, such as the agentic framework in TaskGen. Furthermore, StrictJSON comes in-built with rule-based type checking which increases output reliability. StrictJSON also has error checking capabilities, which uses the JSON parsing errors or type checking errors to feed into the LLM in an iterative feedback loop as an error message to regenerate the JSON again. This is similar to the error feedback mechanism in Voyager \cite{wang2023voyager}.

\textbf{Comparison with json.loads()}: Typically, in order to parse JSON string into a dictionary, the function \texttt{json.loads()} is called. This is not robust to variations of the JSON and can easily fail to parse incorrectly formatted JSON, especially when generating code. StrictJSON is more robust, as it adds a delimiter before and after the key which the regex uses to extract. This regex will still work even if the quotation marks are not closed properly or are missing within the string. See Section \ref{StrictJSON: under the hood} for more details.

\textbf{Why not YAML?} YAML could also potentially be the format for LLM outputs in order to reduce token counts. However, YAML formatting performance has been empirically tested to be poorer than JSON, at least on the GPT models. We posit that this is because current LLMs are extensively trained on web data, of which JSON is more prevalent than YAML since it is the earlier format to be used. This may change as more web data is of YAML format. For now, JSON format is used to get a reliable system working.

This appendix details how to use StrictJSON based on TaskGen v3.2.0.

\subsection{Usage}

\begin{figure}[H]
    \centering
    \includegraphics[width=\textwidth]{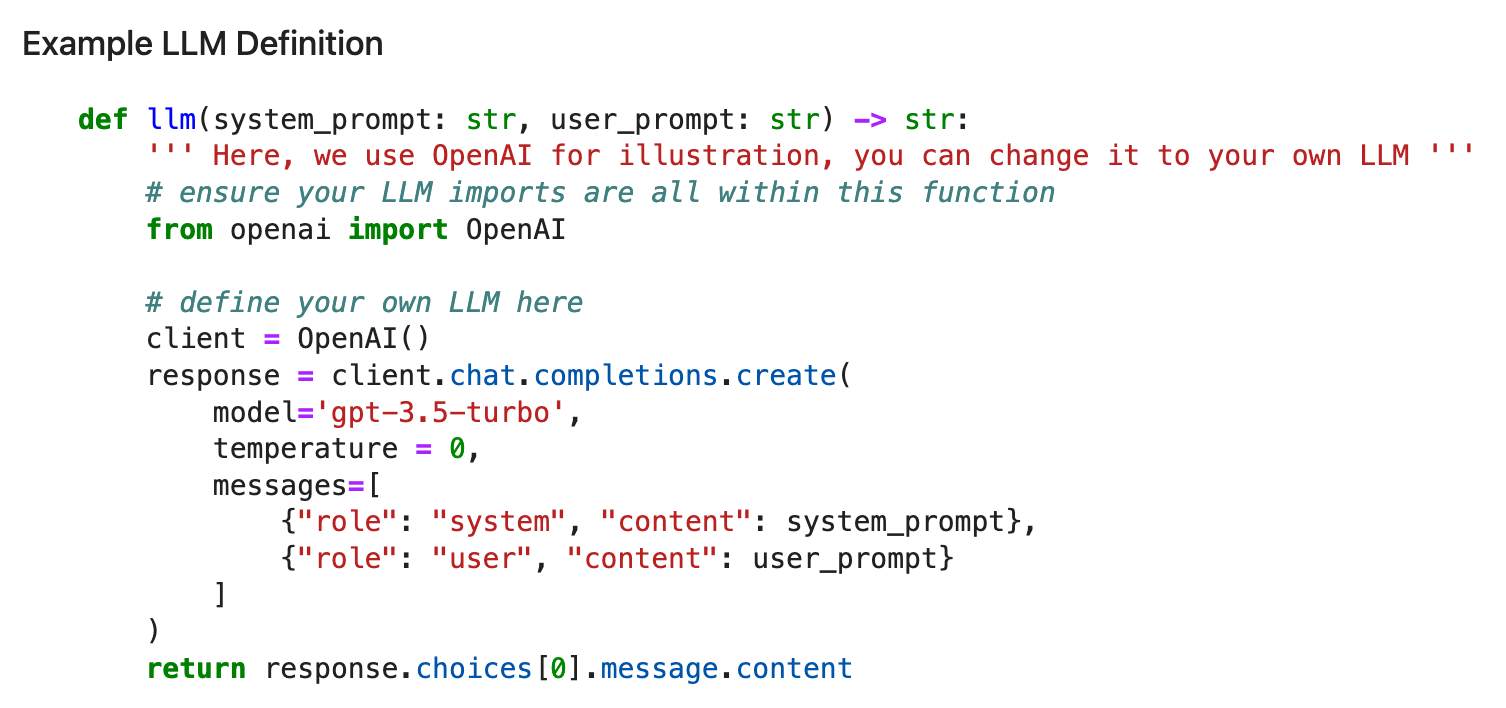}
    \caption{Example LLM Definition}
    \label{fig:llm_definition}
\end{figure}

To use StrictJSON, we firstly need to have an LLM available in order to generate the JSON from the text input given. Fig. \ref{fig:llm_definition} illustrates an example LLM function (named \texttt{llm}) that can be interfaced with StrictJSON. It takes as input the \textbf{system prompt}, which is the overall system message for the LLM, as well as the \textbf{user prompt}, which is what the user typically enters into the LLM for a response. This returns an LLM model response in the form of a string, which is the output of this LLM function. By exposing the entire LLM function to the user, StrictJSON is extremely versatile and can operate with both API-based LLM models and local models. 

\begin{figure}[H]
    \centering
    \includegraphics[width=\textwidth]{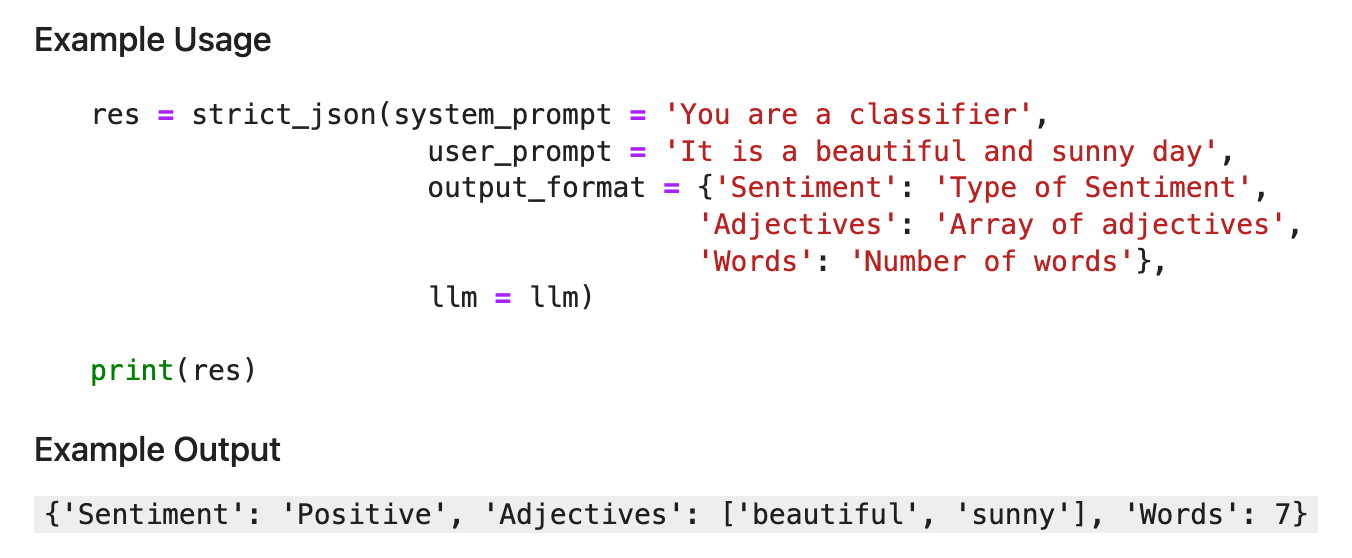}
    \caption{Basic Usage of StrictJSON}
    \label{fig:basic_example}
\end{figure}

In order to use StrictJSON to process the LLM's output, we simply use the \texttt{strict\_json} function. We give it the system prompt, user prompt, and the output format in a dictionary format with keys being the field name and values being the description of the field. For instance, Fig. \ref{fig:basic_example} illustrates how to use StrictJSON to classify a sentence in the user prompt. As can be seen, StrictJSON processes the type of sentiment, an array of adjectives in the sentence, and the number of words all in the same function call.

\begin{figure}[H]
    \centering
    \includegraphics[width=\textwidth]{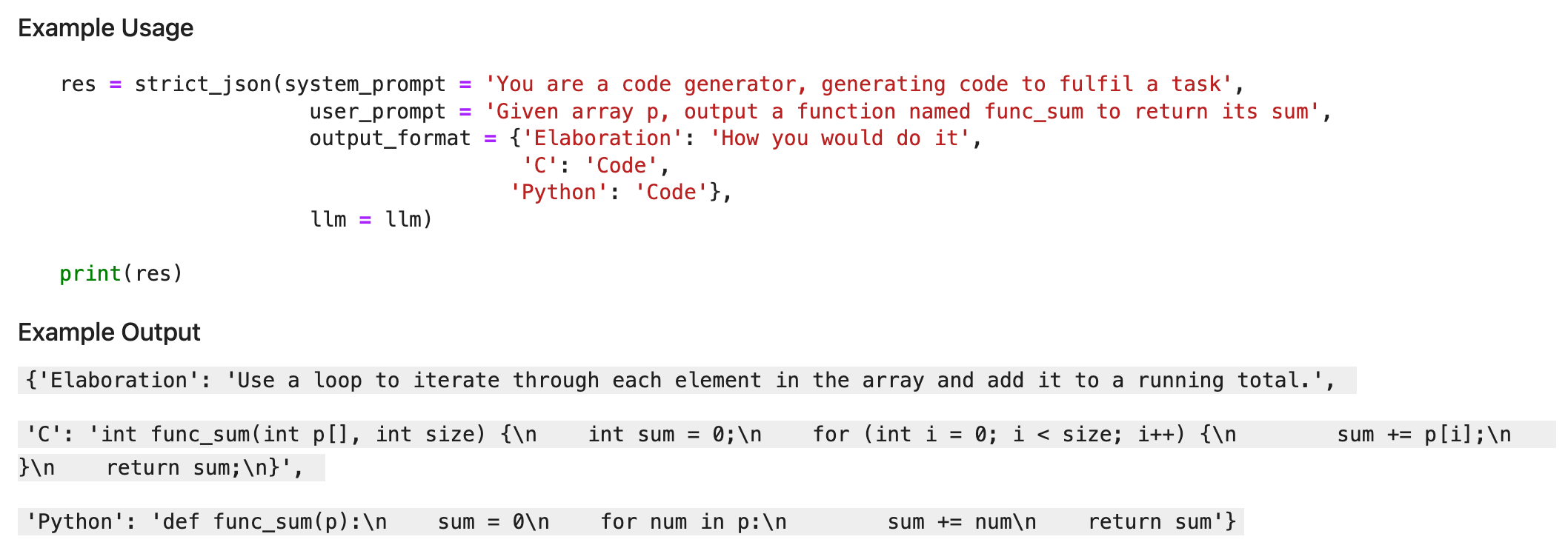}
    \caption{Advanced Usage of StrictJSON for code}
    \label{fig:code_generation}
\end{figure}

StrictJSON is also able to process code reliably, as shown in Fig. \ref{fig:code_generation}.

\begin{figure}[H]
    \centering
    \includegraphics[width=\textwidth]{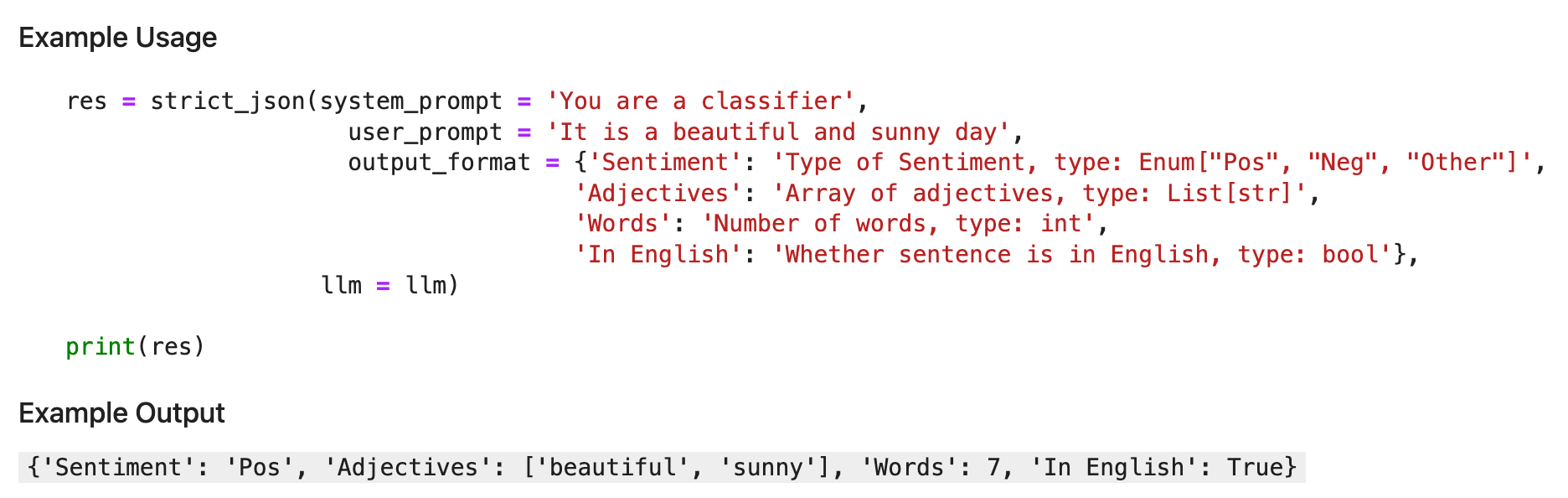}
    \caption{Type Checking in StrictJSON}
    \label{fig:type_checking}
\end{figure}

StrictJSON also supports type checking of the following types: int, float, str, dict, list, array, code, bool, Dict[], List[], Array[], Enum[]. If there is a [], you can nest datatypes within it such as List[int] for a list of integers. Only Dict[] cannot be nested, and Dict[dictionary\_keys] is used instead to enforce the presence of the dictionary\_keys within the dictionary. Fig. \ref{fig:type_checking} illustrates how to use StrictJSON with type checking. This can ensure greater output specificity and greater reliability for downstream tasks.

\subsection{How it works under the hood}
\label{StrictJSON: under the hood}

\begin{figure}[H]
    \centering
    \includegraphics[width=\textwidth]{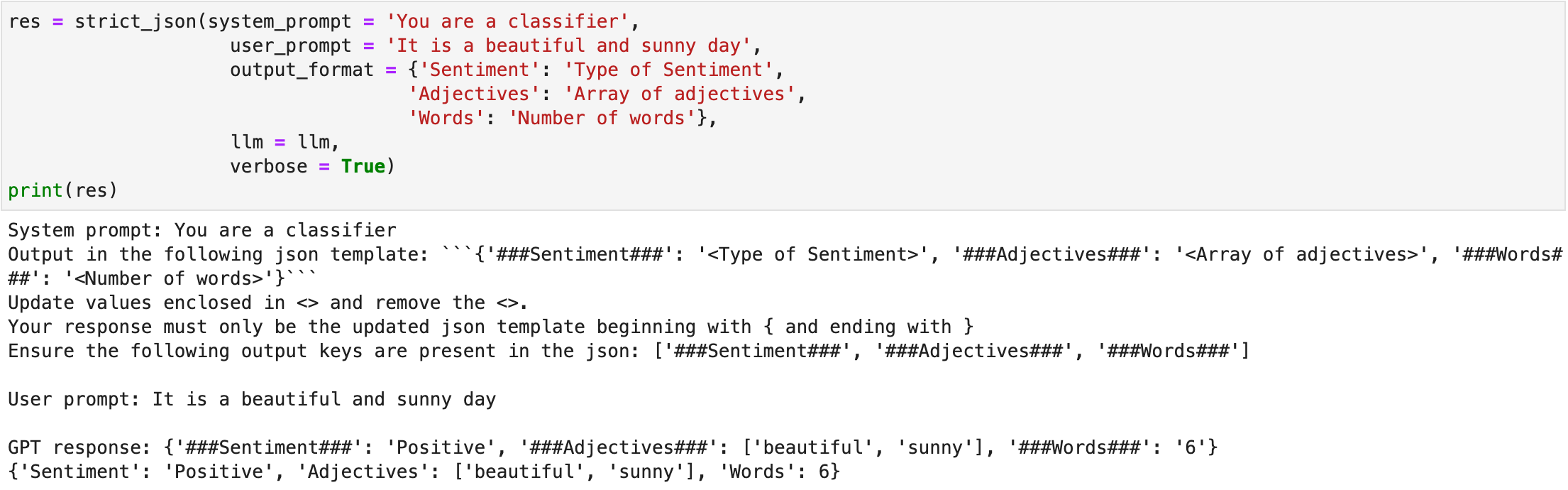}
    \caption{Visualising the actual LLM prompt that StrictJSON uses with \texttt{verbose = True}}
    \label{fig:under_the_hood}
\end{figure}

StrictJSON creates a prompt to the LLM to output JSON in a specified format using delimiters to enclose the output keys, that is more reliable to extract with regex as compared to unmodified keys of JSON. This is because the unmodified keys are just words with quotation marks, like "Sentiment", which may appear in other parts of the JSON and confuse the regex extraction. 

Fig. \ref{fig:under_the_hood} demonstrates how to visualise the actual LLM system and user prompt using \texttt{verbose = True} as a parameter to \texttt{strict\_json}. We can see that we get the LLM to enclose keys with delimiters (default `\#\#\#'), and enclose the JSON values with <>, which the LLM will be instructed to update.

\begin{figure}[H]
    \centering
    \includegraphics[width=\textwidth]{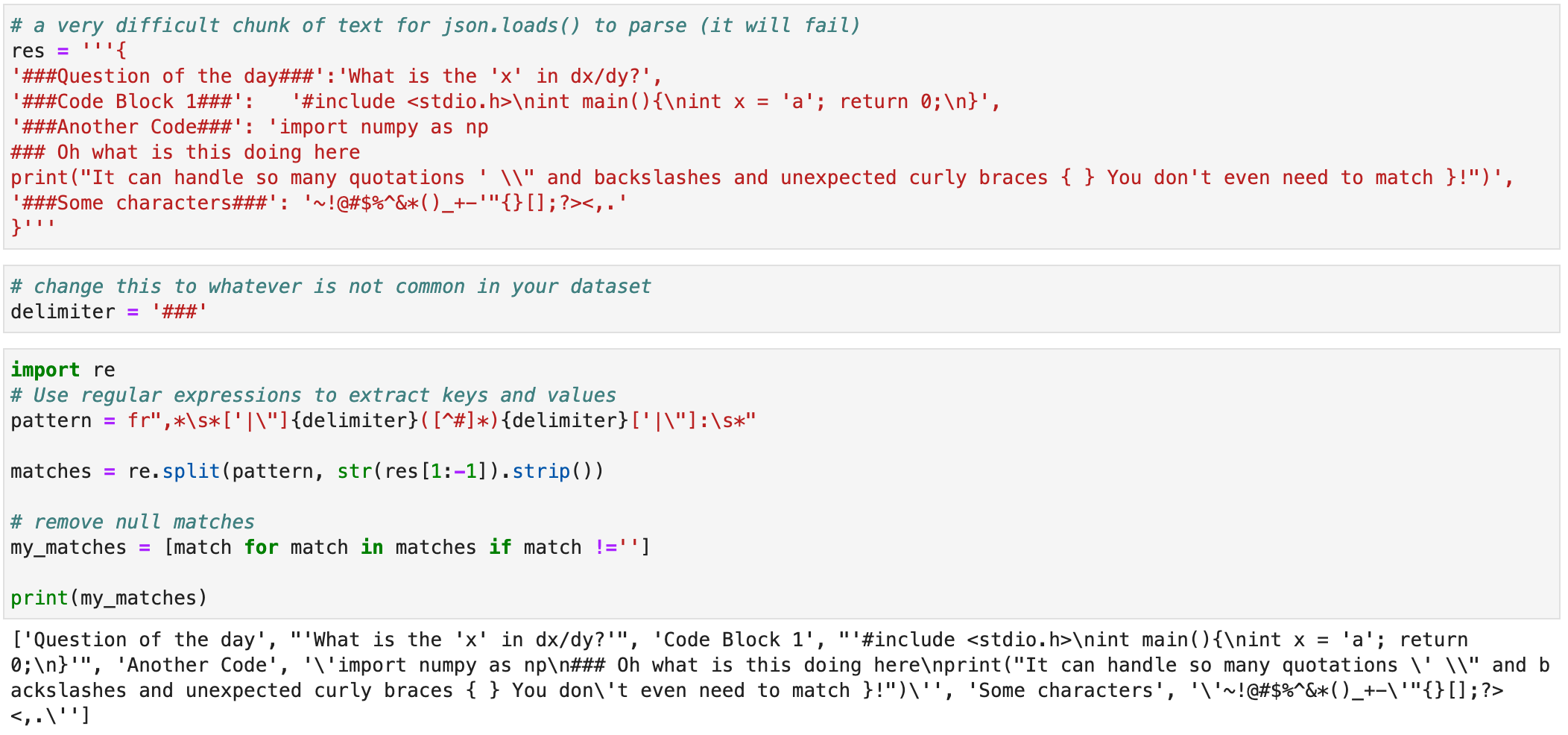}
    \caption{Regex is done on the delimiter + key + delimiter pattern}
    \label{fig:strict_json_regex}
\end{figure}

The regex that is used to parse the LLM output can be seen in Fig. \ref{fig:strict_json_regex}. By extracting keys of the form \texttt{`\#\#\#\{key\}\#\#\#'} or \texttt{"\#\#\#\{key\}\#\#\#"}, we can extract and parse the JSON even when there are mismatched quotation marks, unclosed brackets, and many other issues that will cause \texttt{json.loads()} to fail.

\newpage
\section{TaskGen Details}
\renewcommand{\thefigure}{B\arabic{figure}}
\renewcommand{\thetable}{B\arabic{table}}
\setcounter{figure}{0}
\setcounter{table}{0}
\label{appendix: taskgen}

This appendix details the various modules of TaskGen and how to use them based on TaskGen v3.2.0.

\subsection{Initialising TaskGen}
\begin{figure}[H]
    \centering
    \includegraphics[width=\textwidth]{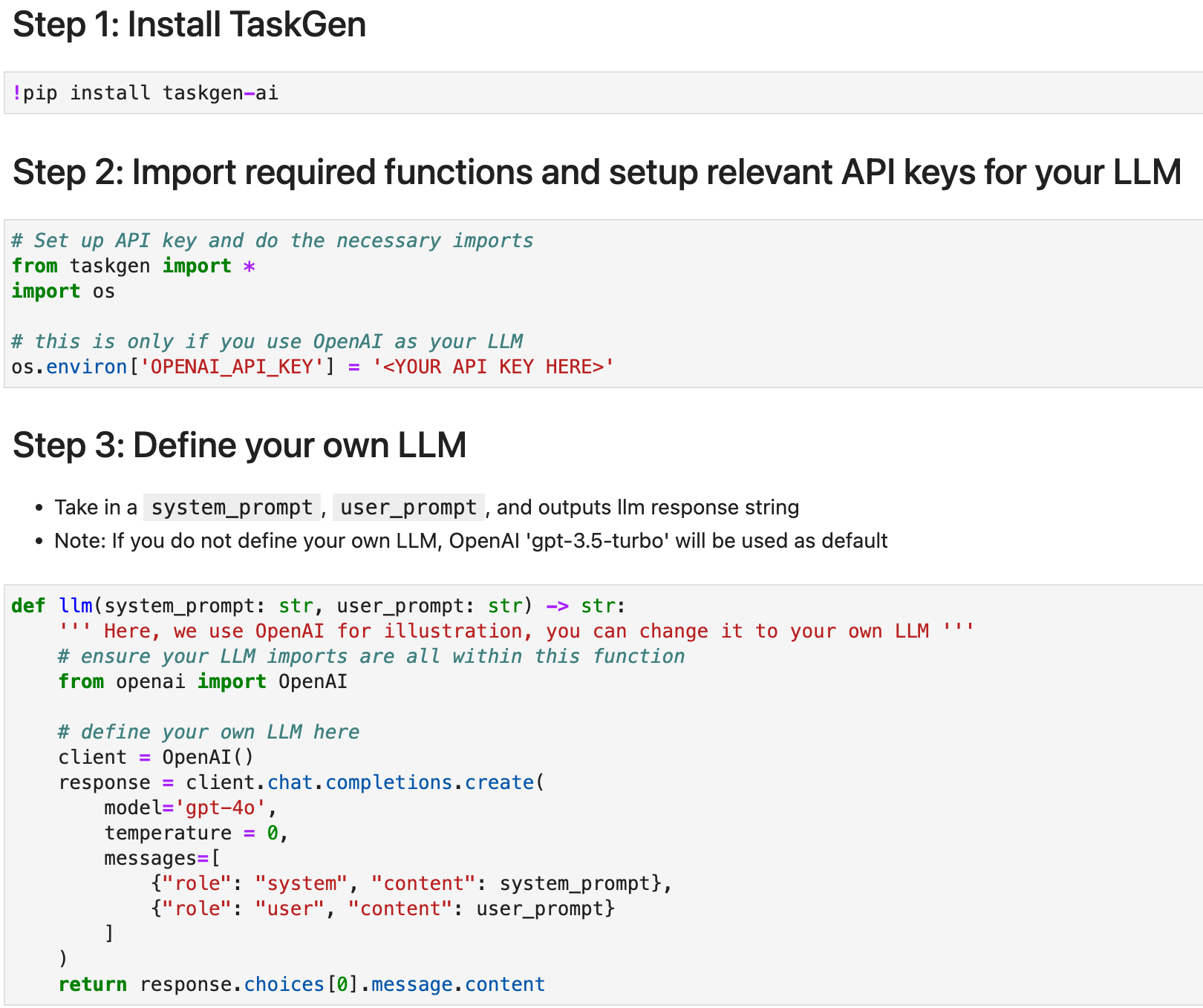}
    \caption{3 Steps to Initialise TaskGen}
    \label{fig:taskgen_init}
\end{figure}

Fig. \ref{fig:taskgen_init} shows how to initialise TaskGen. Here, we use "gpt-4o", but TaskGen can also work with "gpt-3.5-turbo" or equivalent LLM models at the cost of lower performance.

There three steps are:
\begin{enumerate}
\item Install TaskGen
\item Import required functions and setup relevant API keys for your LLM
\item Define your own LLM, which takes in a system prompt and user prompt and outputs the response string from the LLM
\end{enumerate}

\newpage
\subsection{TaskGen Agent Overview}
\subsubsection{Initialising the Agent}
\begin{figure}[H]
    \centering
    \includegraphics[width=\textwidth]{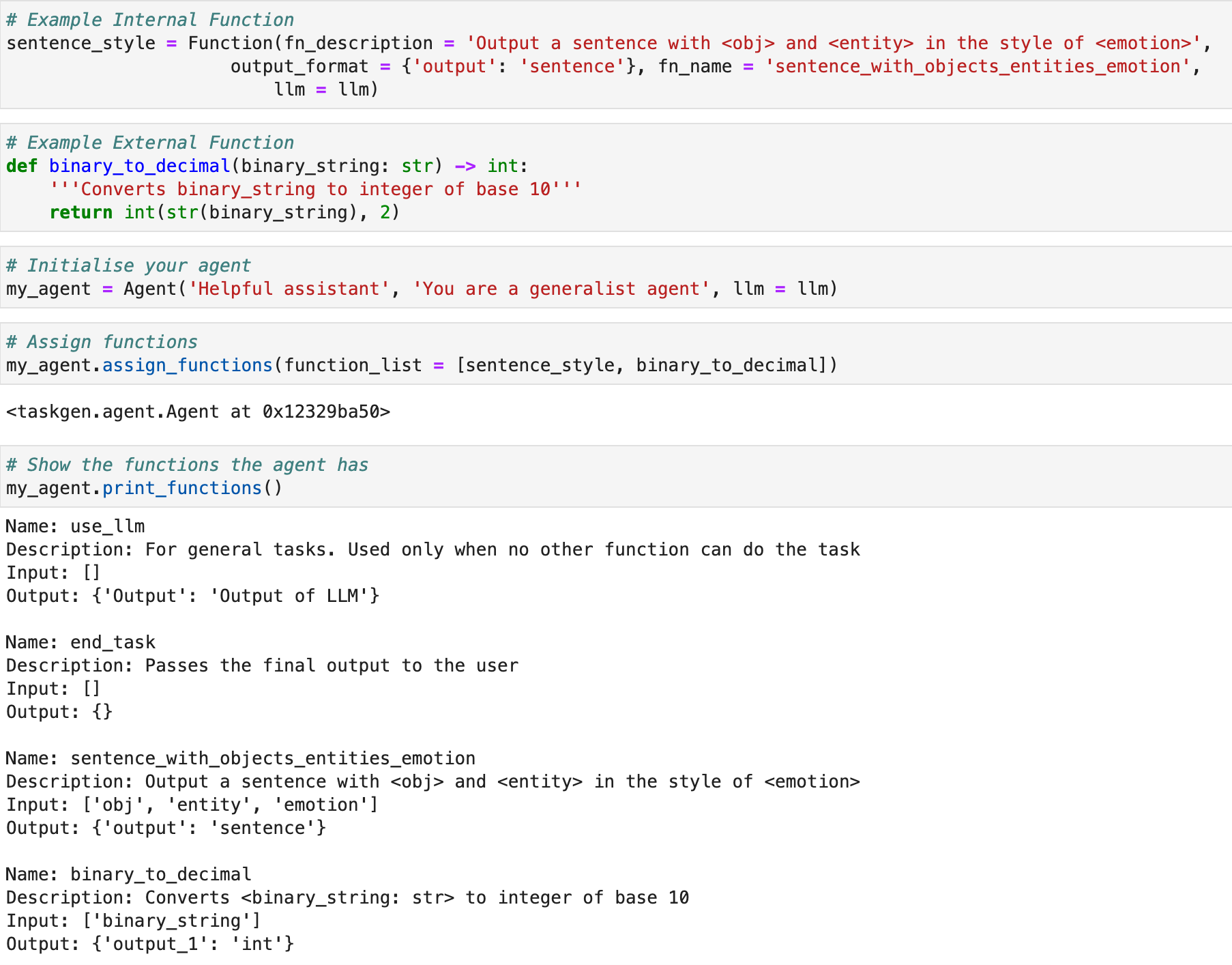}
    \caption{Initialising the Agent}
    \label{fig:taskgen_agent_init}
\end{figure}

Fig. \ref{fig:taskgen_agent_init} shows how to initialise the Agent. 

We firstly define the functions for the Agent. 

This can be of the form of an Internal Functions using \texttt{Function} class, which takes in the function description and output format of the function. We denote the variables in function description via <> enclosing the variable name. The output format is in the style of StrictJSON's output format. The Internal Function uses LLM to process the function, leading to very flexible functions that rule-based solutions may not allow for.

Functions can also be of the form of an External Function, which is very flexible as it is just a Python function. We simply define the function with typing for inputs and outputs, and with a docstring that contains the input parameter names. If any of the typing or docstring is missing, we will omit them from the function description, but the External Function can still work. External Functions allow for both rule-based rigidity and LLM-based flexibility, as an LLM call can be made inside the External Function as well.

After defining our Functions, we define our Agent by calling \texttt{Agent(name, description, llm)}.

Thereafter, we proceed to assign our functions via \texttt{assign\_functions}.

To see how the functions look like, we can also use \texttt{print\_functions} to visualise it. Notice that the functions just consists of Name, Description, Input and Output fields, which is much shorter than the JSON schema or Pydantic way of defining a function.

\newpage
\subsubsection{Running the Agent}
\begin{figure}[H]
    \centering
    \includegraphics[width=\textwidth]{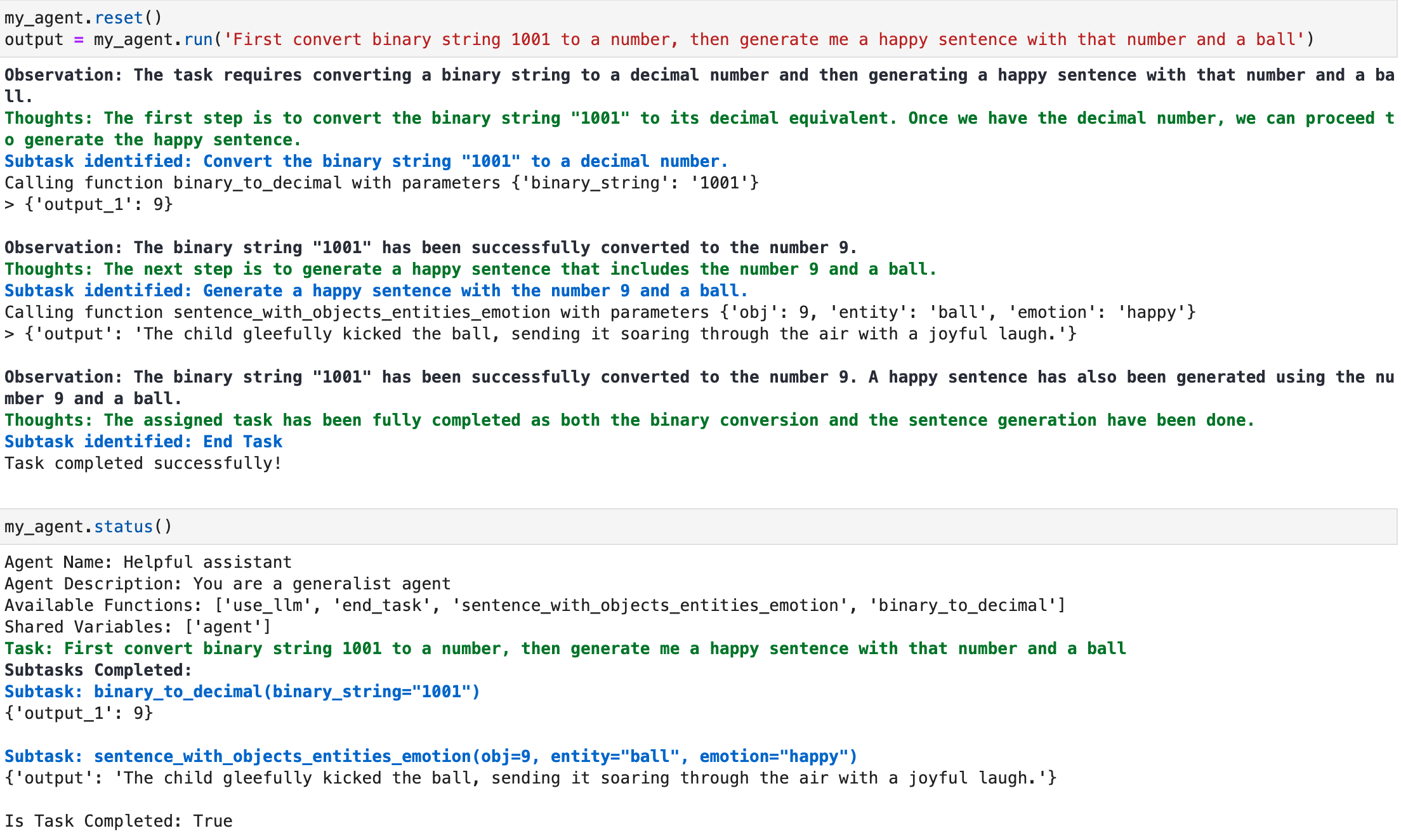}
    \caption{Running the Agent}
    \label{fig:taskgen_agent_run}
\end{figure}

Fig. \ref{fig:taskgen_agent_run} shows how to assign a task and run the Agent by simply calling \texttt{run(task)}. Notice how we can visualise the output via Observation, Thoughts, Action (Subtask) in the traditional ReAct framework. The difference between TaskGen and the original ReAct framework is that the observation here is actually the observation of the \textbf{Subtasks Completed} instead of the Observation of the function's output. By structuring Observation this way, this helps to provide a summary of what has been done so far, which aids in decision making.

We also do not store these Observation and Thoughts as they are just used in decision making at that point of time, but not needed in the longer term. The entire history of what has been done is stored in Subtasks Completed, which can be visualised via \texttt{status()} or via the \texttt{subtasks\_completed} variable of the agent.

Notice also that calling \texttt{status()} also gives us the Agent's details, such as Agent Name, Agent Description, Equipped Functions, Shared Variable Names, Assigned Task, Subtasks Completed, and whether the task is completed. We can call \texttt{status()} anytime to check on how the Agent is performing.

\newpage
\subsubsection{Querying the Agent}
\begin{figure}[H]
    \centering
    \includegraphics[width=\textwidth]{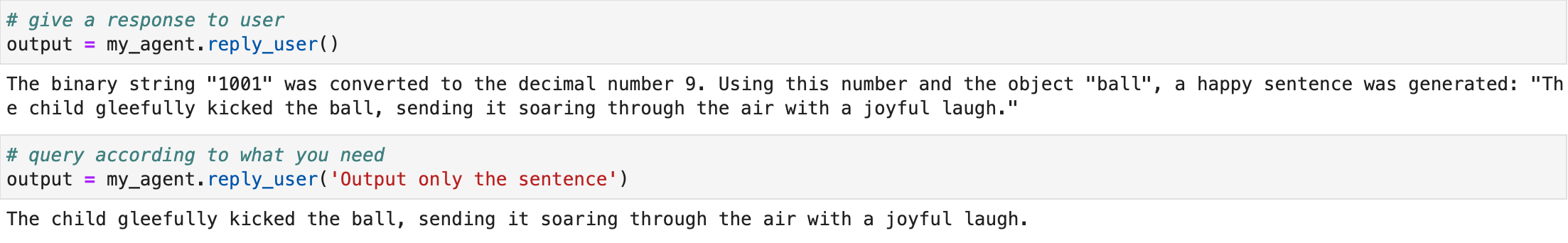}
    \caption{Querying the Agent}
    \label{fig:taskgen_agent_query}
\end{figure}

Fig. \ref{fig:taskgen_agent_query} shows how we can reply the user by simply calling \texttt{reply\_user()} to get the Agent to reply based on what has been done in \textbf{Subtasks Completed}. If \texttt{reply\_user()} is called without any query parameter, it will reply based on the assigned task. If there is a query parameter given, then it will reply based on the query.

This functions as a simple question answer bot, from which we can ask multiple questions about what the Agent has done so far and reply the user.

\subsubsection{Asynchronous Agents}

We can perform whatever we did for the Agent in asynchronous mode too. Such an asynchronous runtime has advantages in that we can run multiple Agents in a shorter time, as we can effectively let other Agents run in the downtime of one Agent.

TaskGen has two main classes - \texttt{Agent} and \texttt{Function}. Their asynchronous equivalents are \texttt{AsyncAgent} and \texttt{AsyncFunction}. Furthermore, the asynchronous version of \texttt{strict\_json} is \texttt{strict\_json\_async}.

\begin{figure}[H]
    \centering
    \includegraphics[width=\textwidth]{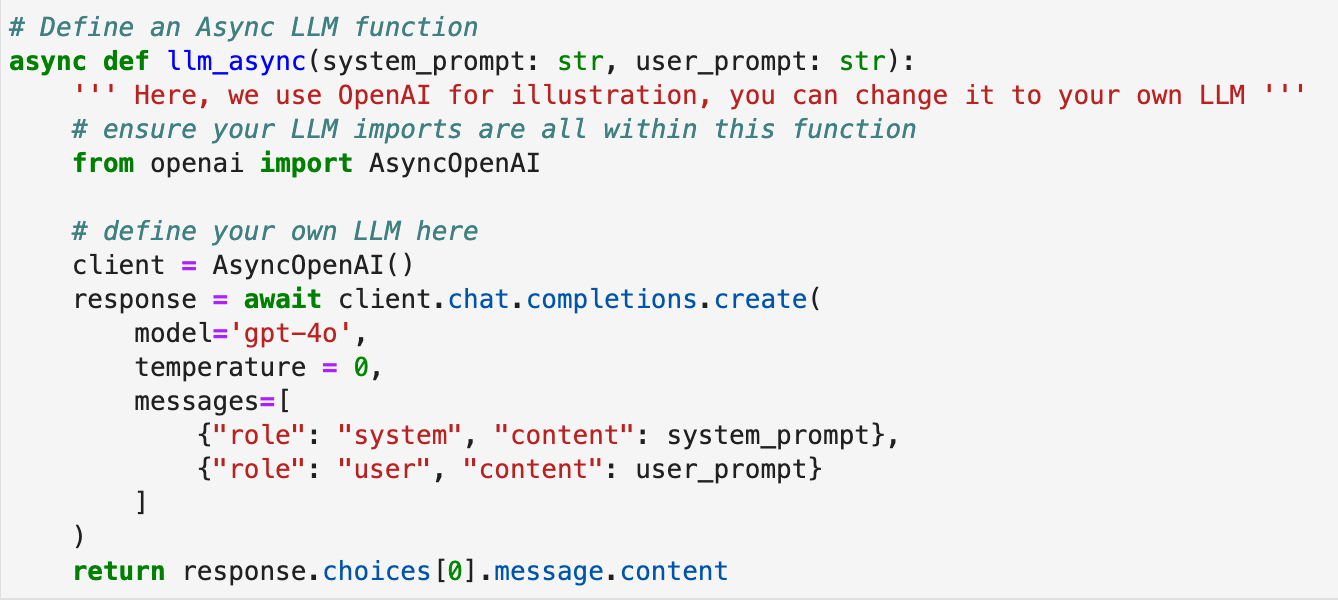}
    \caption{Initialising an Asynchronous Agent}
    \label{fig:taskgen_agent_async_init}
\end{figure}

Fig. \ref{fig:taskgen_agent_async_init} shows how to initialise the asynchronous LLM. Simply define a function that takes in a system prompt and user prompt, and outputs the response string of the LLM operating in asynchronous mode.

\begin{figure}[H]
    \centering
    \includegraphics[width=\textwidth]{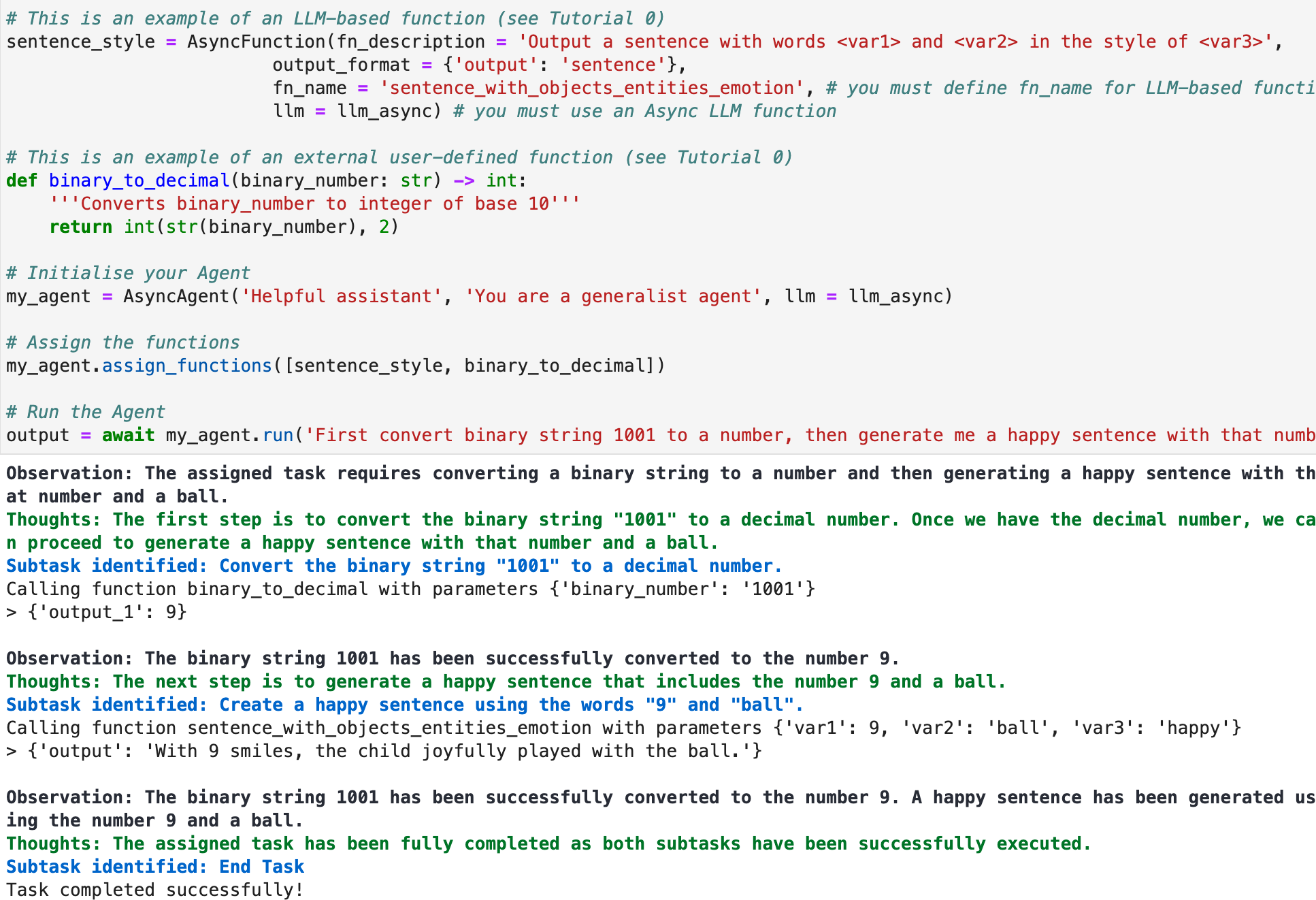}
    \caption{Initialising and Running an Asynchronous Agent}
    \label{fig:taskgen_agent_async_run}
\end{figure}

\begin{figure}[H]
    \centering
    \includegraphics[width=\textwidth]{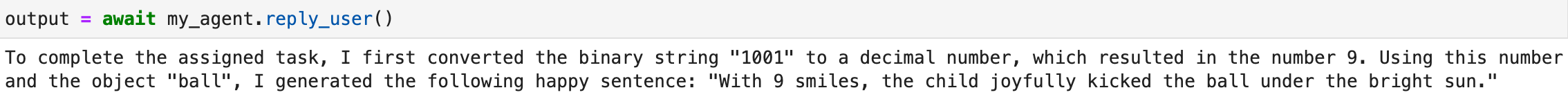}
    \caption{Querying an Asynchronous Agent}
    \label{fig:taskgen_agent_async_query}
\end{figure}

Figs. \ref{fig:taskgen_agent_async_run} and \ref{fig:taskgen_agent_async_query} shows how to initialise and run the \texttt{AsyncAgent} and \texttt{AsyncFunction}. As a general guide, to use the \texttt{AsyncAgent} and \texttt{AsyncFunction}, we do the same as what we would do for the synchronous version, and just put in the llm variable as the asynchronous version of the LLM.

When running the methods of \texttt{AsyncAgent}, we add an \texttt{await} keyword in front of them, like \texttt{await my\_agent.run()} and \texttt{await my\_agent.reply\_user()}. The outputs and how these methods work are similar to the synchronous versions.

\newpage
\subsection{Meta Agent}

Sometimes, due to task complexity, we would like to assign our Agent another Agent as an Equipped Function. Henceforth, our main Agent will be termed the Meta Agent, and the Agent equipped to it be termed the Inner Agent.

\subsubsection{Initialising the Meta Agent}
\begin{figure}[H]
    \centering
    \includegraphics[width=\textwidth]{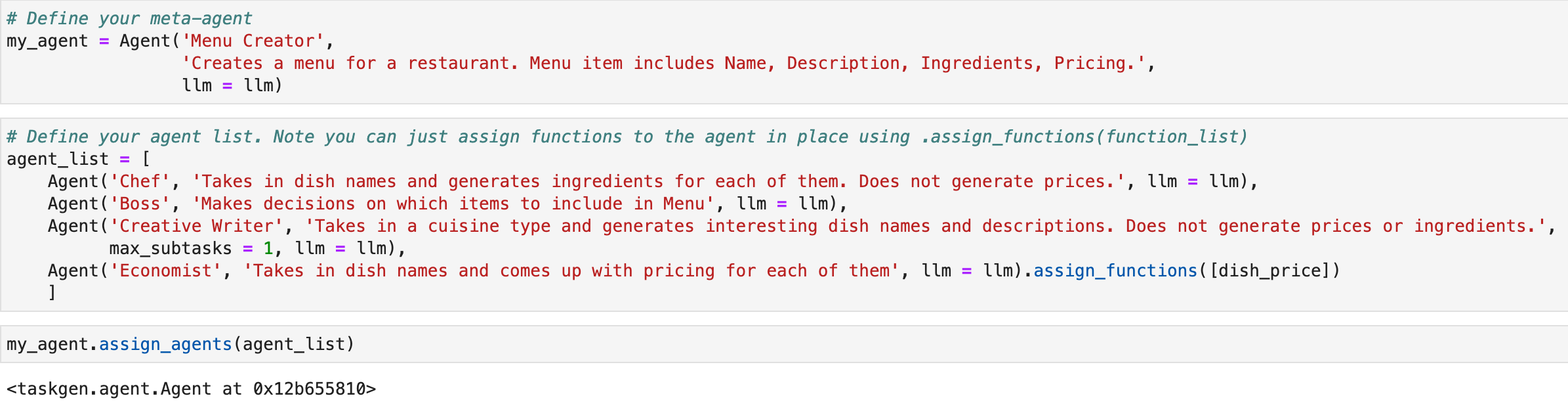}
    \caption{Initialising the Meta Agent}
    \label{fig:taskgen_meta_agent_init}
\end{figure}

Fig. \ref{fig:taskgen_meta_agent_init} show how to initialise the Meta Agent. It is generally the same process as initialising functions to the Agent, except that this function is of class "Agent". Note that we can specify how each Inner Agent should behave, including the max\_subtasks it should run for and what LLM it should use.

The Inner Agents will have full access to the \textbf{Subtasks Completed} and \textbf{Shared Variables} of the Meta Agent, and all the Equipped Functions of the Inner Agents will have access to these as well. This helps ensure that the context of the Meta Agent is fed downwards to the Inner Agents, and the Inner Agents can also change the Shared Memory of the Meta Agent.

\newpage
\subsubsection{Running the Meta Agent}
\begin{figure}[H]
    \centering
    \includegraphics[width=\textwidth]{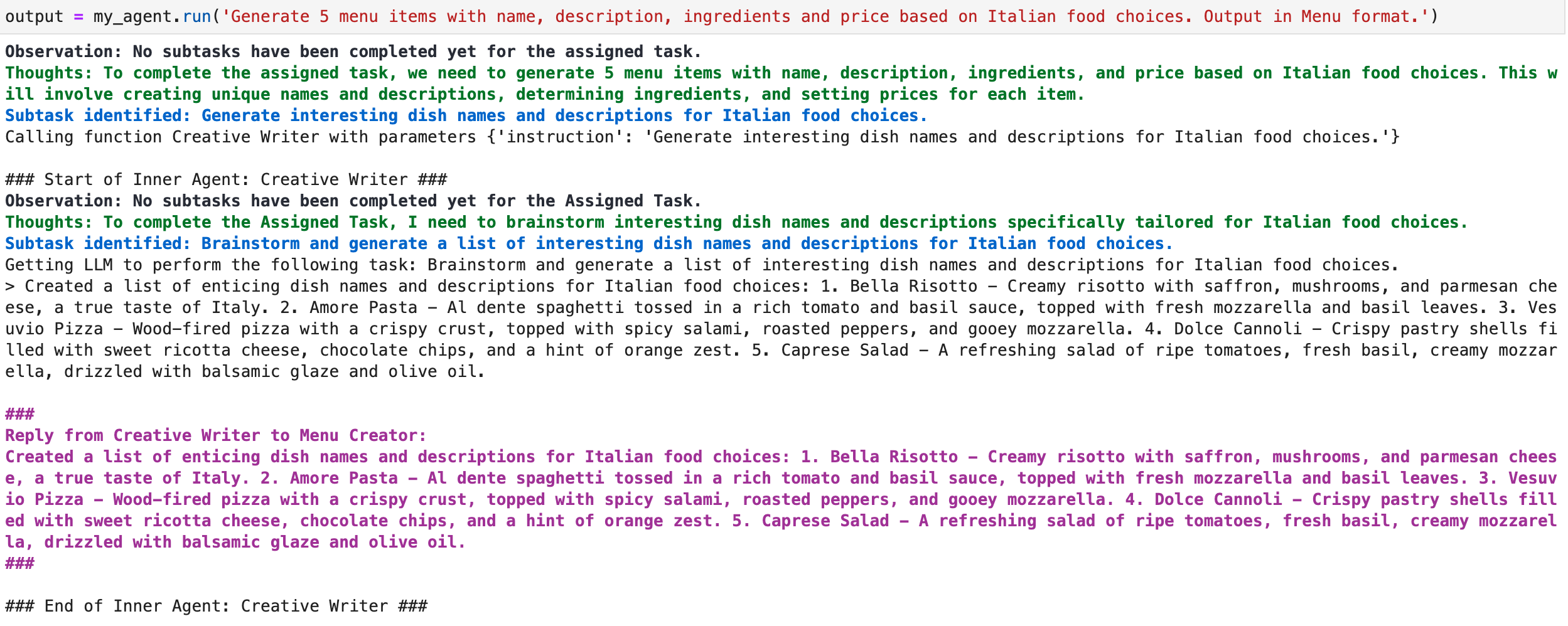}
    \caption{Running the Meta Agent (Part 1 - Creative Writer)}
    \label{fig:taskgen_meta_agent_run_1}
\end{figure}

\begin{figure}[H]
    \centering
    \includegraphics[width=\textwidth]{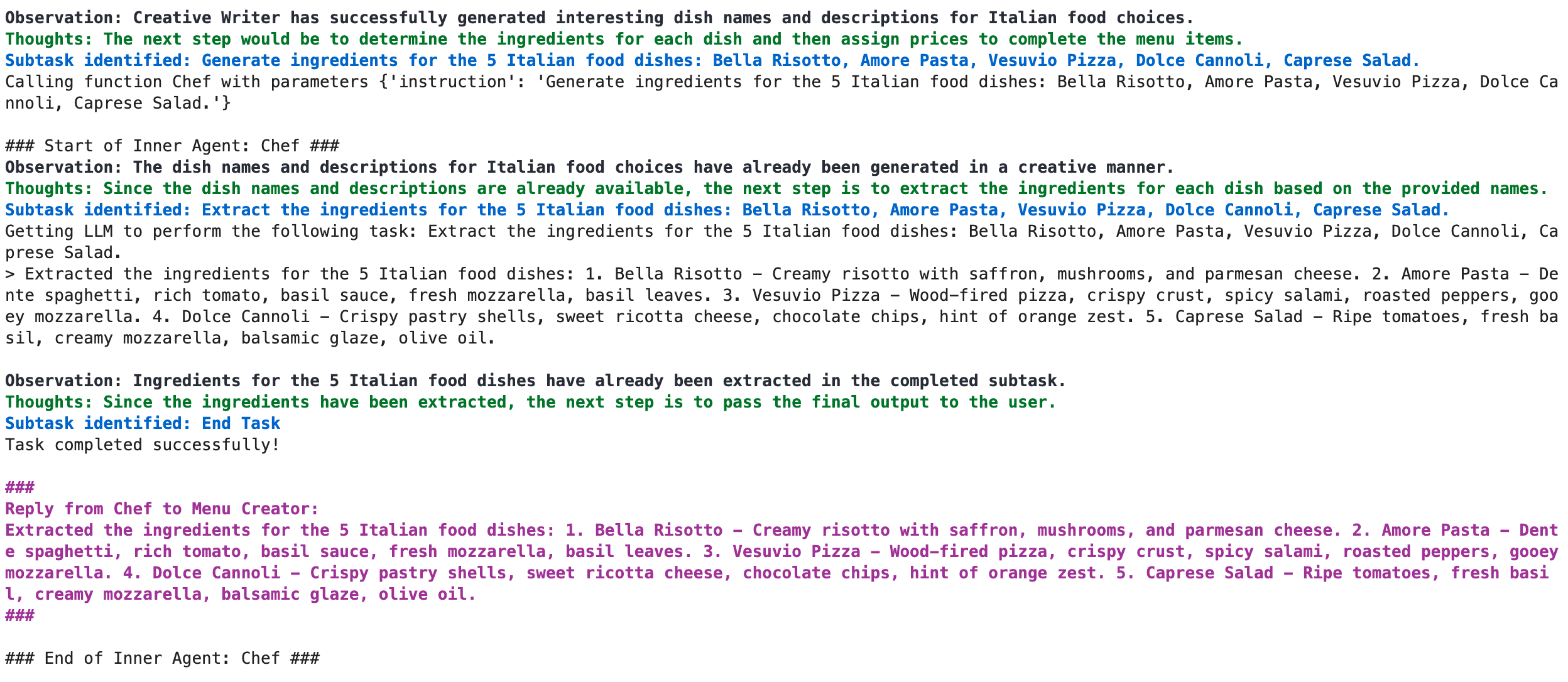}
    \caption{Running the Meta Agent (Part 2 - Chef)}
    \label{fig:taskgen_meta_agent_run_2}
\end{figure}

\newpage
\begin{figure}[H]
    \centering
    \includegraphics[width=\textwidth]{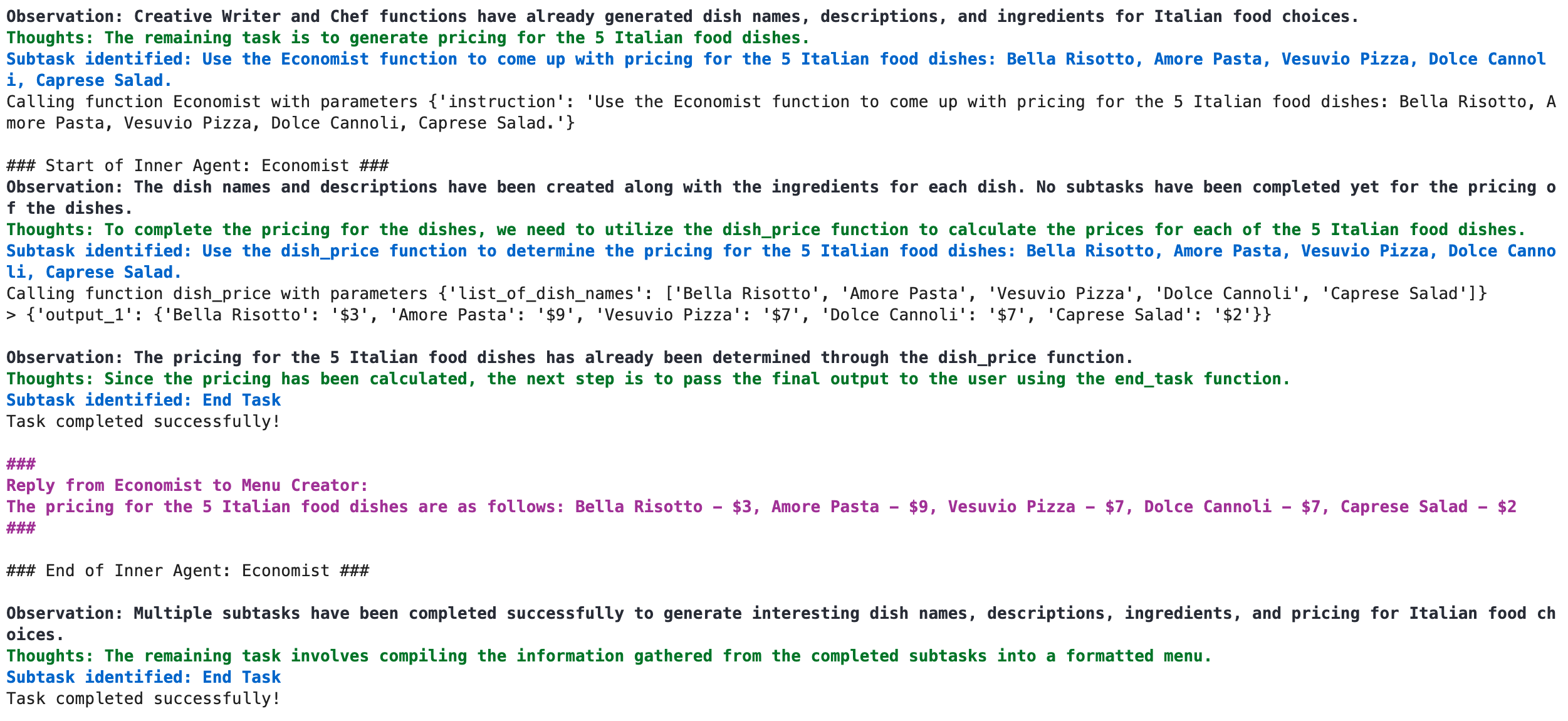}
    \caption{Running the Meta Agent (Part 3 - Economist)}
    \label{fig:taskgen_meta_agent_run_3}
\end{figure}

Figs. \ref{fig:taskgen_meta_agent_run_1}, \ref{fig:taskgen_meta_agent_run_2} and \ref{fig:taskgen_meta_agent_run_3} show the process of running the Meta Agent by simply calling \texttt{run()} and showcase the responses of the respective Creative Writer, Chef, Economist Inner Agents. 

Notice that if we call the Inner Agent as the function, we will generally repeat the Observation, Thoughts, Action (Subtask Identified) loop at the Inner Agent level. This kind of recursiveness helps to make the implementation of the Inner Agent easy, and we can stack as many Inner Agents as we would like to scale up the system.

We give the Inner Agent the full awareness of the Meta Agent's Assigned Task, \textbf{Subtasks Completed} and \textbf{Shared Variables}. When the Inner Agent ends the subtask, it does not give all the information back to the Meta Agent, but instead call \texttt{reply\_user()} to consolidate important information to put into \textbf{Subtasks Completed} (reply text shown in magenta). This helps to minimise the information stored in Shared Memory, which helps to reduce the overall context length, as many details done by the Inner Agent do not need to be known by the Meta Agent.

The Agents should generally be given context and Equipped Functions appropriate for their level of processing. In practice, such a hierarchical structure of Agents help with decomposing a complex problem into bite-sized bits, with the Agents at the higher levels focusing on the broader picture, while the Agents at the lower levels will do more of the specific details needed. This structure can be used to do most tasks that have such a hierarchical nature.

\newpage
\subsubsection{Visualising the Meta Agent's Status}
\begin{figure}[H]
    \centering
    \includegraphics[width=\textwidth]{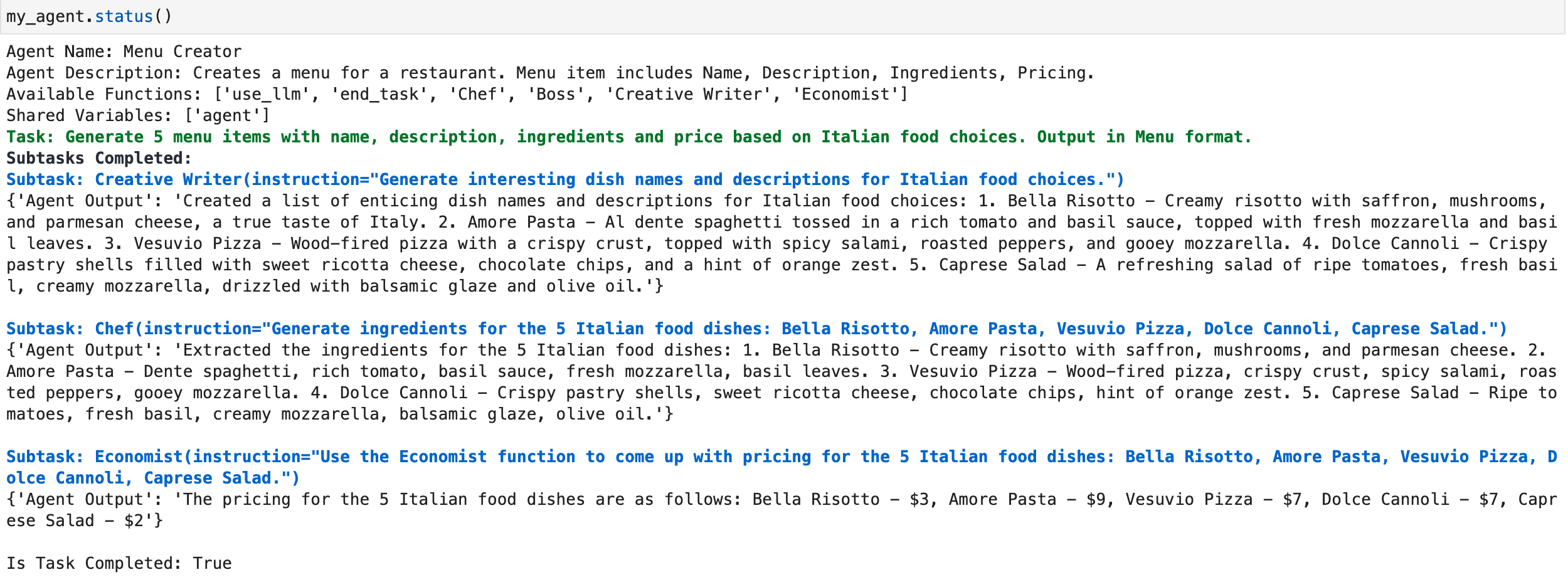}
    \caption{Visualising the Meta Agent's Status}
    \label{fig:taskgen_meta_agent_status}
\end{figure}

Fig. \ref{fig:taskgen_meta_agent_status} shows how to use \texttt{status()} to see the Meta Agent's status, including \textbf{Subtasks Completed}.

Here, we can see that the Inner Agents like Chef, Boss, Creative Writer, Economist are the Equipped Functions of the Meta Agent.

Furthermore, the Subtasks Completed shows which Inner Agent is called and what instruction was passed to each of them, along with their reply as the output when the subtask has ended.

\newpage
\subsubsection{Querying the Meta Agent}
\begin{figure}[H]
    \centering
    \includegraphics[width=\textwidth]{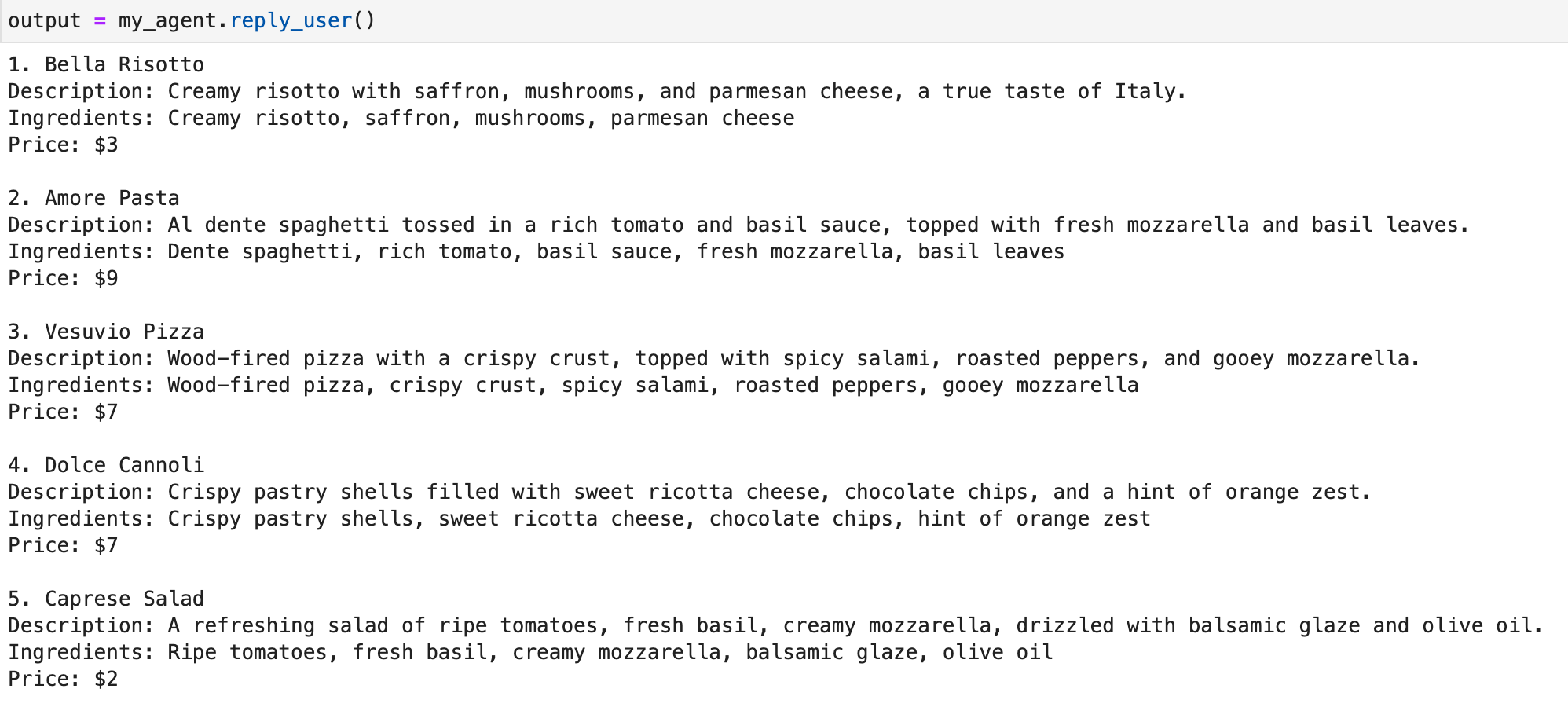}
    \caption{Querying the Meta Agent}
    \label{fig:taskgen_meta_agent_query}
\end{figure}

Fig. \ref{fig:taskgen_meta_agent_query} shows how to query the Meta Agent after the task is run using \texttt{query()}.

Here, we can see that the Agent is able to use the information in \textbf{Subtasks Completed} to give a coherent answer to what the user was asking, namely, to create a menu with 5 dishes with name, description, ingredients and price. In general, the more detailed the description of the Assigned Task, the better the answer by the Agent.

\newpage
\subsection{Shared Variables}

\subsubsection{Initialising Shared Variables}
\begin{figure}[H]
    \centering
    \includegraphics[width=\textwidth]{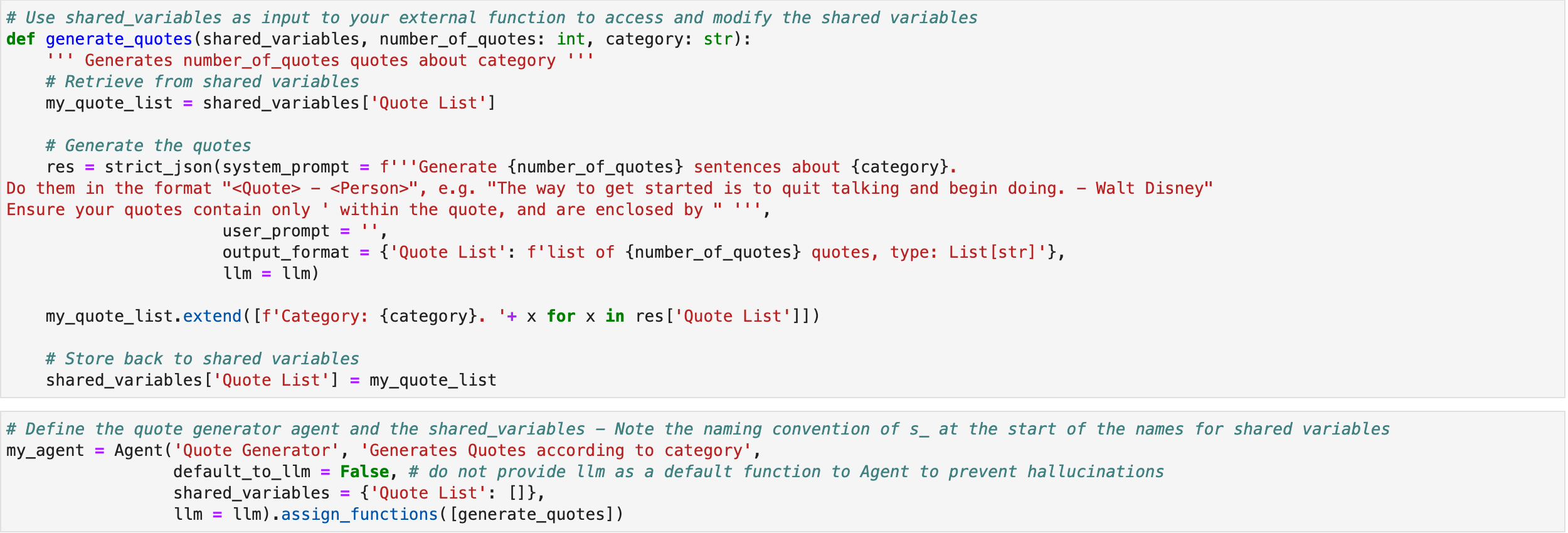}
    \caption{Initialising Shared Variables}
    \label{fig:taskgen_shared_variables_init}
\end{figure}

Fig. \ref{fig:taskgen_shared_variables_init} shows how to initialise the \textbf{Shared Variables}. In general, we call \texttt{shared\_variables} as a variable in the External Function, and proceed to extract and modify the relevant \texttt{shared\_variables} as appropriate for the Equipped Function. Here in \texttt{generate\_quotes}, we store the new generated quotes in the shared variable "Quote List".

Then, in order to use this shared variable in the Equipped Functions, we need to initialise the \texttt{shared\_variables} of the Agent. Here, we can see that we initialise "Quote List" as an empty list [].

\subsubsection{Modifying Shared Variables at Runtime}
\begin{figure}[H]
    \centering
    \includegraphics[width=\textwidth]{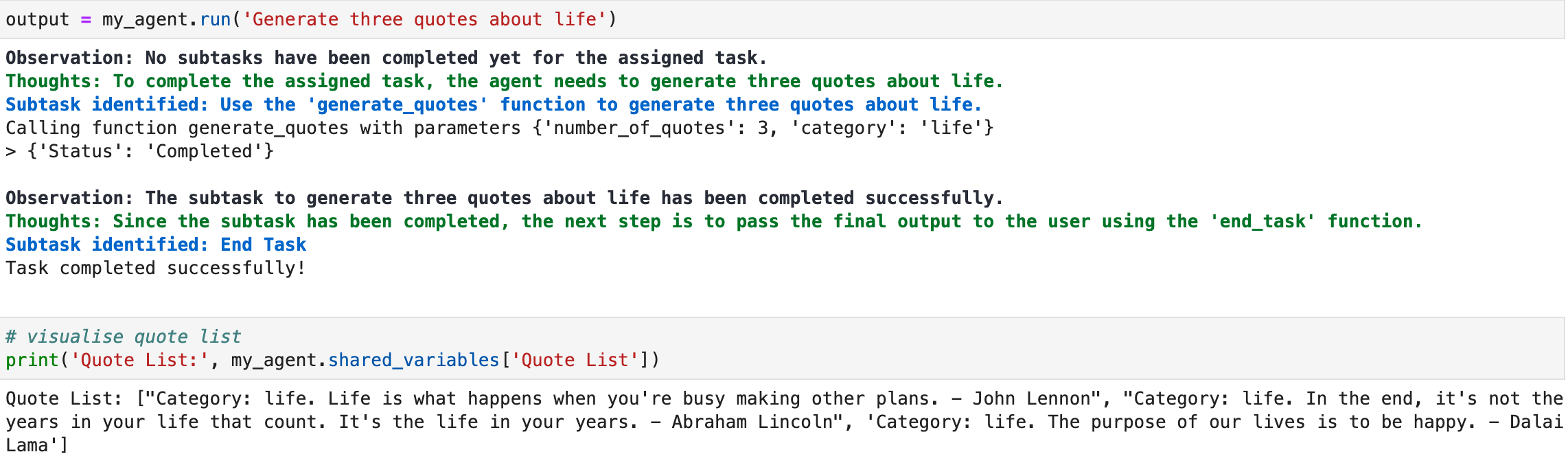}
    \caption{Modifying Shared Variables at runtime}
    \label{fig:taskgen_shared_variables_run}
\end{figure}

Fig. \ref{fig:taskgen_shared_variables_run} shows how we can modify \textbf{Shared Variables} at runtime. The function \texttt{generate\_quotes} was called, but the quotes did not appear in \textbf{Subtasks Completed} since \texttt{generate\_quotes} does not return any output. Rather, we store the generated quotes in the shared variable "Quote List". This helps reduce the overall context length for the Agent as the details for the quotes do not matter for this situation - only the fact that the quotes are generated does. This is a template for how we can use LLM as an Operating System (OS), by just simply returning whether or not an action was completed in \textbf{Subtasks Completed}, and storing the details in \textbf{Shared Variables} as needed.

\newpage
\subsection{Global Context}
\subsubsection{Initialising Global Context}
\begin{figure}[H]
    \centering
    \includegraphics[width=\textwidth]{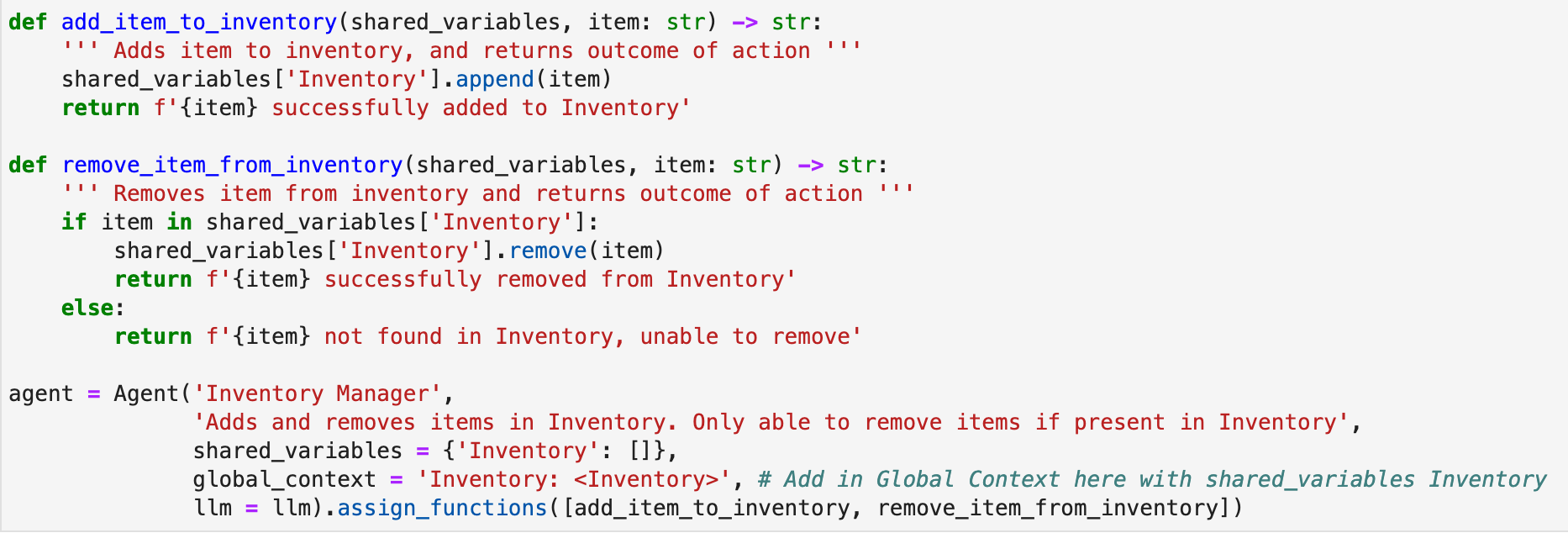}
    \caption{Initialising Global Context}
    \label{fig:taskgen_global_context_init}
\end{figure}

Fig. \ref{fig:taskgen_global_context_init} shows how we can initialise the Global Context by simply initialisng Agent with a \texttt{global\_context} variable. This contains the additional prompt we want to give the Agent, and we express whatever we want to replace with \textbf{Shared Variables} with a <> enclosing the shared variable name. 

Here, in this Inventory Manager Agent, we want to expose the inventory items to the Agent, so we give it the \texttt{global\_context} of "Inventory: <Inventory>", which at runtime, the <Inventory> will be replaced by the actual value in the shared variable "Inventory".

Placing information in \textbf{Global Context} helps the Agent to maintain the most updated picture when the Agent makes its decisions, which is very useful for dynamically changing environments where the Agent would need to continually assess and re-evaluate its situation in the environment.

\newpage
\subsubsection{Running Agent with Global Context}
\begin{figure}[H]
    \centering
    \includegraphics[width=\textwidth]{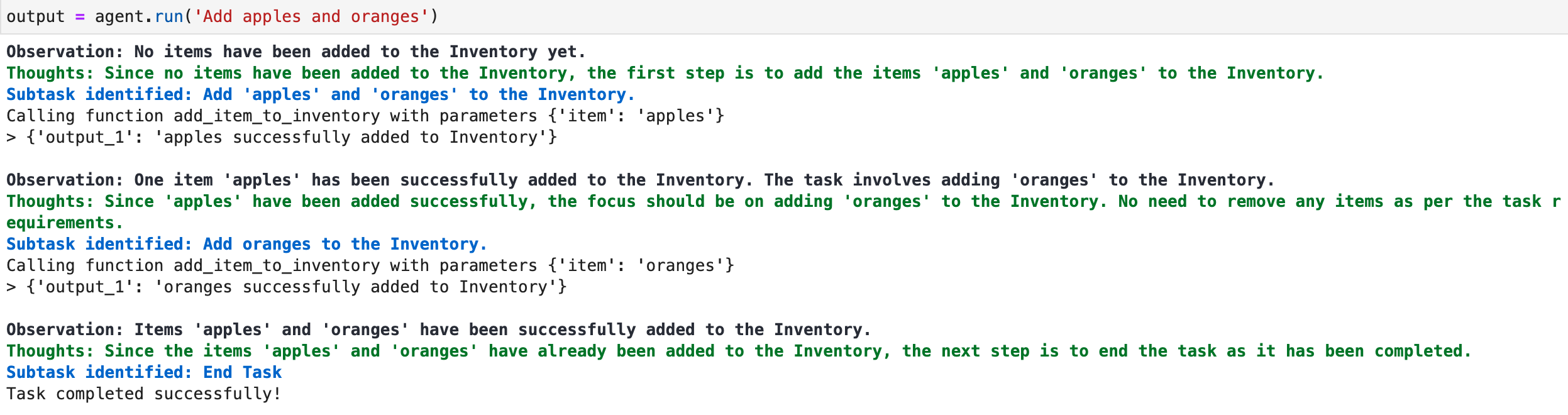}
    \caption{Running Agent with Global Context (Part 1)}
    \label{fig:taskgen_global_context_run_1}
\end{figure}

\begin{figure}[H]
    \centering
    \includegraphics[width=\textwidth]{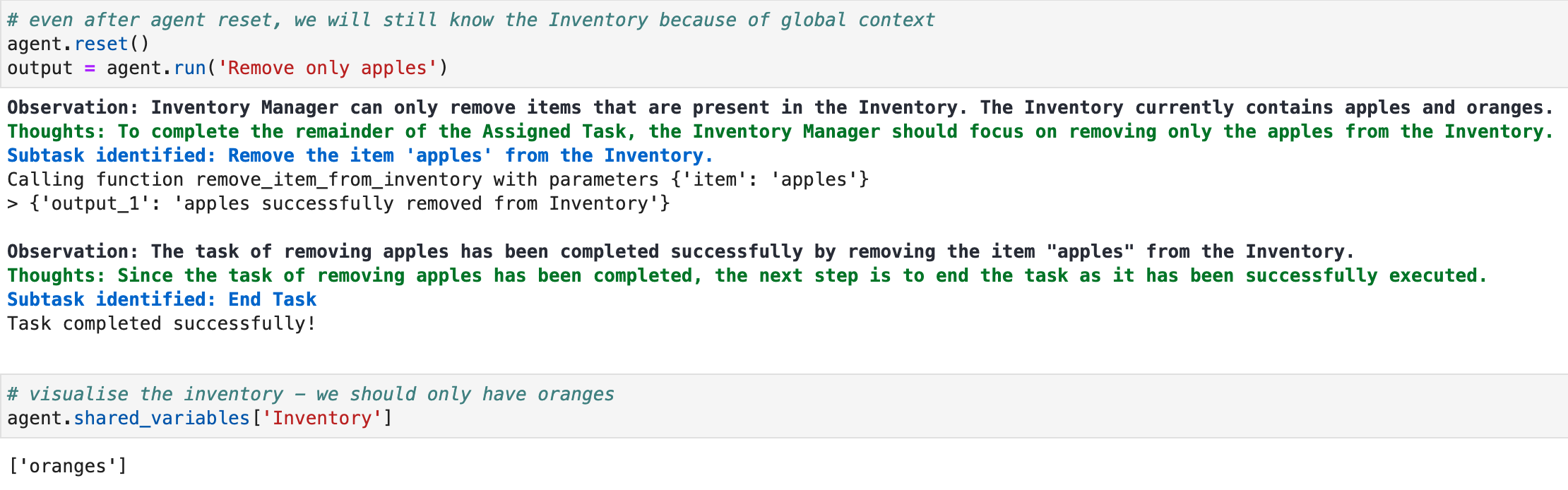}
    \caption{Running Agent with Global Context (Part 2)}
    \label{fig:taskgen_global_context_run_2}
\end{figure}

Figs. \ref{fig:taskgen_global_context_run_1} and \ref{fig:taskgen_global_context_run_2} show how Global Context can be used when running tasks with the Agent. Typically, we do not carry over information across new tasks. However, if we store a persistent state in \textbf{Shared Variables}, such as "Inventory", we can actually expose this "Inventory" variable to the Agent via \textbf{Global Context}.

Hence, as can be seen, after running the task to add apples and oranges, although we reset the Agent and clear its \textbf{Subtasks Completed}, the Agent is still able to know that there are apples and oranges in the inventory and proceed to remove the apples in the next task.

In fact, this practice of continually clearing the Subtasks Completed via \texttt{reset()} and using \textbf{Global Context} to carry over information between tasks is very helpful for Agentic decision making, as the amount of information the Agent needs to focus on is significantly reduced for every future task.

\newpage
\subsection{Memory}

\subsubsection{Initialising Function Memory}
\begin{figure}[H]
    \centering
    \includegraphics[width=\textwidth]{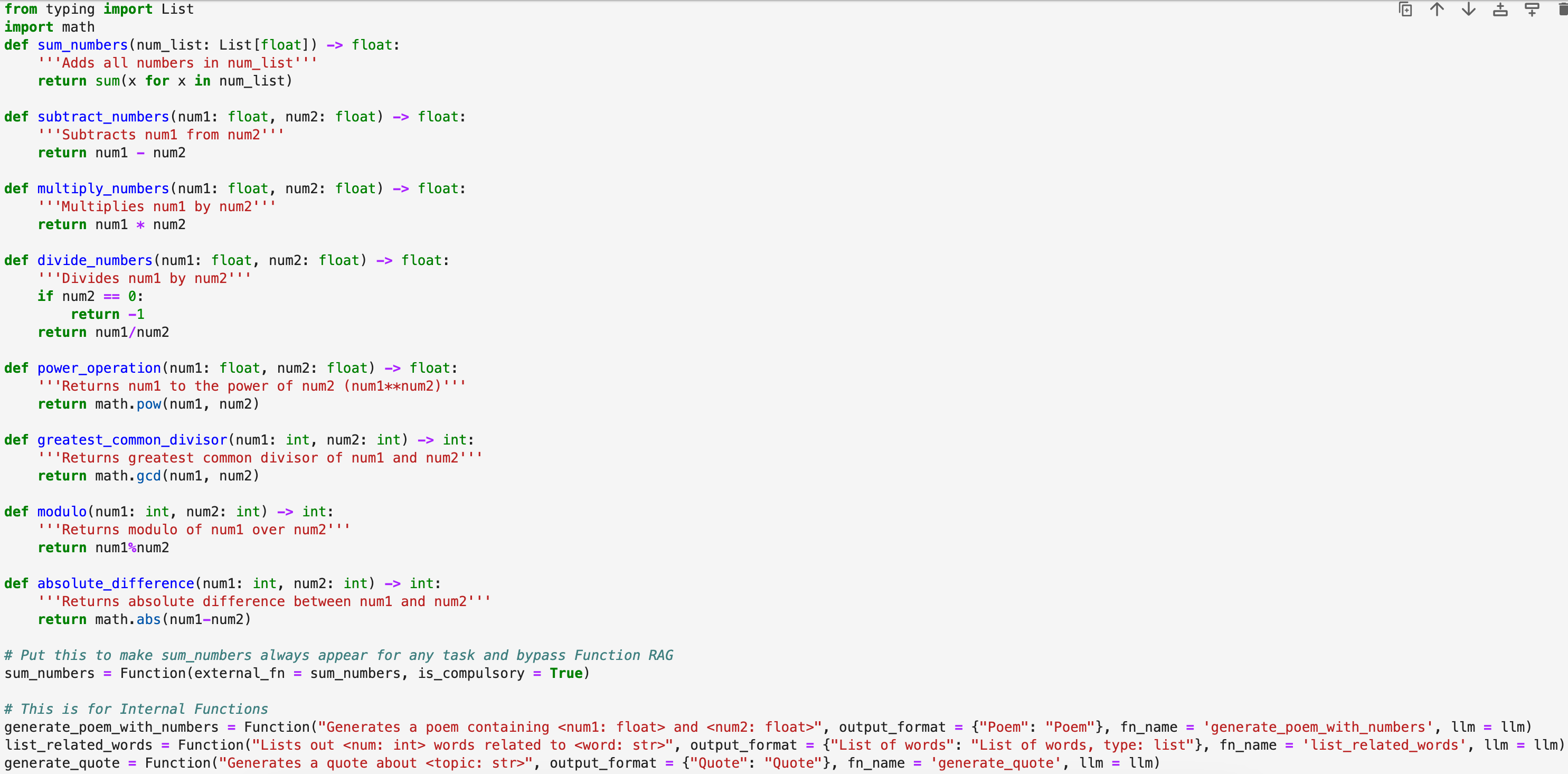}
    \caption{Initialising Functions}
    \label{fig:taskgen_function_memory}
\end{figure}

Fig. \ref{fig:taskgen_function_memory} depicts how we can define External Functions using a normal Python function format with input and output typing and a docstring containing the input variable names, as well as Internal Functions by defining the function description and output format. In order to ensure that certain functions do not go through RAG to filter functions, we can additionally set the \texttt{is\_compulsory} variable to be \texttt{True} when initialising the \texttt{Function} class of TaskGen.

\newpage
\begin{figure}[H]
    \centering
    \includegraphics[width=\textwidth]{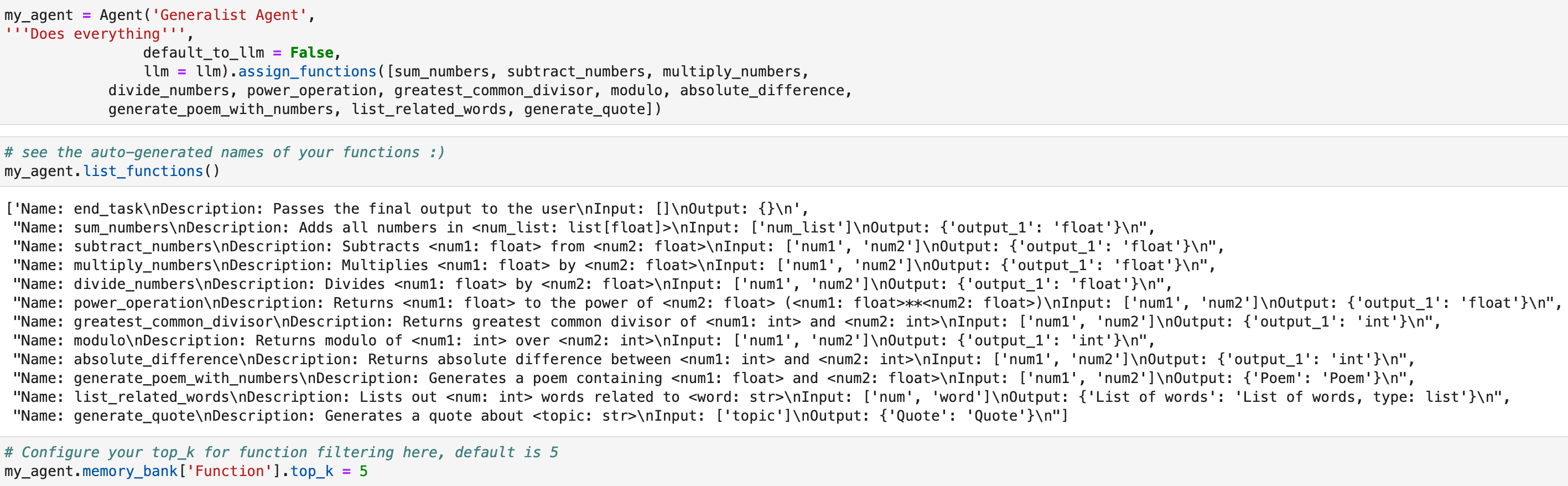}
    \caption{Equipping Agent with Functions}
    \label{fig:taskgen_function_memory_init}
\end{figure}

Fig. \ref{fig:taskgen_function_memory_init} depicts how we can assign the functions using \texttt{assign\_functions} to the LLM. We remove the \texttt{use\_llm} function by setting \texttt{default\_to\_llm} to \texttt{False} in the Agent's initialisation.

We can preview the entire list of functions using \texttt{list\_functions()}. Notice that both the Internal and External Functions are converted to the same format of Name, Description, Input and Output according to the \texttt{Function} class parameters.

Since there are too many Equipped Functions for the Agent to use reliably, TaskGen automatically filters the Equipped Functions (excluding \texttt{use\_llm} and \texttt{end\_task}) down to a \texttt{top\_k} value of 5 based on semantic matching of the function's name and description to the Assigned Task. We can also change this value by modifying the \texttt{top\_k} parameter in the Agent's Memory Bank for \texttt{Function}. There are many other parameters that can be customised, and we encourage the interested reader to check out "Tutorial 3 - Memory" for more details.

\subsubsection{Using Function Memory}
\begin{figure}[H]
    \centering
    \includegraphics[width=\textwidth]{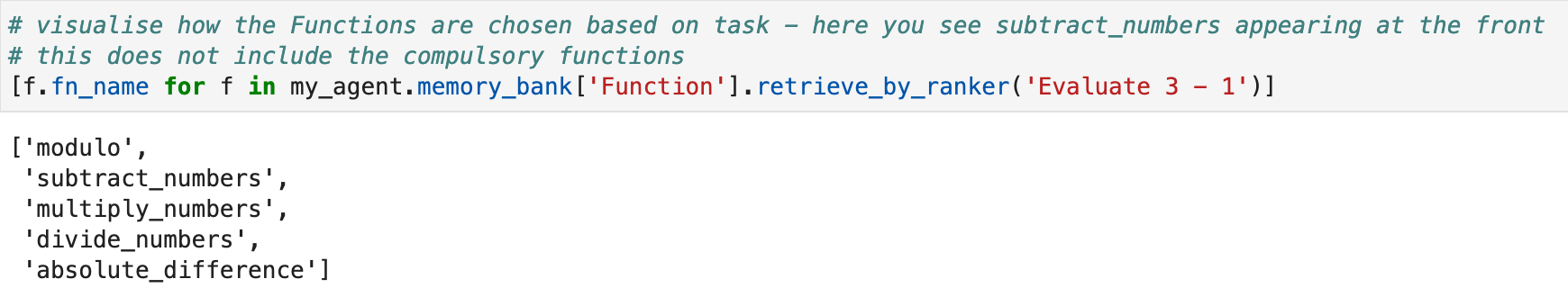}
    \caption{Filtering Functions by Task}
    \label{fig:taskgen_function_memory_filter}
\end{figure}

Fig. \ref{fig:taskgen_function_memory_filter} shows how we can retrieve relevant functions by a ranker (default: OpenAI's "text-embedding-3-small", can be customised to other providers as well). Here, the Assigned Task is to evaluate 3 - 1, and as expected, the Equipped Function \texttt{subtract\_numbers} appear in the list of \texttt{top\_k = 5} functions filtered.

We note here that the \texttt{retrieve\_by\_ranker} function uses cosine similarity to filter the functions according to similarity to the Assigned Task, which may not always be the best approach to do so if the embedding space is not informative of the similarity required. Hence, users are free to customise their own ranker function or to customise the entire \texttt{retrieve\_fn} that takes in a task and outputs the \texttt{top\_k} most similar memories. These changes can be done by simply modifying the "Memory" class accordingly. 

\begin{figure}[H]
    \centering
    \includegraphics[width=\textwidth]{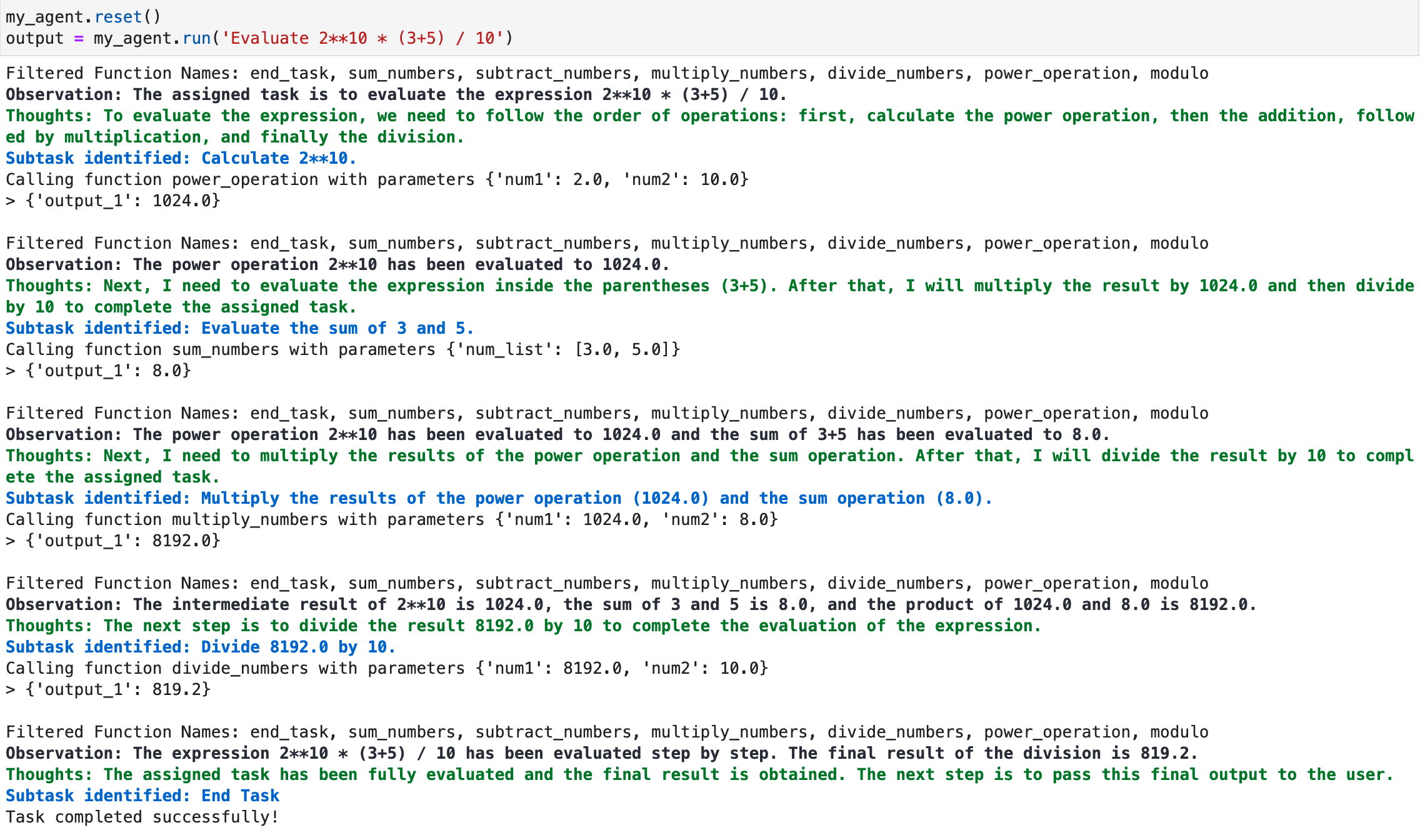}
    \caption{Running Task with Filtered Functions}
    \label{fig:taskgen_function_memory_run}
\end{figure}

Fig. \ref{fig:taskgen_function_memory_run} shows how we can run a task using \texttt{run()}, and the filtering of functions is done automatically at the backend. Do note that the Agent can only use the filtered functions, so if there are functions that are missed out due to failure in retrieving them via RAG, performance may decrease.

\textbf{Current Thoughts by Developer:} The recommended approach for Agents using TaskGen now is actually not to use memory-based filtering of functions, but instead to define each Agent with only a limited set of functions, and to use Inner Agents with a limited set of functions to cover the spectrum of tasks needed if the main Agent has too many functions to use. This reduces the dependency on filtering functions correctly, and ensures quality response by the Agent.

\subsubsection{Storing Additional Task-based Memory in Memory Bank}

\begin{figure}[H]
    \centering
    \includegraphics[width=\textwidth]{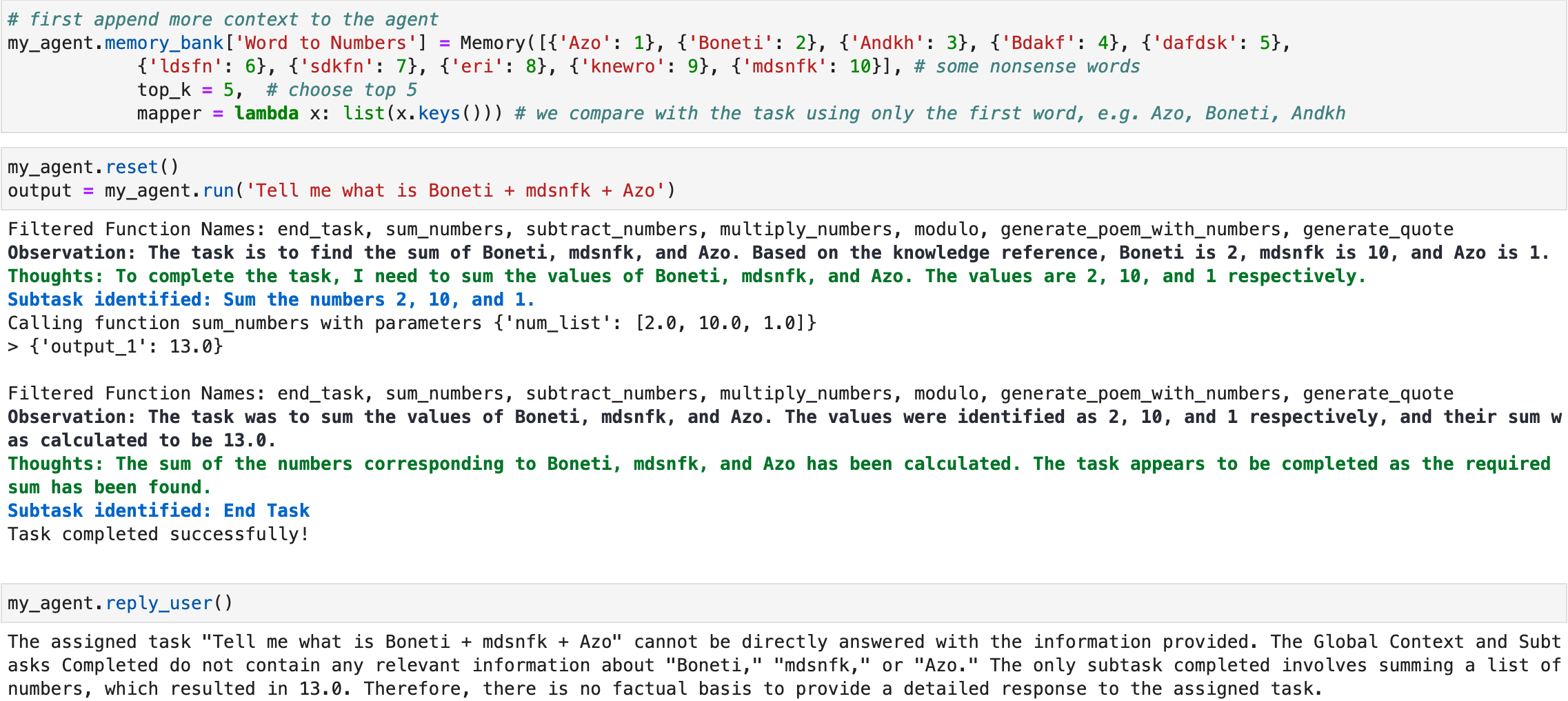}
    \caption{Storing Additional Task-Based Information in Memory Bank}
    \label{fig:taskgen_task_based_memory_1}
\end{figure}

Fig \ref{fig:taskgen_task_based_memory_1} shows how we can incorporate task-based memory in the \textbf{Memory Bank}. We simply define a new key in the \textbf{Memory Bank} Python dictionary. In this case, we define a new key "Word to Numbers" and add in the mapping of various nonsense words to their numerical equivalents. We can also do the same for multiple keys to add in some additional context based on the task. This task-based addition of relevant context is an extremely powerful concept that enables the Agent to work across a wide variety of tasks using the same format. It functions like a general plug-and-play Agent that is infused with specific task-based knowledge based on the Assigned Task.

Here, we can see that by adding in the knowledge of the various nonsense words and their numerical equivalents, the Agent is able to compute a sum such as "Boneti + mdsnfk + Azo".

As the task becomes more complex, storing and using memory of various abstraction spaces will be extremely critical for solving arbitrary tasks.

\begin{figure}[H]
    \centering
    \includegraphics[width=\textwidth]{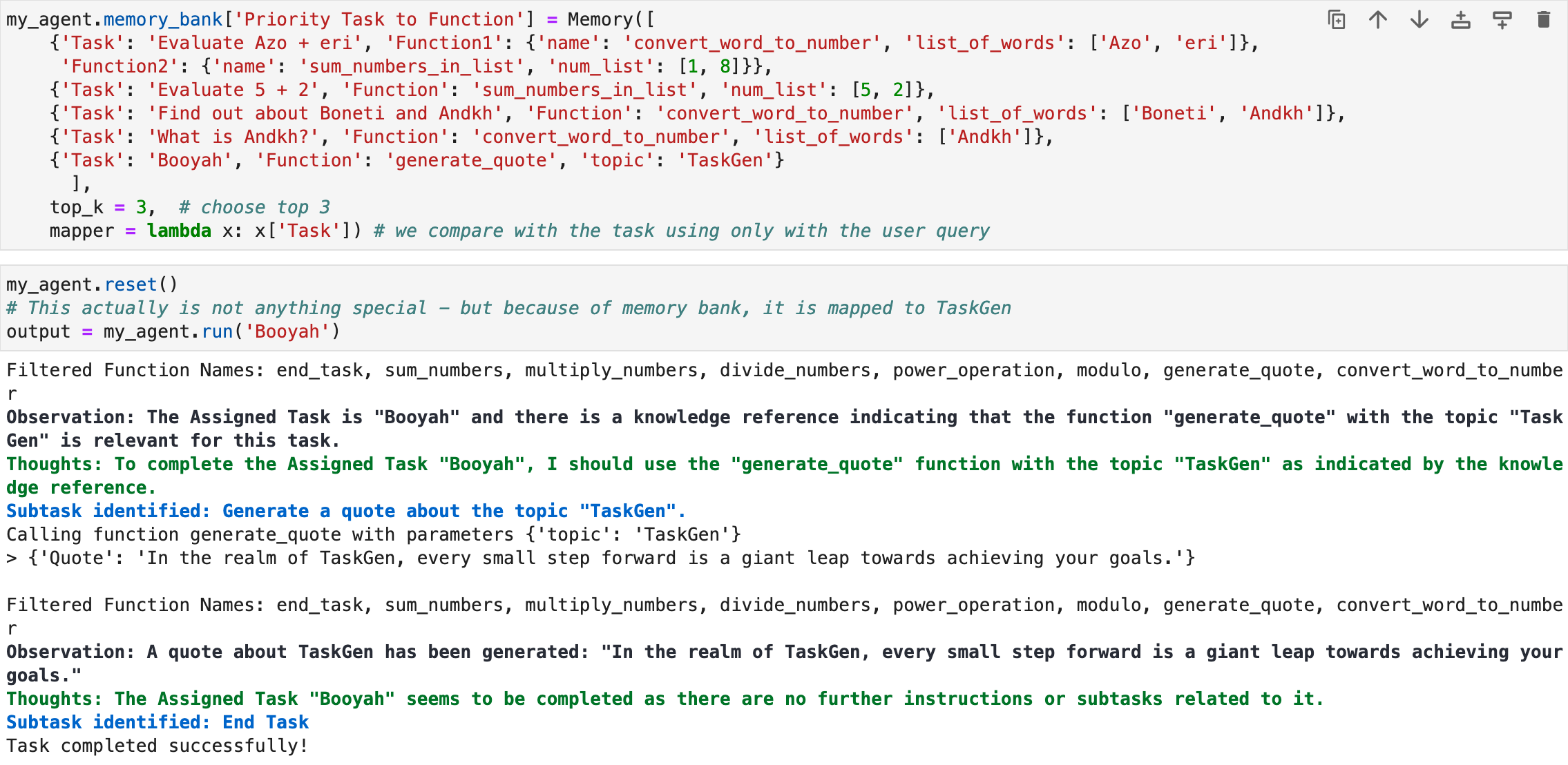}
    \caption{Storing Task to Function mappings in Memory Bank}
    \label{fig:taskgen_task_based_memory_2}
\end{figure}

Fig. \ref{fig:taskgen_task_based_memory_2} shows how we can also use \textbf{Memory Bank} to store various task to function mappings. This memory could be based off the ground truth mappings, or they could be learned on the go by simply storing what worked best during the earlier tasks. Conditioned with such a task to function mapping, an Agent is better able to match future tasks to what has been done effectively in the past.

Here, we can see that by default the task "Booyah" conveys no specific meaning. If we do not have the memory bank of "Priority Task to Function", the Agent will most likely generate a quote about "Booyah". However, when conditioned with the mapping of a task of "Booyah" to the function "generate quote" with topic "TaskGen", we see this being carried out when the Agent is given the task "Booyah".

\textbf{Current Thoughts by Developer:} Having memory in \textbf{Memory Bank} actually biases the Agent greatly to do the previous actions / instructions given in memory. While this may be ideal in cases where the environment does not change, we find that actually storing too much memory in \textbf{Memory Bank} may help to decrease adaptability of the Agent to new scenarios. We are still testing, and are proposing a multi-agent approach to solving new environments. Such a multi-agent approach will contain some agents with past memory, and some agents without any past memory, and we will select the most performant agent in the environment. Such a multi-agent approach will increase robustness and reward either experience if doing actions according to past memory is the best way in the current environment, or exploration if doing something new is the better approach. Increasingly, we come to think of intelligence as not just one single Agent doing tasks, but a group of Agents exploring and exploiting the environment together and learning from one another. This will be a future direction of TaskGen to increase robustness and adaptability for Agents to do well in dynamic environments.

\newpage
\subsection{Conversation Class - Beta Version}

As many applications of LLM involve some form of chatbot or personal assistant, we have decided to create a wrapper class \texttt{ConversableAgent} that takes in an Agent and interfaces it with a conversational interface.

In addition to the shared variables in Agent, \texttt{ConversableAgent} adds on three more:
\begin{enumerate}
    \item \textbf{Persistent Memory.} This stores memory which we want to persist over the entire conversation and it will be updated after each turn of the conversation.
    \item \textbf{Conversation.} This stores the actual conversation itself.
    \item \textbf{Summary of Conversation.} This stores the summary of the entire conversation, which will be used to provide a global context to the Agent.
\end{enumerate}

In general, when given a task, the \texttt{ConversableAgent} firstly performs the actions needed to answer the User's query. The \texttt{ConversableAgent} would then use the summarised actions (if any), \textbf{Global Context}, Summary of Past Conversation, Past Conversation, Persistent Memory to reply the User. The \texttt{ConversableAgent} will also update the Summary of Conversation.

After the reply to the User, \texttt{ConversableAgent} will append the User's message and the Agent's reply to Conversation, and update the Persistent Memory accordingly.

Overall, the main goal is to imbue a conversation with persistent states such as Persistent Memory and Summary of Conversation, so as to be able to create more wholesome and natural conversation.

\textbf{Insights by Developer:} Conversation is not the main means of solving the User's query, so as to make the task solving portion concise. The task is solved first, before the 
Agent is given the chance to reply the User. In earlier iterations of \texttt{ConversableAgent}, when we had given the LLM function directly to the Agent, it is quite likely that the Agent will use the LLM function to hallucinate an outcome for the task that has never happened. This is an interesting finding that the task executor and the response to User portion of \texttt{ConversableAgent} should be implemented separately to minimise hallucinations.

\newpage
\begin{figure}[H]
    \centering
    \includegraphics[width=\textwidth]{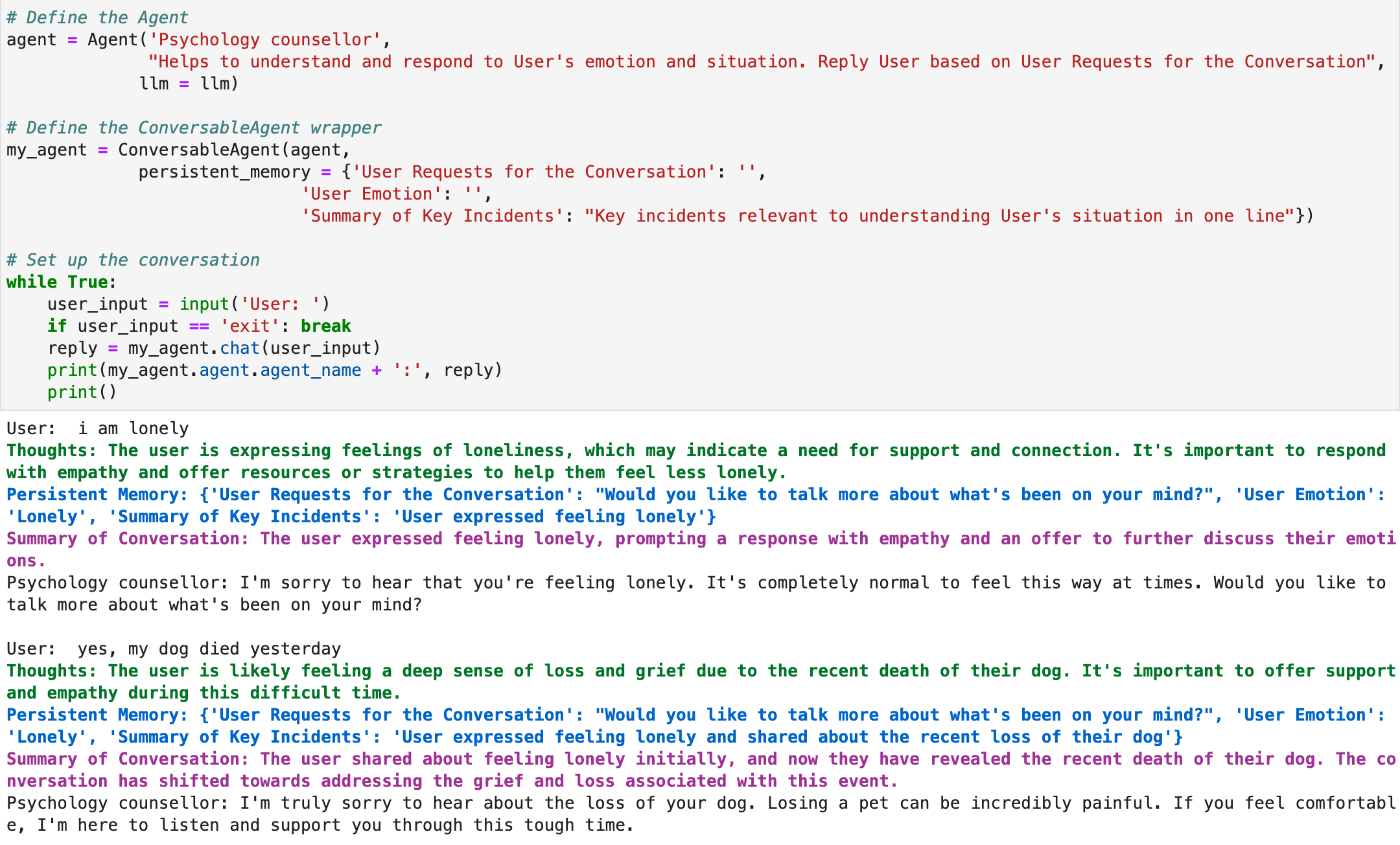}
    \caption{Running ConversableAgent without additional Equipped Functions}
    \label{fig:taskgen_conversableagent_run}
\end{figure}

Fig. \ref{fig:taskgen_conversableagent_run} shows how we can implement a Psychology Counsellor Agent by wrapping the baseline agent in a \texttt{ConversableAgent} class, and giving it Persistent Memory of User Request for the Conversation, User Emotion, Summary of Key Incidents. Note that the \texttt{persistent\_memory} variable takes the same form as the \texttt{output\_format} of the \texttt{strict\_json} function.

\newpage
\begin{figure}[H]
    \centering
    \includegraphics[width=\textwidth]{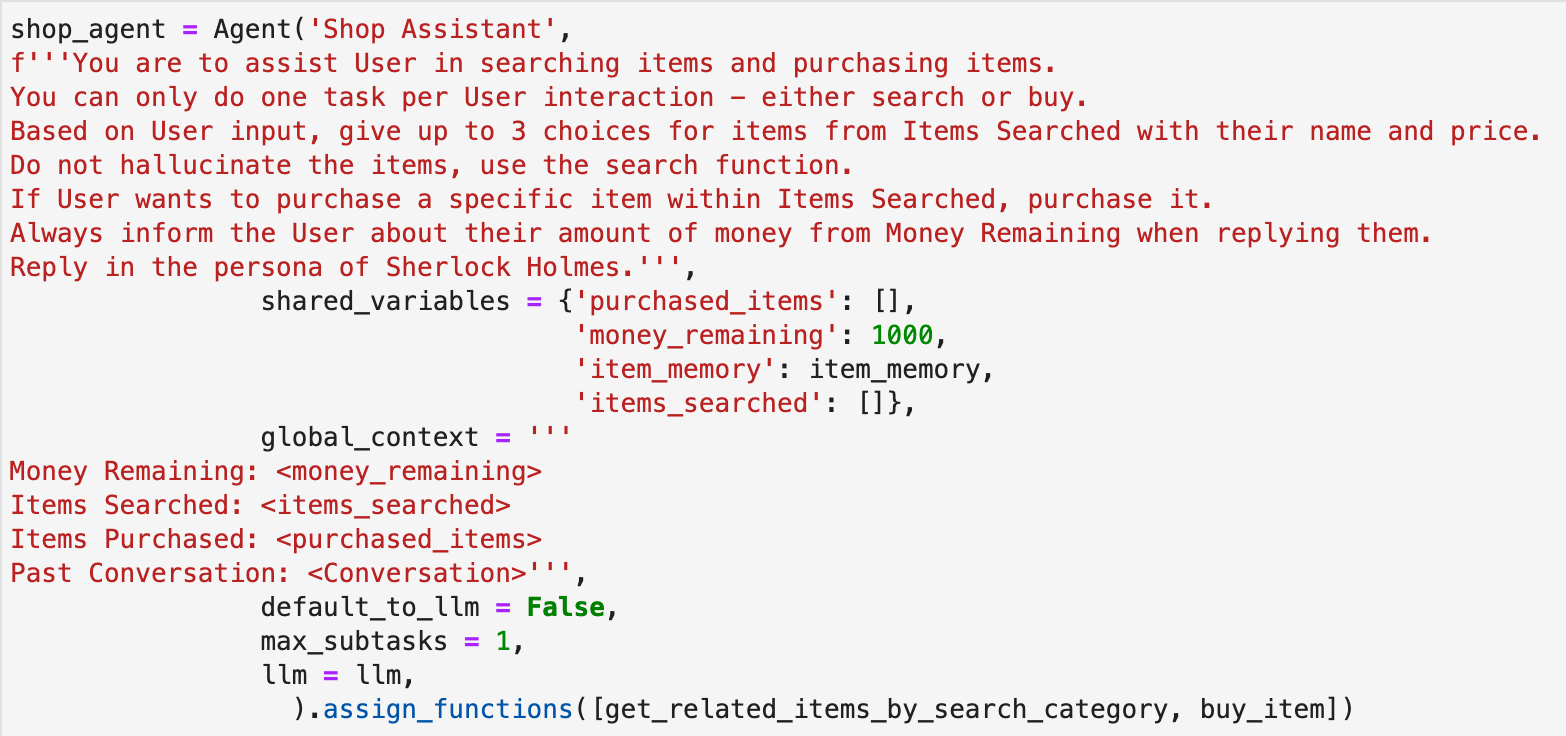}
    \caption{Initialising ConversableAgent with Equipped Functions}
    \label{fig:taskgen_conversableagent_action_init}
\end{figure}

\begin{figure}[H]
    \centering
    \includegraphics[width=\textwidth]{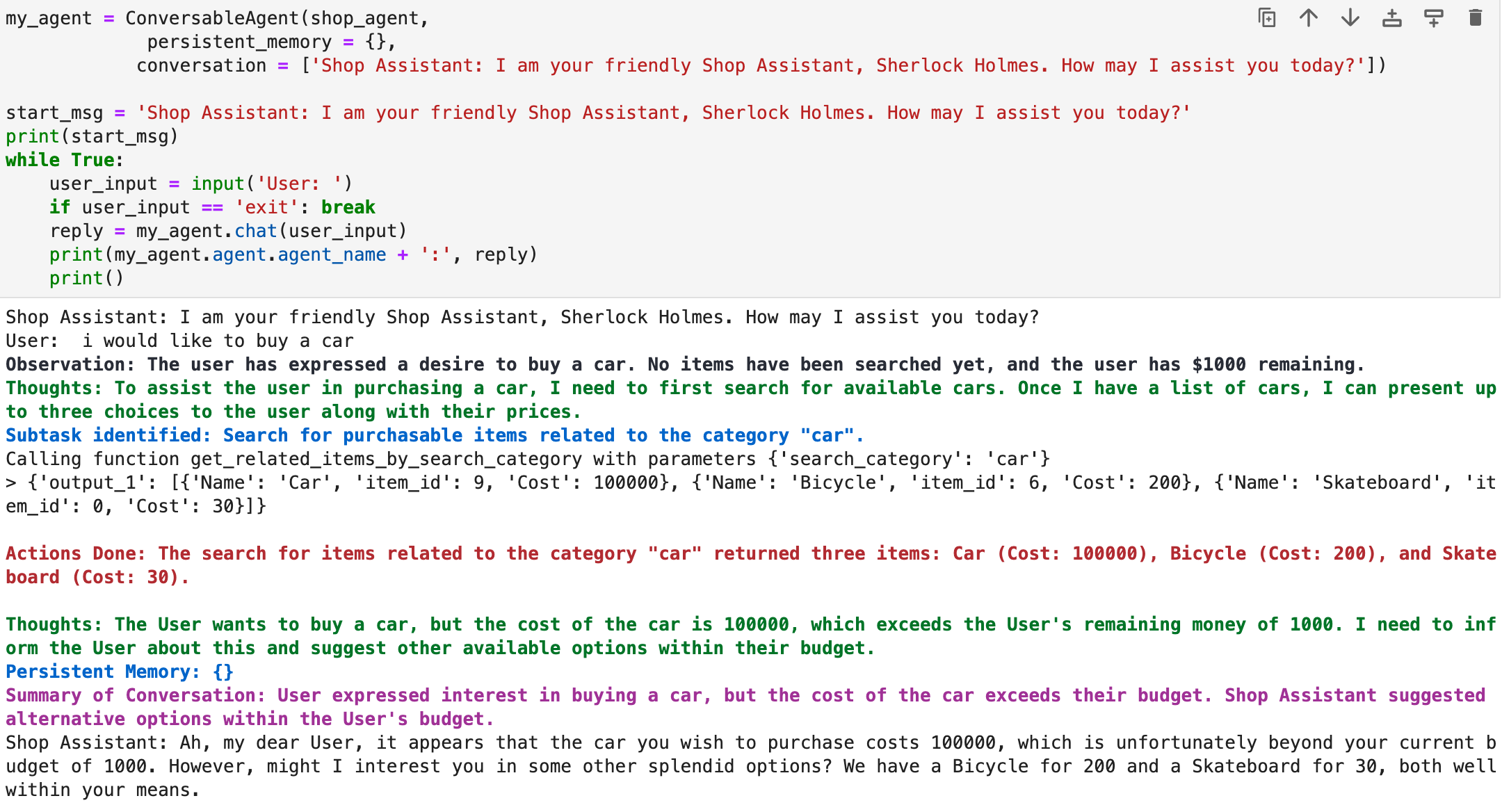}
    \caption{Running ConversableAgent with additional Equipped Functions}
    \label{fig:taskgen_conversableagent_action_run}
\end{figure}

Figs. \ref{fig:taskgen_conversableagent_action_init} and \ref{fig:taskgen_conversableagent_action_run} shows how to initialise and run a Shop Assistant that can search and purchase items for the User and responds in the persona of Sherlock Holmes. The Shop 
Assistant is given Global Context (and Shared Variables) of Money Remaining, Items Searched, Items Purchased and Past Conversation.

When replying the User, the relevant functions are firstly called in response to the User's message, and the Shop Assistant Agent then references what has been done in the action summary (red text titled Actions Done) to inform the User accordingly.

\newpage
\section{Community Contributions to TaskGen}
\renewcommand{\thefigure}{C\arabic{figure}}
\renewcommand{\thetable}{C\arabic{table}}
\setcounter{figure}{0}
\setcounter{table}{0}
\label{appendix: community contribs}

This section elucidates the methodology by which users of TaskGen can contribute to the library, thereby fostering the growth of the TaskGen community.

\subsection{Motivation for Community Contribution}

TaskGen, an open-source repository, actively encourages contributions from its user base to enhance the library's functionality and accessibility. As users of TaskGen, individuals are incentivised to develop sophisticated agents utilising the framework and subsequently contribute these agents for the benefit of the broader community. This approach aligns with the ethos of open-source development and aims to cultivate a collaborative ecosystem where users can build upon each other's contributions. The overarching vision is to establish a marketplace of powerful agents leveraging the TaskGen framework, ultimately increasing the repository's utility through reusability.

\begin{figure}[H]
\centering
\includegraphics[width=0.5\linewidth]{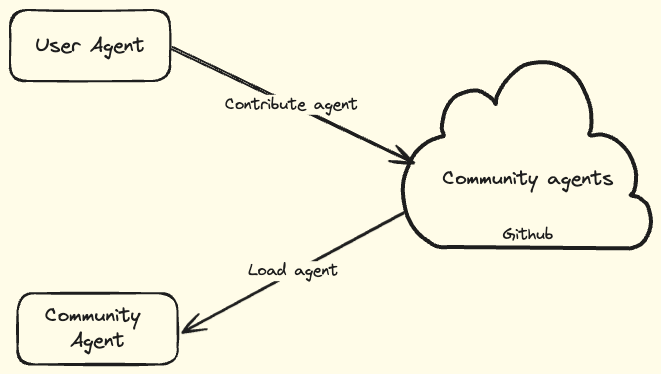}
\caption{Community contributions to TaskGen}
\label{fig:TaskgenContributionFlow}
\end{figure}

\subsection{Key Features of the Contribution Process}

To facilitate seamless community involvement, significant efforts have been invested in streamlining the contribution process. Notable features include:

\begin{enumerate}
    \item \textbf{Simplified Contribution:} Users can contribute their agents through a single function invocation.
    \item \textbf{Minimal Prerequisites:} The process requires only a configured GitHub profile, eliminating the need for local git setup or repository cloning.
    \item \textbf{Comprehensive Support:} The contribution mechanism accommodates various configurations, including max\_subtasks, summarise\_subtasks\_count, memory\_bank, shared\_variables, global context settings, sub\_agents, and both internal and external functions.
    \item \textbf{Efficient Integration:} Accepted contributions can be loaded as agents with a single line of Python code.
\end{enumerate}

\subsection{Technical Implementation}

The contribution process involves the following steps:

\begin{enumerate}
    \item \textbf{Environment Configuration:} Users must set the GITHUB\_USERNAME and GITHUB\_TOKEN environment variables.
    \item \textbf{Agent Contribution:} Invocation of the \texttt{contribute\_agent} function on the user's agent.
\end{enumerate}

To load a contributed agent, users can utilise the \texttt{load\_community\_agent} class method from the Agent class, specifying the agent name.\footnote{We recommend you pull the latest version of taskgen to get the most recent community agents.}

The backend process of the \texttt{contribute\_agent} function encompasses:

\begin{enumerate}
    \item Creation of a TaskGen fork for the user (if not already existing).
    \item Generation of a Python representation of the user's Agent, including subclasses for the agent and sub-agents, along with external functions and configurations.
    \item Utilization of low-level GitHub APIs to commit the agent's Python representation to a branch in the user's fork.
    \item Initiation of a Pull Request to the main TaskGen repository.
\end{enumerate}

\subsection{Examples}

To illustrate the contribution and usage process, we provide the following examples:

\subsubsection{Contributing an Agent}

The following code snippet demonstrates how to create and contribute an agent:

\begin{lstlisting}[style=Python]
from taskgen import *

# Create your agent by specifying name and description
my_agent = Agent('Helpful assistant', 'You are a generalist agent')

# Example External Function
def binary_to_decimal(x: int) -> int:
    ''' Convert input <x: a binary number in base 2> to base 10 '''
    return int(str(x), 2)

# Example Internal Function
sentence_style = Function(fn_description = 'Output a sentence with <obj> and <entity> in the style of <emotion>', 
                     output_format = {'output': 'sentence'}, fn_name = 'sentence_with_objects_entities_emotion')

# Assign functions
my_agent.assign_functions(function_list = [binary_to_decimal, sentence_style])

# Contribute your agent
os.environ['GITHUB_USERNAME'] = '<your GitHub username>'
os.environ['GITHUB_TOKEN'] = '<your GitHub token>'
my_agent.contribute_agent(author_comments = 'This is a generalist agent')
\end{lstlisting}

\subsubsection{Loading a Community Agent}

To load a contributed agent, users can employ the following simple code:

\begin{lstlisting}[style=Python]
from taskgen import *
agent = Agent.load_community_agent("Helpful Assistant")
\end{lstlisting}

\newpage

\subsubsection{Generated Code}

The contribution process generates a Python representation of the agent. Below is an example of the generated code:

\begin{lstlisting}[style=Python]
from taskgen import Agent, Function, Memory, Ranker
import math


# Author: @name_of_author
# Author Comments: This is a generalist agent
class HelpfulAssistant_abc(Agent):
    def __init__(self):
        var_binary_to_decimal = Function(
            fn_name="binary_to_decimal",
            fn_description=''' Convert input <<x: int>: a binary number in base 2> to base 10 ''',
            output_format={'output_1': 'int'},
            examples=None,
            external_fn=binary_to_decimal,
            is_compulsory=False)

        var_sentence_with_objects_entities_emotion = Function(
            fn_name="sentence_with_objects_entities_emotion",
            fn_description='''Output a sentence with <obj> and <entity> in the style of <emotion>''',
            output_format={'output': 'sentence'},
            examples=None,
            external_fn=None,
            is_compulsory=False)

        super().__init__(
            agent_name="Helpful assistant",
            agent_description='''You are a generalist agent''',
            max_subtasks=5,
            summarise_subtasks_count=5,
            memory_bank={'Function': Memory(memory=[], top_k=5, mapper=lambda x: x.fn_name + ': ' + x.fn_description, approach='retrieve_by_ranker', ranker=Ranker(model='text-embedding-3-small', ranking_fn=None)),},
            shared_variables={},
            get_global_context=None,
            global_context='''''',
            default_to_llm=True,
            code_action=False,
            verbose=True,
            debug=False
        )

        self.assign_functions(
            [var_binary_to_decimal,var_sentence_with_objects_entities_emotion]
        )

        self.assign_agents(
            []
        )

# Supporting Functions
def binary_to_decimal(x: int) -> int:
    ''' Convert input <x: a binary number in base 2> to base 10 '''
    return int(str(x), 2)
\end{lstlisting}

These examples demonstrate the simplicity of contributing and loading agents, as well as the structure of the generated code that encapsulates the agent's functionality.

\subsection{Future Work and Community Feedback}

While efforts have been made to support diverse agent configurations, it is acknowledged that there may be limitations in the current contribution process. Users are encouraged to provide feedback by raising issues on the GitHub repository to continually improve this process.


\newpage
\section{Dynamic Maze Navigation}
\renewcommand{\thefigure}{D\arabic{figure}}
\renewcommand{\thetable}{D\arabic{table}}
\setcounter{figure}{0}
\setcounter{table}{0}
\label{appendix: maze}
\subsection{Maze Navigation}
We evaluate TaskGen with a StrictJSON Planner, \textbf{Shared Variables} and \textbf{Global Context} in a dynamic maze navigation environment. It manages to solve the hardest 40x40 dynamic grid world all the time, faring better than prior methods in Learning, Fast and Slow \cite{tan2023learning}.

\subsubsection{Background}

\begin{figure}[H]
\begin{center}
\includegraphics[width = \textwidth]{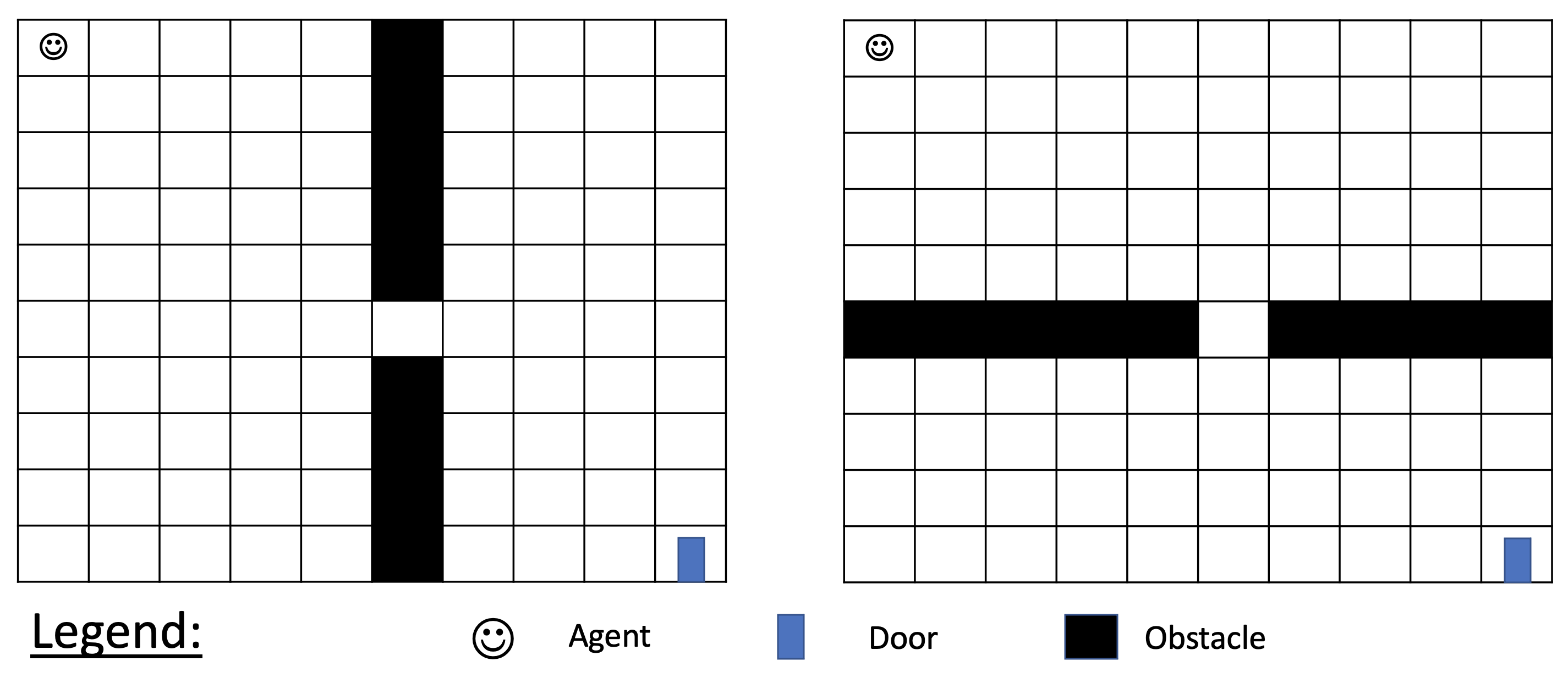}
\end{center}
\vspace{-5mm}
\caption{A sample maze environment of size 10x10. The actual experiment is 40x40. By default, the agent's start state is at the top left and the door is at the bottom right, but it can be varied. Obstacles change after half of the total number of episodes. \textbf{(Left)} Obstacles in the first half form a vertical wall with a gap in the centre across the mid-point. \textbf{(Right)} Obstacles in the second half from a horizontal wall with a gap in the centre across the mid-point.}
\label{fig: maze}
\vspace{-3mm}
\end{figure}

\subsubsection{Experimental Setup}
The environment used is a 2D grid world, where there are 40 by 40 squares. There are also some grid squares which are denoted as obstacles and are not traversable. The agent starts off at a grid square and is supposed to head towards the door (exit) position.

The obstacles change mid-way, and the start and end points vary randomly with each episode. This is a difficult environment to evaluate learning as it is continuously changing. See Fig. \ref{fig: maze} for an illustration. 

\textbf{State Space.} The agent is provided with both its own position and the door (exit) position.

\textbf{Reward.} This is a sparse reward environment and the agent will only be counted as completing the episode and receive a reward of 1 if it manages to reach the door before $40\times 40$ time steps. Otherwise, it will receive a reward of 0.

\textbf{Action Space.} The available action space is discrete from the set \{Up, Down, Left, Right\}. There is no wraparound, and the agent will remain in its existing position should it collide with the edges of the grid or with an obstacle. 

\textbf{Agents.} We use a TaskGen Agent using "gpt-4o" as the LLM. We pit its performance against Fast \& Slow (F\&S) and three RL-based agents - Proximal Policy Optimisation (PPO) \cite{schulman2017proximal}, Trust Region Policy Optimisation (TRPO) \cite{schulman2015trust} and Advantage Actor-Critic (A2C) \cite{mnih2016asynchronous}.

\newpage
\subsubsection{TaskGen Agent}

\begin{figure}[H]
    \centering
    \includegraphics[width=\textwidth]{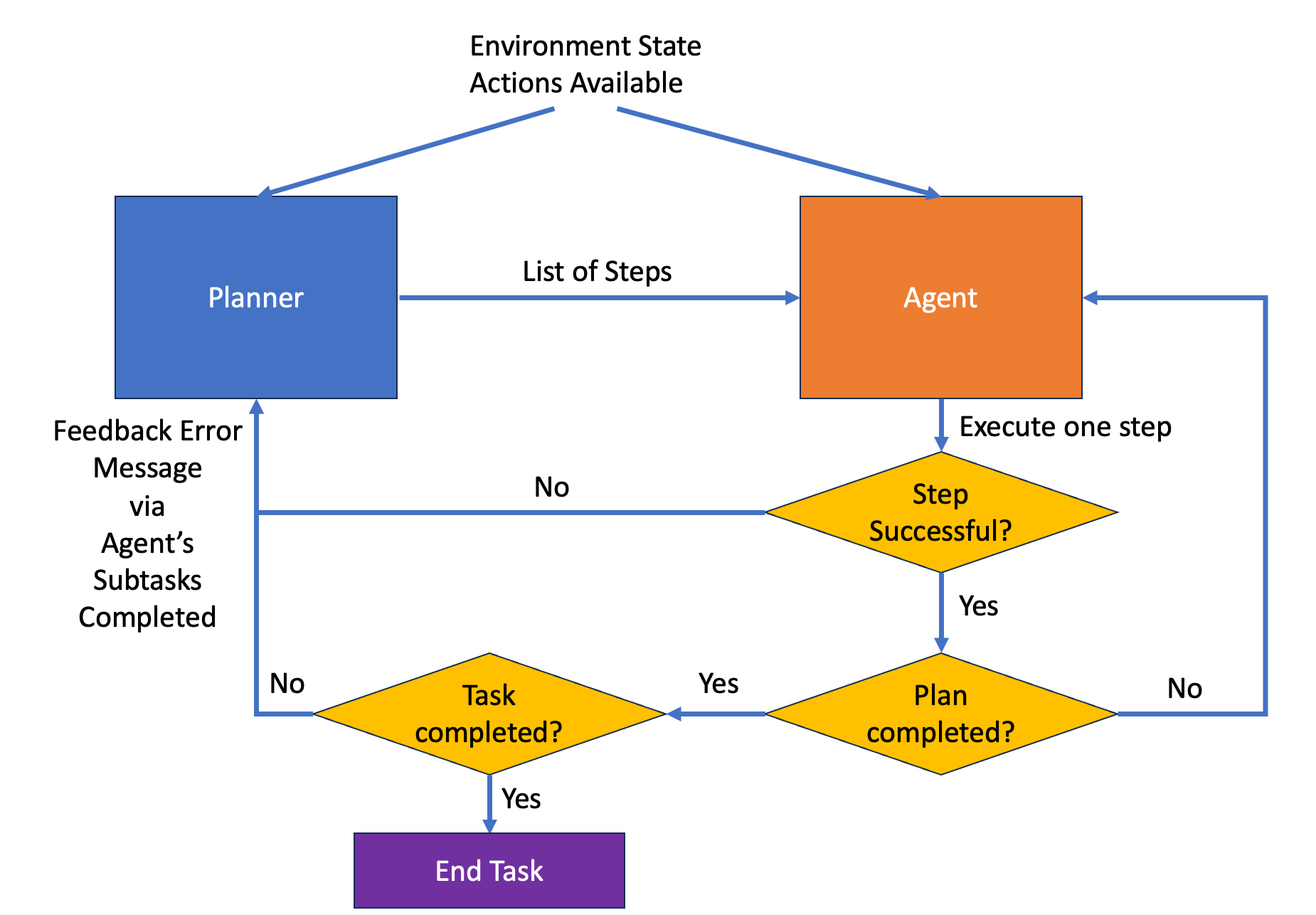}
    \caption{Schema of TaskGen Planner Interface with TaskGen Agent}
    \label{fig:taskgen_planner}
\end{figure}

\textbf{Introduction.} As the environment is huge and difficult to navigate just by exploring and thinking on-the-go, we use a Planner to craft an overall plan. This is actually a alpha-version of the Planner for TaskGen which is not released officially yet. The reason why we need Planner instead of just simply a Meta Agent is because we want to use rule-based methods to ensure each part of the plan is followed for continuity. 

\textbf{Overall Schema.} Both the Planner and the Agent will have access to the environmental states via \textbf{Global Context}, and actions available. The difference in roles is that the Planner is in charge of the bigger picture from end-to-end and is in charge of deriving a list of steps to get the Agent from the current state to the goal state. The Agent will focus on the more immediate situation, and will seek to execute the most immediate step from the list of steps that the Planner has planned out. In order to ensure continuity of plan, we follow the Planner's plan all the way until a step failed, which will then cue replanning. In the dynamic maze environment, this is if an obstacle is encountered or if the Agent reaches out of bounds. After the entire plan is executed, we will check for task completion before breaking out of the loop. If the task is completed (i.e. agent is at the exit position), we end the task. Otherwise, we will get the Planner to replan again. This is shown in Fig. \ref{fig:taskgen_planner}.

\textbf{Planner.} For the Planner, we use a \texttt{strict\_json} function with the inputs of Start Position, Exit Position, Obstacle Locations and Subtasks Completed. The Planner will use CoT prompting to get a plan from current position to exit position. A sample CoT generation for the plan is as follows:
\begin{enumerate}[leftmargin=*, nosep, wide=0pt]
\item Example Start Position: (2, 0)
\item Example Exit Position: (2, 4)
\item Example Obstacle Positions: ["Obstacle from (0, 1) to (5, 1)"]
\item Example Obstacle Position Layout: There is a wall of obstacles from (0, 1) to (5, 1)
\item Example Thoughts: I need to get from (2, 0) to (0, 4)
There are obstacles in the way. Since (2, 1) to (5, 1) has obstacles, I am only able to go past the wall via (6, 1)
\item Example Plan: ["Move down 4 times from (2, 0) to (6, 0)", 
"Move right 4 times from (6, 0) to (6, 4)",
"Move up 4 times from (6, 4) to (2, 2)"]
\end{enumerate}

\textbf{Agent.} The Agent is equipped with a \texttt{move} function that takes in an action and the number of times to execute it. We first reset the \textbf{Subtasks Completed} of the Agent before running the task, to prevent past history from affecting the current task. The task is the most immediate step of the Plan. We also provide the Agent with its current position and exit position in \textbf{Global Context}. As the Agent traverses the environment, we also update the Obstacle Locations encountered. If the obstacle is not present, it will be removed from memory. If the obstacle is present but not in memory, it will be added to memory. If there is no error in execution of the task, we proceed to the next item of the Plan. Otherwise, we will get the Planner to re-plan.

\textbf{Differences from prior methods.} As LLM-based methods require more semantic understanding of the world to work, we give the TaskGen agent the full specifications of the environment description and meanings of each action. Furthermore, to facilitate faster execution, we allow the TaskGen Agent to execute the same action multiple times at a go. This is possible as the LLM is able to express arbitrary output which prior methods struggle with. We are also able to externally store the observed obstacle positions, and input these positions as in-context prompt to the TaskGen Agent. The obstacle positions are grouped continuously before being fed to the LLM, like (0, 0), (0, 1) and (0, 2) will get grouped as obstacles from (0, 0) to (0, 2). This is because the LLM is not very good at exact grid positions, and abstractions like these help with understanding a wall of obstacles better. Another significant difference is that in order to reduce the number of turns, instead of letting the Agent bump into an obstacle to discover its presence, we give the TaskGen Agent a 3x3 square view of vision centered on itself to discover all nearby obstacles. This is also more realistic as in real life an agent should have some vision to see what is in the world.

\newpage
\subsubsection{Overall Results}



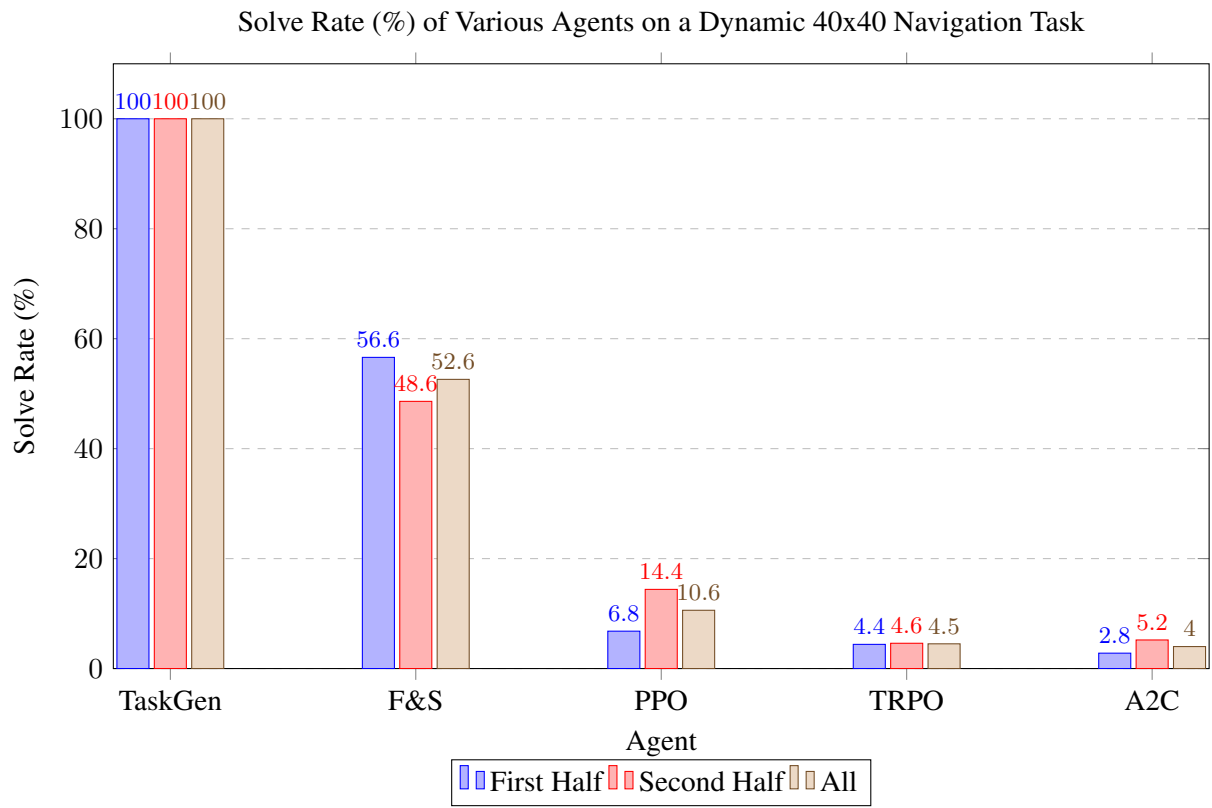
\begin{figure}[H]
    \centering
    \begin{tikzpicture}
        \begin{axis}[
            ybar,
            bar width=12pt,
            title={Solve Rate (\%) of Various Agents on a Dynamic 40x40 Navigation Task},
            xlabel={Agent},
            ylabel={Solve Rate (\%)},
            symbolic x coords={TaskGen, F\&S, PPO, TRPO, A2C},
            xtick=data,
            nodes near coords,
            nodes near coords align={vertical},
            nodes near coords style={font=\small},
            legend style={at={(0.5,-0.15)}, anchor=north, legend columns=-1},
            enlarge x limits={abs=0.75cm},
            ymin=0,
            ymax=110,
            ymajorgrids=true,
            grid style=dashed,
            width=\textwidth,
            height=0.6\textwidth,
        ]

        \addplot coordinates {(TaskGen, 100) (F\&S, 56.6) (PPO, 6.8) (TRPO, 4.4) (A2C, 2.8)};
        \addlegendentry{First Half}

        \addplot coordinates {(TaskGen, 100) (F\&S, 48.6) (PPO, 14.4) (TRPO, 4.6) (A2C, 5.2)};
        \addlegendentry{Second Half}

        \addplot coordinates {(TaskGen, 100) (F\&S, 52.6) (PPO, 10.6) (TRPO, 4.5) (A2C, 4.0)};
        \addlegendentry{All}

        \end{axis}
    \end{tikzpicture}
    \caption{Solve rate (\%) of various agents on a dynamic 40x40 navigation task for the first half, second half of episodes after obstacle positions change, and across all episodes.}
    \label{fig:solve_rate_40x40}
\end{figure}

For the RL agents and F\&S, we show the results across 100 episodes averaged across 10 seeds, as per the original paper. For the TaskGen agent, we evaluate it with few-shot prompting without any training across 20 episodes with environment changeover after 10 episodes across 1 seed.

Overall, Fig. \ref{fig:solve_rate_40x40} shows that TaskGen performs the best compared to all other agents such as F\&S, PPO, TRPO, A2C. This is significant, as it shows that for complex environments, perhaps having a Planner is critical for success and continuity of actions to achieve a long-term goal. It also shows the versatility of TaskGen to be reconfigured for an agentic task such as navigation.

\newpage
\subsubsection{Detailed Run-through of TaskGen Agent}



\begin{figure}[H]
    \centering
    \begin{tikzpicture}
        \begin{axis}[
            title={Comparison of Actual Steps taken vs Minimum Steps possible for each episode for TaskGen Agent},
            xlabel={Episode Number},
            ylabel={Steps Taken},
            xmin=1, xmax=20,
            ymin=0, ymax=250,
            xtick={1,2,3,4,5,6,7,8,9,10,11,12,13,14,15,16,17,18,19,20},
            ytick={0,50,100,150,200,250},
            legend pos=north east,
            ymajorgrids=true,
            grid style=dashed,
            width=\textwidth,
            height=0.5\textwidth,
        ]

        \addplot[
            color=blue,
            mark=square,
            ]
            coordinates {
            (1,62)(2,43)(3,36)(4,34)(5,46)(6,56)(7,47)(8,42)(9,44)(10,38)
            (11,63)(12,36)(13,41)(14,47)(15,28)(16,46)(17,44)(18,19)(19,35)(20,30)
            };
        \addlegendentry{Minimum Steps}

        \addplot[
            color=red,
            mark=*,
            ]
            coordinates {
            (1,210)(2,46)(3,58)(4,34)(5,46)(6,182)(7,49)(8,42)(9,44)(10,38)
            (11,68)(12,62)(13,52)(14,48)(15,28)(16,75)(17,47)(18,21)(19,57)(20,34)
            };
        \addlegendentry{Actual Steps}

        \draw[dashed, thick] (axis cs:10.5,0) -- (axis cs:10.5,250);
        \node[anchor=west] at (axis cs:10.5,125) {Obstacle Change};

        \end{axis}
    \end{tikzpicture}
    \caption{Comparison of Actual Steps taken vs Minimum Steps possible for each episode for TaskGen Agent. Note the obstacle positions are changed after episode 10.}
    \label{fig:steps_per_episode}
\end{figure}
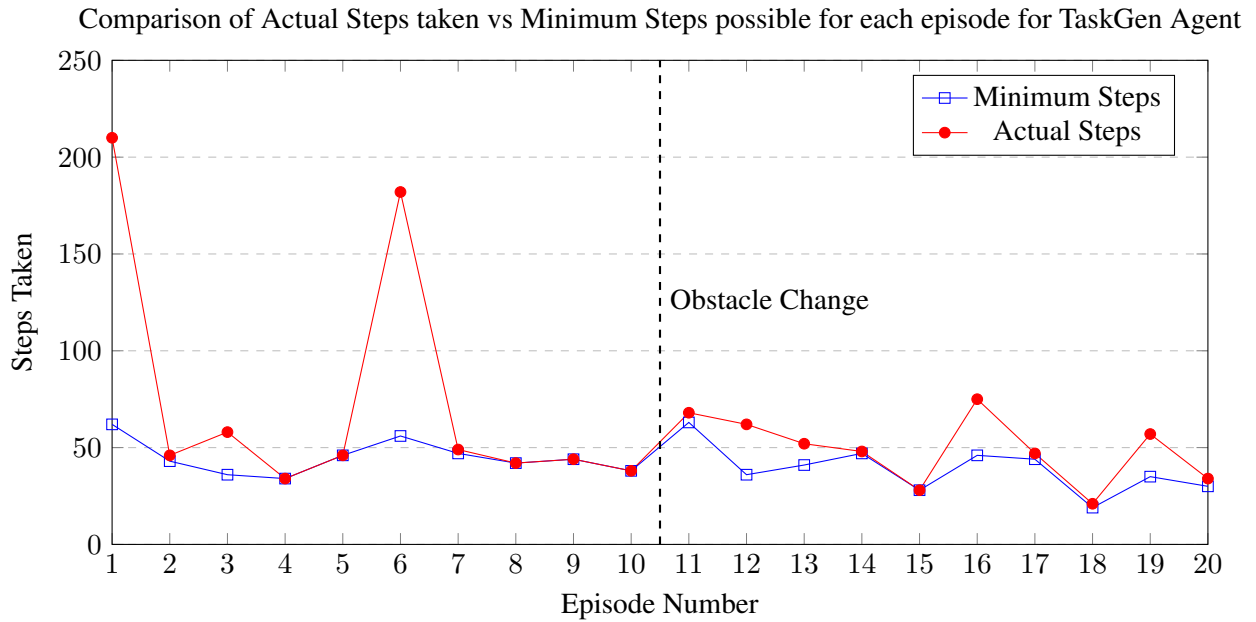

Fig. \ref{fig:steps_per_episode} shows that in general, TaskGen Agent is able to solve the episodes quite efficiently with planning. The main cause of the higher actual steps for some environments is when some obstacle positions are unknown, the Planner is not perfect and sometimes chooses a position to backtrack to that does not work out. When the obstacles in the path are known, the Planner can usually generate a perfect plan, or can correct itself quickly mid-way.

In general, as the number of episodes increase in the same environment, the better the knowledge of the obstacles, and hence the better the plan and the generated actions. Even if the obstacles are changed, like in Episode 11, the Planner and TaskGen Agent combined are able to navigate and still clear the environment.

\subsubsection{Insights}
As we are trying to test TaskGen's ability to solve the maze, we intentionally use LLM as the Planner and LLM as the executor for the maze environment. However, such a logical pathfinding task is best done by rule-based deterministic methods like Breadth-First Search or A* search algorithm. It may be better to treat pathfinding with known obstacles as a problem solved by traditional pathfinding algorithms, and simply get TaskGen to call such a function to do pathfinding.

It is noted that the Planner was not able to perform well without few-shot prompting of something which might occur in the environment (i.e. a wall with a gap). This is a huge downside for using LLM as an optimiser, as it does not optimise well. LLMs also do not understand 2D text grids perfectly, and hence, the spatial awareness for the Planner is lacking, resulting in less robust plans.

We have also tried to use native TaskGen without the Planner, but the LLM was not able to see the big picture that well, resulting in LLM going to the same squares again and again trying to navigate past the wall. Planning is a difficult problem for an LLM and it is best to offload that to a rule-based Planner.

\newpage
\section{Escape Room Solving in TextWorld}
\renewcommand{\thefigure}{E\arabic{figure}}
\renewcommand{\thetable}{E\arabic{table}}
\setcounter{figure}{0}
\setcounter{table}{0}
\label{appendix: text-world}

\subsection{Introduction to Interactive Fiction / Escape Room Environments}
This appendix describes the implementation of an interactive fiction player as an agent. Interactive fiction is a genre of computer game that pre-dates GUIs, with many of the games (and tools) originating in the 1980s. The Microsoft TextWorld project has delivered a system for building arbitrary games and provides a framework for building agents to navigate these games.

The key game system in interactive fiction is the discovery of the game world. Players are presented with limited information at one time and are required to recall or rediscover elements of the game "world". Within advanced games, the game world may change without player interaction but this behaviour is not present in TextWorld.

Another aspect of interactive fiction is discovering how to interact with it. Players issue commands on each turn, typically in a terse pseudo-code, and the game attempts to interpret them. TextWorld has optional support for providing the player/agent the list of acceptable commands at each turn. Agents in this experiment will utilise these hints, if provided by the game. By design, the TaskGen agent developed does not depend on any specific input from the game and the few shot examples are not present in the test/development environment but chosen to represent "reasonably complex" commands where subject and object are each qualified.

Interactive fiction games may have counter-intuitive problems to solve to succeed in the game. e.g.: to cook a carrot, grill it directly on a stove. For this developer, carrots aren't often grilled, and things that are grilled are rarely done so directly on a stove.

\subsection{Conversation Class in TaskGen}

The TaskGen agent developed for this paper used the new Conversation Class interface, building on the existing escape room example.

\begin{figure}[H]
    \centering
    \includegraphics[width=\linewidth]{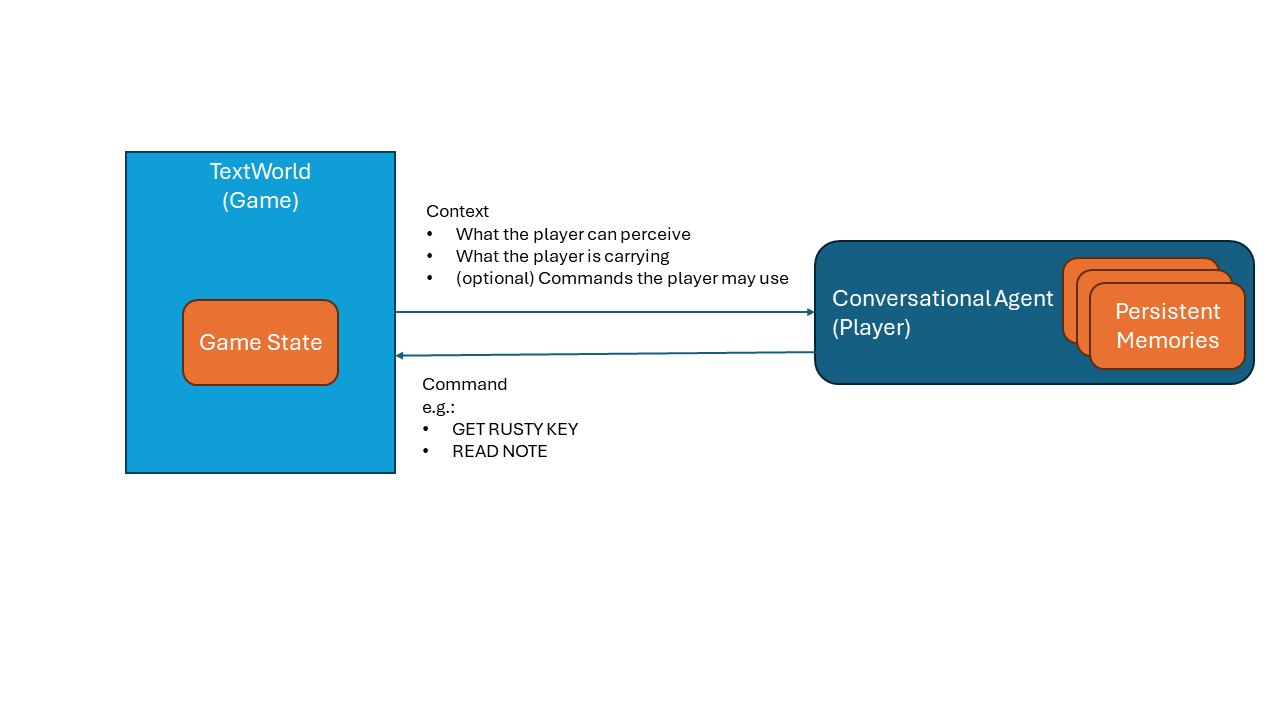}
    \caption{Block diagram of TextWorld and agent.}
    \label{fig:enter-label}
\end{figure}

The Persistent Memories of the player/agent are aligned to the core systems of interactive fiction: commands, rooms (locations in the game), objectives. The "Summary of Conversation" in the player/agent was useful for the agent as it allowed the agent to reflect on futile behaviour and move on to alternative solutions.

Structuring the Persistent Memory as "arrays" made the memories much more effective at guiding the agent.

\subsection{Agents Used}

\textbf{Random Agent.} The "random" agent is part of the TextWorld project and selects from the available commands at random. Without the commands provided, it cannot act.

\textbf{LLM Only Agent.} The LLM Only "gpt-4o" agent is simply a chat session, where each "observation" of context from the game to the player is another chat message. "gpt-4o" was remarkably effective in these circumstances.

\textbf{TaskGen Agent.} The input to the TaskGen Agent is the same as the LLM Only agent. The TaskGen Agent utilises the Conversation Class with Persistent Memories to store a continued awareness of the environment, which could potentially help it to make better decisions. 

\subsection{Experiment Setup}
The tests are from the TextWorld project examples. The variations relate to how detailed the "goals" (intermediate steps) are and how common the feedback from the game (points). As a proxy to goal solving, we can treat each point as one goal fulfilled, and so the total percentage of points earned in a game relative to the total points is a proxy for total solve rate of all goals.

We halt each game at 100 turns, and run each game 10 times for each agent and report averaged scores. All commands were truncated to prevent fatal buffer overflows in the 1980's game engine.

We vary three aspects of the environment. The first is goal description - detailed, brief, none. The second is rewards - dense or sparse. The third is whether commands are provided or not. We test the agents across the following environments, and report the score obtained: 
\begin{enumerate}
    \item Detailed Dense (commands provided)
    \item Brief Sparse (commands provided), 
    \item None Sparse (commands provided), 
    \item Detailed Dense (commands not provided), 
    \item Brief Sparse (commands not provided), 
    \item None Sparse (commands not provided)
\end{enumerate}


\newpage
\subsection{Results}
\begin{figure}[H]
    \centering
    \begin{tikzpicture}
        \begin{axis}[
            ybar,
            bar width=10pt,
            width=\textwidth,
            height=0.7\textwidth,
            xlabel={Agent},
            ylabel={Score (\% out of total possible points)},
            symbolic x coords={random, LLM Only, TaskGen},
            xtick=data,
            nodes near coords,
            ymin=0, ymax=110,
            legend style={at={(0.5,-0.25)}, anchor=north, legend columns=2},
            enlarge x limits={abs=1.5cm},
            ylabel near ticks,
            xticklabel style={rotate=45, anchor=east},
            title={Score for each Agent across 6 environments}
        ]
        
        \addplot coordinates {(random,42) (LLM Only,93) (TaskGen,96)};
        \addplot coordinates {(random,0) (LLM Only,20) (TaskGen,30)};
        \addplot coordinates {(random,0) (LLM Only, 20) (TaskGen,30)};
        \addplot coordinates {(random,0) (LLM Only,57) (TaskGen,88)};
        \addplot coordinates {(random,0) (LLM Only,0) (TaskGen,0)};
        \addplot coordinates {(random,0) (LLM Only,0) (TaskGen,0)};
        
        \legend{
            Detailed Dense (commands provided), 
            Brief Sparse (commands provided), 
            None Sparse (commands provided), 
            Detailed Dense (commands not provided), 
            Brief Sparse (commands not provided), 
            None Sparse (commands not provided)
        }
        
        \end{axis}
    \end{tikzpicture}
    \caption{Score for each Agent across 6 environments}
    \label{fig:textworld_score}
\end{figure}
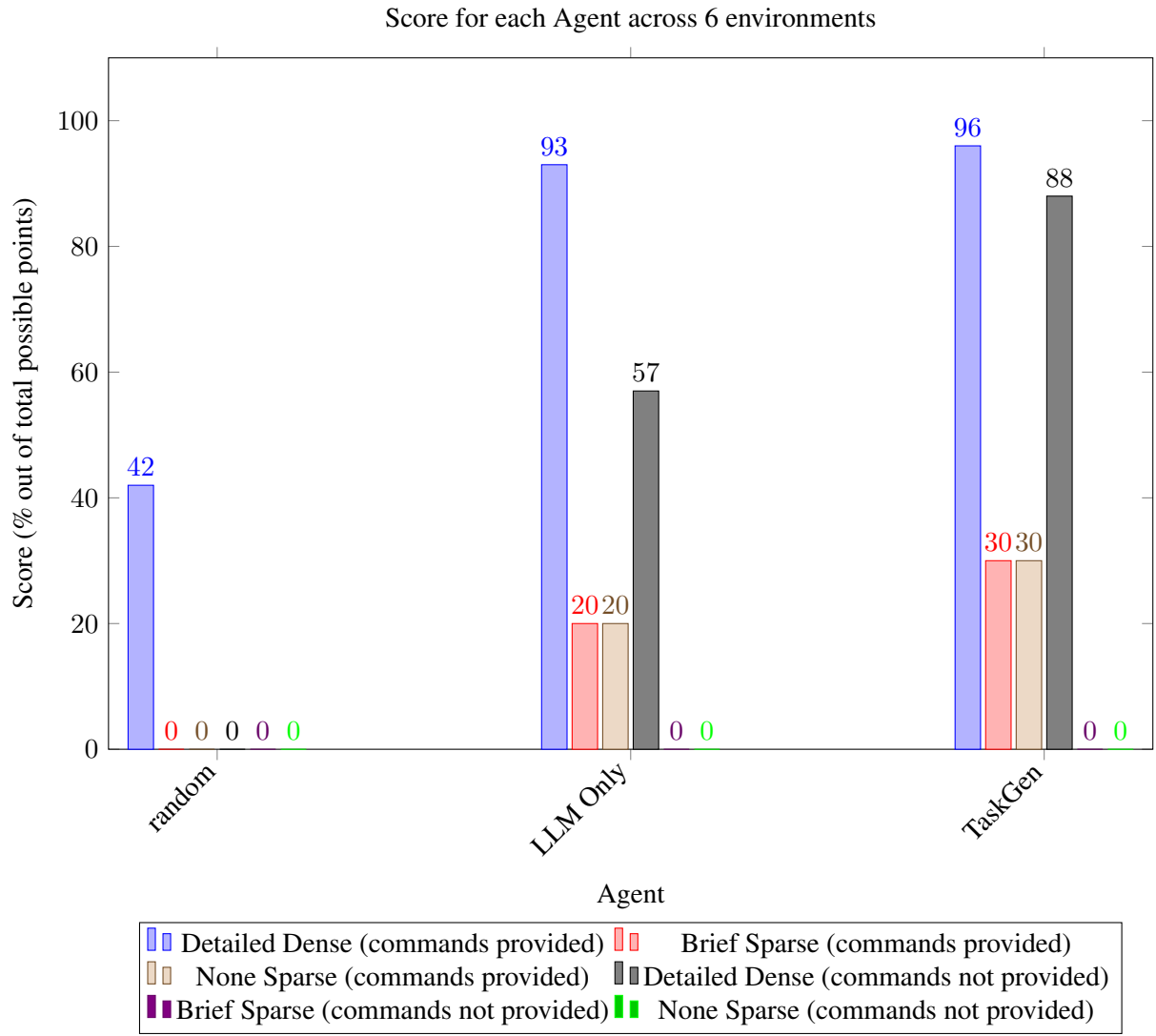

Fig. \ref{fig:textworld_score} shows the scores for the various agents on 6 different kinds of environments. In general, the TaskGen Agent performs the best, having a higher overall score (and hence solve rate) than LLM Only and random. It is to be noted that the None Sparse environments may be too difficult for the agents, as no goal is provided and the agents will need to search around until the right sequences of commands are done. This highlights that having goals is very important for solving an environment efficiently.

\subsection{Insights}
Reviewing the results, it seems likely that the LLM Only and TaskGen agents could (in some circumstances) have been more successful, if given more turns. 

It has been observed that when there are multiple objectives, the TaskGen Agent may not complete all of them dutifully and may think that it has completed some when it has not. This is a problem that could potentially be solved with a rule-based plan follower like that of the Planner in Appendix \ref{appendix: maze}.

\newpage
\section{Web-Browsing Agents}
\renewcommand{\thefigure}{F\arabic{figure}}
\renewcommand{\thetable}{F\arabic{table}}
\setcounter{figure}{0}
\setcounter{table}{0}
\label{appendix: web-browsing-agent}
This appendix details the implementation of Web-Browsing Agents in TaskGen.

The goal is to introduce the idea of Agents using a program/web application using TaskGen's current capability and performing actions based on user's query.

\subsection{Agent Diagram/Flow}

\begin{figure}[H]
\centering
\includegraphics[width=\linewidth]{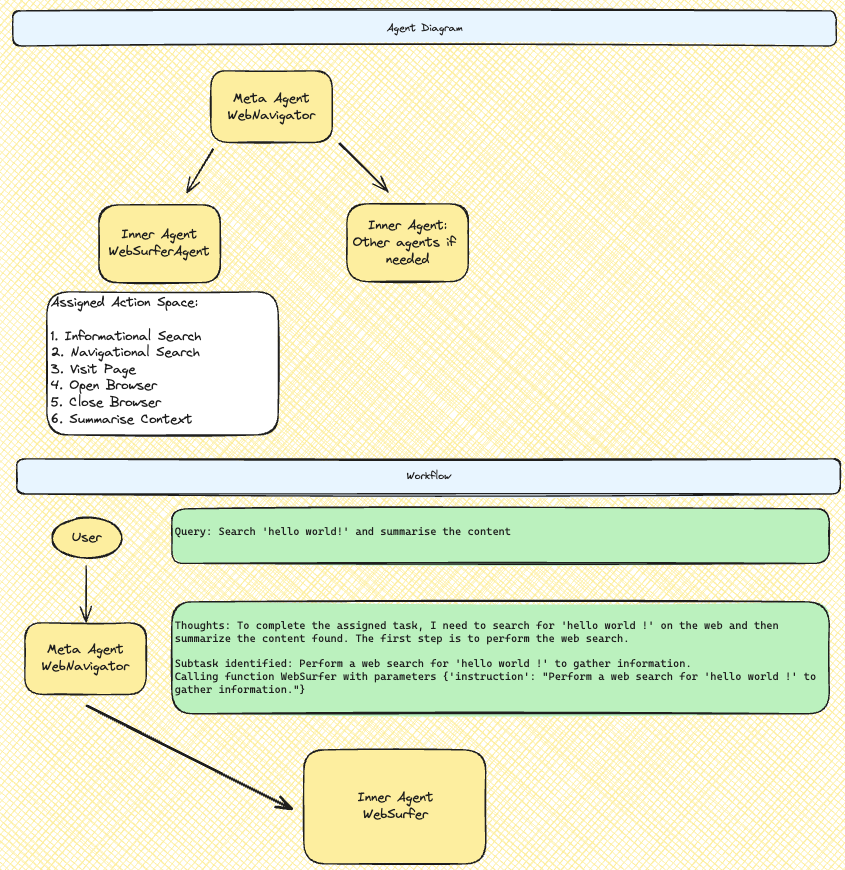}
\caption{Agent Diagram and beginning of agent flow. Consists of 1 Meta Agent and 1 Inner Agent. User performs a query via terminal CLI that starts off agent flow.}
\label{fig:WebBrowserAgentDiagram}
\end{figure}

\begin{figure}[H]
\centering
\includegraphics[width=\linewidth]{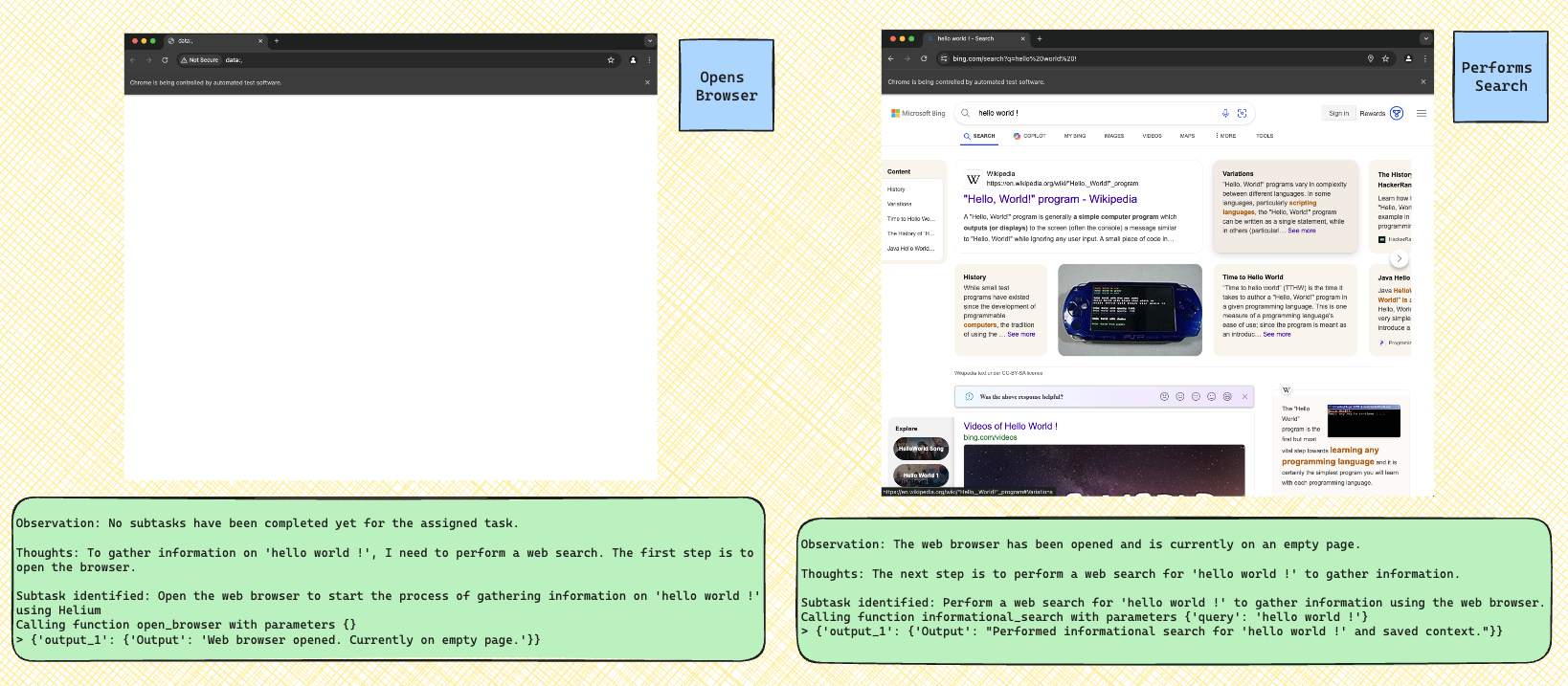}
\caption{Inner Agent opens browser and performs search}
\label{fig:WebBrowserAgentFlow1}
\end{figure}

\begin{figure}[H]
\centering
\includegraphics[width=\linewidth]{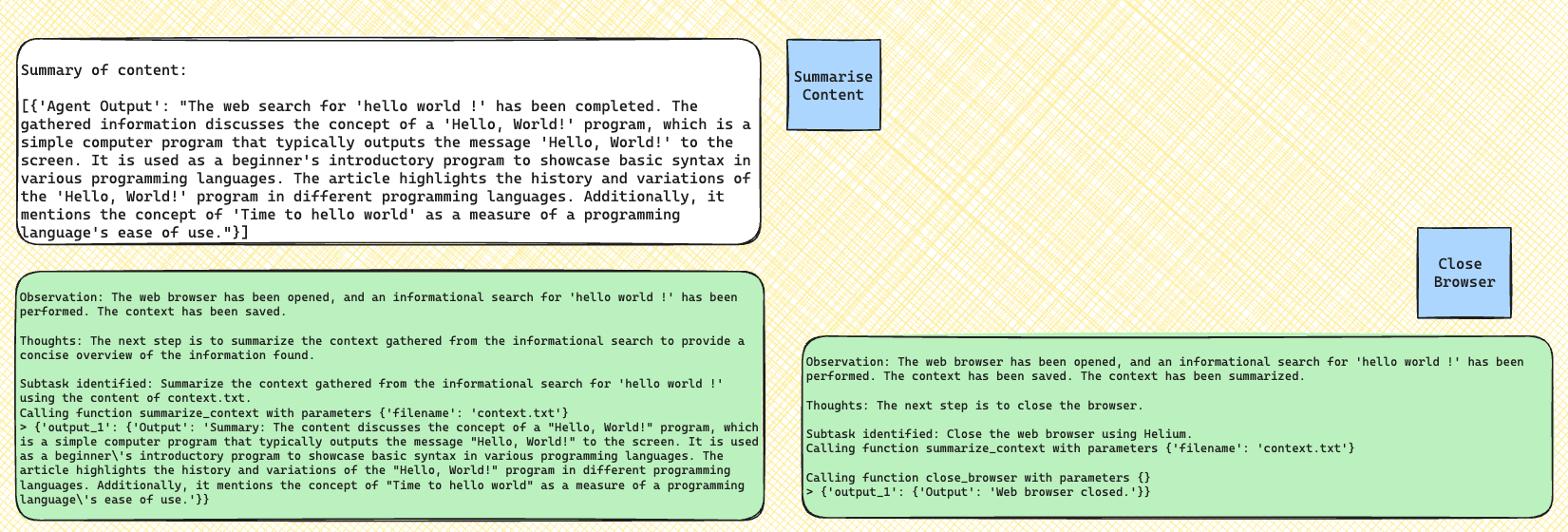}
\caption{Inner Agent opens browser and performs search}
\label{fig:WebBrowserAgentFlow2}
\end{figure}

\newpage
\subsection{Agent's Action Space}

\begin{itemize}
    \item \textbf{informational\_search}:
    \begin{itemize}
        \item \textbf{Description}: Performs a search query on Bing and saves the context of the search results page.
        \item \textbf{Steps}:
        \begin{enumerate}
            \item Navigates to Bing with the given search query.
            \item Captures the current state of the browser (URL, title, and page source).
            \item Extracts and saves relevant content from the page to a file.
        \end{enumerate}
    \end{itemize}

    \item \textbf{navigational\_search}:
    \begin{itemize}
        \item \textbf{Description}: Performs a search query on Bing, clicks the first result, and saves the context of the resulting page.
        \item \textbf{Steps}:
        \begin{enumerate}
            \item Navigates to Bing with the given search query.
            \item Clicks on the first search result link.
            \item Captures the current state of the browser (URL, title, and page source).
            \item Extracts and saves relevant content from the page to a file.
        \end{enumerate}
    \end{itemize}

    \item \textbf{visit\_page}:
    \begin{itemize}
        \item \textbf{Description}: Visits a specified URL and saves the context of the page.
        \item \textbf{Steps}:
        \begin{enumerate}
            \item Navigates to the given URL.
            \item Captures the current state of the browser (URL, title, and page source).
            \item Extracts and saves relevant content from the page to a file.
        \end{enumerate}
    \end{itemize}

    \item \textbf{open\_browser}:
    \begin{itemize}
        \item \textbf{Description}: Opens a web browser using Helium.
        \item \textbf{Steps}:
        \begin{enumerate}
            \item Starts a Chrome browser session.
            \item Returns a message indicating the browser has been opened.
        \end{enumerate}
    \end{itemize}

    \item \textbf{close\_browser}:
    \begin{itemize}
        \item \textbf{Description}: Closes the web browser using Helium.
        \item \textbf{Steps}:
        \begin{enumerate}
            \item Kills the current browser session.
            \item Returns a message indicating the browser has been closed.
        \end{enumerate}
    \end{itemize}

    \item \textbf{summarise\_context}:
    \begin{itemize}
        \item \textbf{Description}: summarises the content saved in a file (default: \texttt{context.txt}) using OpenAI's GPT model.
        \item \textbf{Steps}:
        \begin{enumerate}
            \item Reads the content from the specified file.
            \item Sends the content to OpenAI's API to generate a summary.
            \item Returns the generated summary.
        \end{enumerate}
    \end{itemize}
\end{itemize}

\newpage
\subsection{TaskGen Code for Web Browsing}

We use a Meta Agent with Inner Agents to solve the task of web browsing. Some of the code used as as shown below:

\subsubsection{Function Definition}
\begin{lstlisting}[style=Python]
def informational_search(query: str) -> str:
    go_to(f"https://www.bing.com/search?q={query}")
    header, content = _browser_state()
    save_context_to_file(header, content)
    return {
        "Output": f"Performed informational search for '{query}' and saved context."
    }
\end{lstlisting}

\subsubsection{Meta Agent Creation}
\begin{lstlisting}[style=Python]
WebSurfer = Agent(
    "WebSurfer",
    "Performs web searches and navigates web pages. Always open the browser at the start of the task and close the browser at the end.",
    model="gpt-4o",
    default_to_llm=False,
).assign_functions(fn_list_3)
\end{lstlisting}

\subsubsection{Boss Agent}
\begin{lstlisting}[style=Python]
bossagent = Agent(
    "WebNavigator",
    "Assists user to navigate the web. Always open the browser at the start of the task and close the browser at the end.",
    model="gpt-4o",
    default_to_llm=False,
)
\end{lstlisting}






\newpage
\subsection{Results of Web Browsing Agent}

\begin{figure}[H]
\centering
\includegraphics[width=\textwidth]{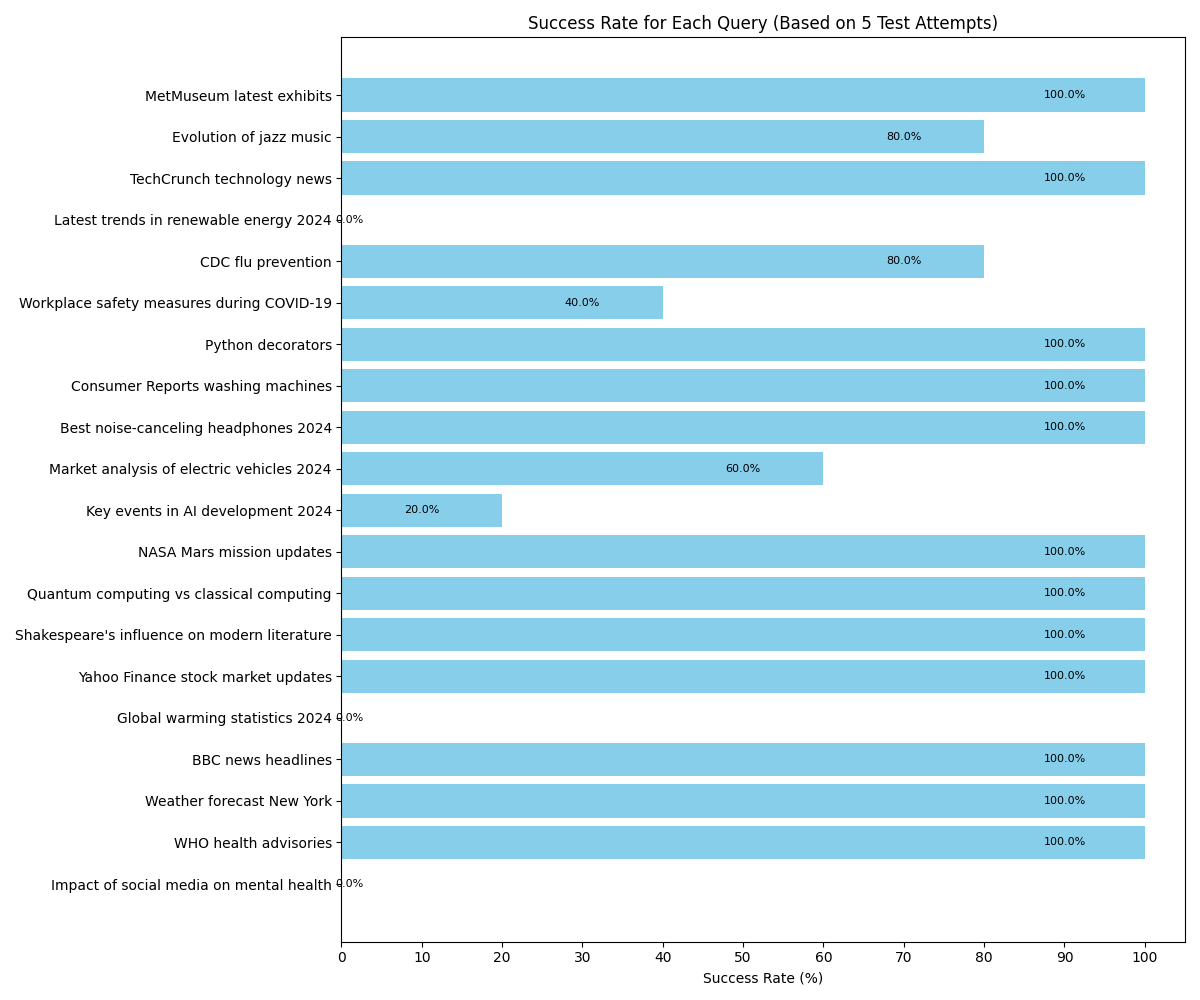}
\caption{Graphical representation of the success rates for each query tested. Each query was tested 5 times, and the success rate is calculated as the percentage of successful attempts out of these 5 tests. The chart compares the effectiveness of different queries, providing a clear visualization of the success rate for each query.}
\label{fig:WebBrowserAgentResult}
\end{figure}

Fig. \ref{fig:WebBrowserAgentResult} shows the result of the Agent executing various queries. We achieve an overall solve rate of \textbf{69\%}. While better prompt engineering may significantly improve the solve rate, our focus is to show a working initial interface, so we did not over-engineer for this use case.

\newpage
\section{MATH Dataset}
\renewcommand{\thefigure}{G\arabic{figure}}
\renewcommand{\thetable}{G\arabic{table}}
\setcounter{figure}{0}
\setcounter{table}{0}
\label{appendix: MATH}

\begin{figure}[H]
    \centering
\includegraphics[width=\linewidth]{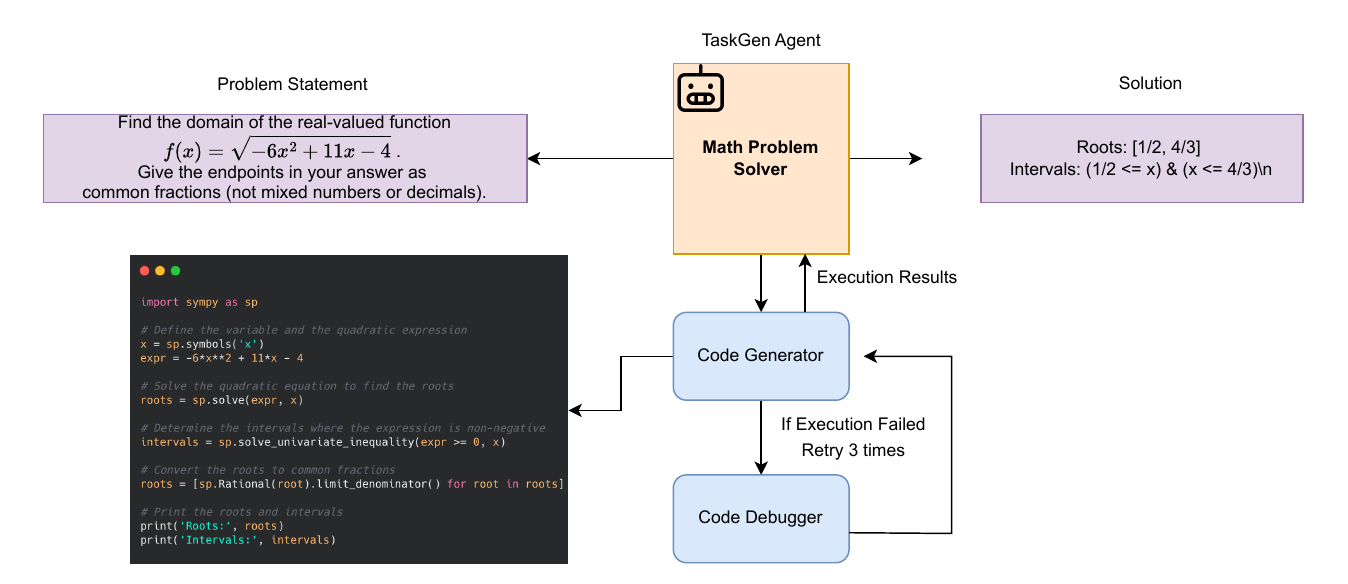}
    \caption{Math Problem Solver Agent}
    \label{fig:math-agent}
\end{figure}

Leveraging LLMs for solving mathematical problems has become a widely researched area \cite{Zhou2023SolvingCM,Yang2023GPTCS}. In this section, we explore the use of the TaskGen agent to tackle complex mathematical problems across various domains. For our evaluations, we utilised the MATH dataset \cite{Hendrycks2021MeasuringMP}, which contains over 12,500 competition-level mathematical problems. We focused specifically on the most challenging subset, \textbf{Level 5}, across the following categories: Algebra, Pre-Algebra, Intermediate Algebra, Number Theory, and Counting and Probability. We randomly selected 20 problems from the test set of each category, resulting in a total of 100 problems in our test set, to assess the TaskGen agent's ability to solve these tasks.

To solve these challenging problems, we employed a TaskGen Agent called ``Math Problem Solver'' powered with GPT4o as depicted in Figure \ref{fig:math-agent}. This agent is equipped with a specialised functions that facilitates the generation, execution, and debugging of code necessary to tackle the given tasks. The function has access to essential Python libraries, including \texttt{numpy}, \texttt{sympy}, \texttt{math}, and \texttt{random}.

\begin{figure}[H]
    \centering
    \includegraphics[width=\linewidth]{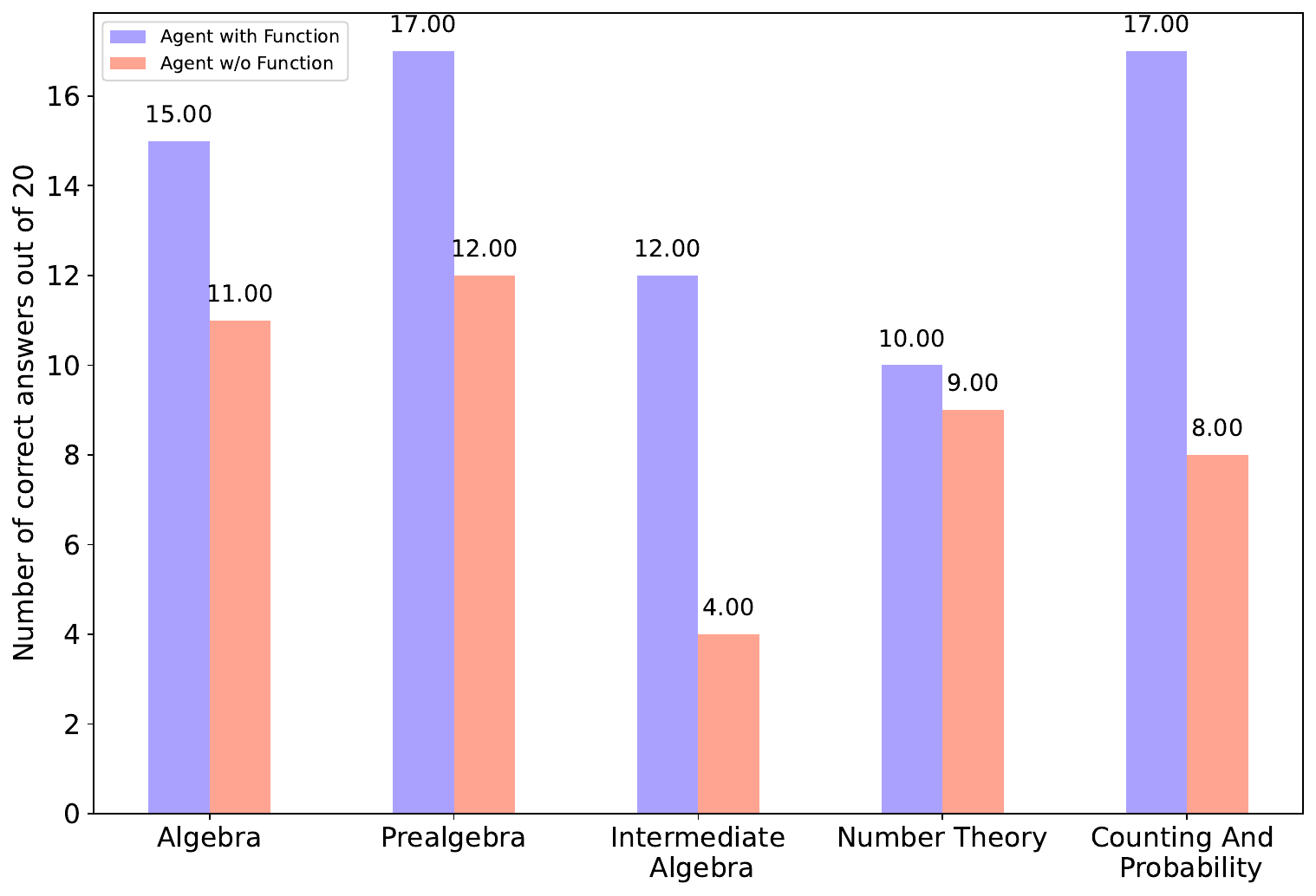}
    \caption{Quantitative results of TaskGen agents on the subset of the MATH dataset.}
    \label{fig:math-eval}
\end{figure}

\newpage
\paragraph{Evaluation Result.} In Figure \ref{fig:math-eval}, we provide the evaluation results of our TaskGen agent equipped with the function described above and the agent without any equipped functions.

From our experiments, we found that on the challenging Level-5 problems, the TaskGen agent with Equipped Functions achieved an average accuracy of \textbf{71\%}, while the TaskGen agent without the Equipped Functions achieved only \textbf{44\%} accuracy. For evaluation, we manually verified the generated solution against the provided ground truth solution. These results demonstrate that, in order to solve these challenging Level-5 problems, equipping the agent with code generation and debugging capabilities leads to more accurate solutions of mathematical problems.

\newpage
\section{RAG-based Question Answering on NaturalQuestions Dataset}
\label{appendix:naturalquestions_rag}

\renewcommand{\thefigure}{H\arabic{figure}}
\renewcommand{\thetable}{H\arabic{table}}
\setcounter{figure}{0}
\setcounter{table}{0}

In this section, we describe the development and functionality of a Retrieval-Augmented Generation (RAG) system using TaskGen. This system integrates one TaskGen agent, known as the ``User Agent,'' along with two critical TaskGen functions: \textbf{ContextFetchFunction} and \textbf{AnswerFunction}. These components constitute the fundamental operations of our system.

\subsection{System Overview}

\begin{enumerate}
    \item \textbf{ContextFetchFunction}: This function accepts a user's query and a batch number, retrieving the relevant context. It is designed to incrementally fetch more context if the initial retrieval proves inadequate.
    \item \textbf{AnswerFunction}: After receiving context, this function generates an answer based on the context available. If the context is insufficient to resolve the query, \textbf{AnswerFunction} returns ``no answer.''
    \item \textbf{User Agent}: The orchestrator of the entire Q\&A cycle, the User Agent is responsible for managing both the \textbf{ContextFetchFunction} and the \textbf{AnswerFunction}. It initiates the process by retrieving context for the user's query and continues to fetch additional context in subsequent batches until a satisfactory answer is found or the interaction limit is reached.
\end{enumerate}

\subsection{Detailed Process}

\begin{itemize}
    \item \textbf{Query Submission}: The user submits a query to the User Agent.
    \item \textbf{Context Retrieval}: The User Agent invokes \textbf{ContextFetchFunction} with the initial query and a starting batch number (0).
    \item \textbf{Answer Generation}: With the context obtained, the User Agent next activates \textbf{AnswerFunction}. If the context sufficiently answers the query, a response is generated. Otherwise, it issues ``no answer.''
    \item \textbf{Incremental Fetching}: If ``no answer'' is received, the User Agent increments the batch number and re-engages \textbf{ContextFetchFunction} to obtain more context. This iterative process is capped at five interactions (max interactive retrieval count).
\end{itemize}

\begin{figure}[H]
\centering
\includegraphics[width=0.8\textwidth]{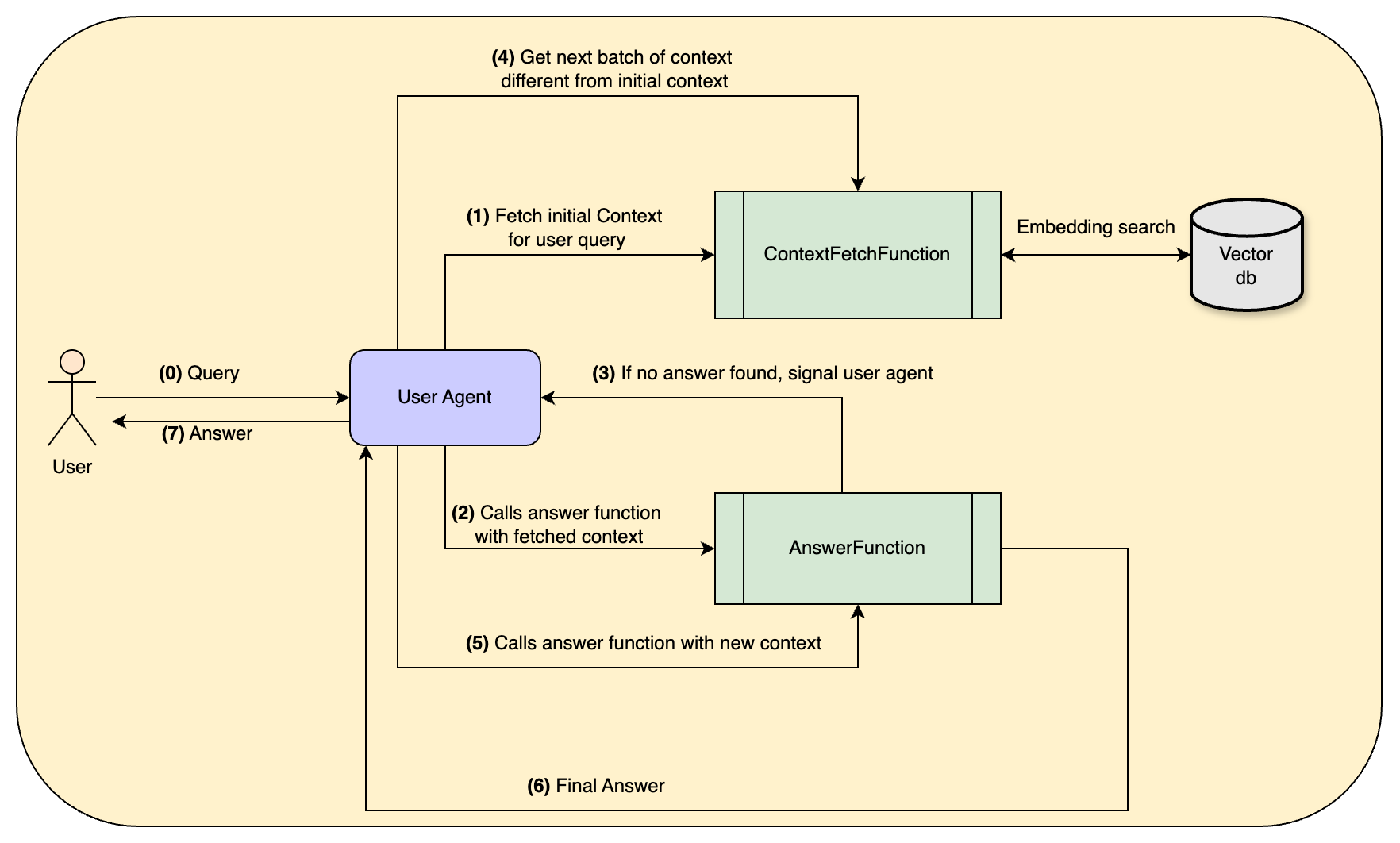}
\caption{Illustration of the Interactive Retrieval-Augmented Generation (RAG) Question and Answer Flow. The diagram sequentially represents the process from step (0) to (7), detailing the interaction between the User, User Agent, ContextFetchFunction, and AnswerFunction. Each numbered marker (num) in the diagram corresponds to a specific step in the query-answer cycle. }
\end{figure}

\subsection{Example}

Consider the query: ``What is the capital of France?''

\begin{itemize}
    \item \textbf{First Batch (Batch 0)}:
    \begin{verbatim}
    1. Paris is a major city in France.
    2. France is known for its culture and cuisine.
    \end{verbatim}
    \textit{AnswerFunction Output}: ``no answer'' (the context does not explicitly state Paris as the capital).

    \item \textbf{Second Batch (Batch 1)}:
    \begin{verbatim}
    1. The capital of France is Paris.
    2. Paris is famous for the Eiffel Tower.
    \end{verbatim}
    \textit{AnswerFunction Output}: ``The capital of France is Paris.''
\end{itemize}

This example illustrates the system's ability to handle complex queries by sequentially enhancing the context until a definitive answer can be provided.

\subsection{Technical Framework and Evaluation Methodology}

This section outlines the technical and methodological specifics employed in our study to develop and evaluate the Retrieval-Augmented Generation (RAG) system.

\subsubsection{Technology Stack}
\begin{itemize}
    \item \textbf{TaskGen Agents and Functions}: Our system utilises TaskGen's capabilities extensively. The core components, namely the \textbf{User Agent}, \textbf{ContextFetchFunction}, and \textbf{AnswerFunction}, are powered by GPT-3.5. This model was chosen for its lower cost and robust performance in natural language understanding and generation.
    \item \textbf{Embedding Storage and Retrieval}: We employed \textbf{Postgres PGvector} to manage the storage and retrieval of embeddings. For Retrieval we use \( k = 10 \) configuring our system to fetch top 10 most relevant vector embeddings for each query
    \item \textbf{Embedding Model}: The \textbf{text-embedding-ada-002} model was used to convert text data into vector embeddings. These embeddings represent the textual data in a format amenable to similarity comparisons and retrieval operations.
\end{itemize}

\subsubsection{Dataset and Benchmarking}
\begin{itemize}
    \item \textbf{Dataset}: The \href{https://github.com/google-research-datasets/natural-questions}{Natural Questions dataset} was chosen for its comprehensive collection of real-world questions. Our study focuses on the first 2,000 entries of the development split validation set, providing a balanced mix of complexity and coverage.
    \item \textbf{Evaluation Metrics}: To assess the effectiveness of our RAG system, we used Google's \href{https://github.com/google-research-datasets/natural-questions/blob/master/nq_eval.py}{nq\_eval script}. This script is widely recognised for its rigour in measuring the precision and accuracy of answers provided by question-answering systems.
\end{itemize}

\newpage
\subsubsection{Evaluation Results}
We conducted a comprehensive evaluation to compare the performance of non-interactive versus interactive (via TaskGen) retrieval method. The non-interactive retrieval approach involves a single invocation of an LLM using context from the vector database to answer the query. This method assumes that the initial context contains all the necessary information to generate an answer. In contrast, the interactive retrieval method dynamically fetches and refines context based on the ongoing interaction with the user's query, allowing for a more adaptive and potentially accurate response as additional information is incorporated in successive retrieval steps.

\begin{figure}[H]
\centering
\includegraphics[width=0.9\textwidth]{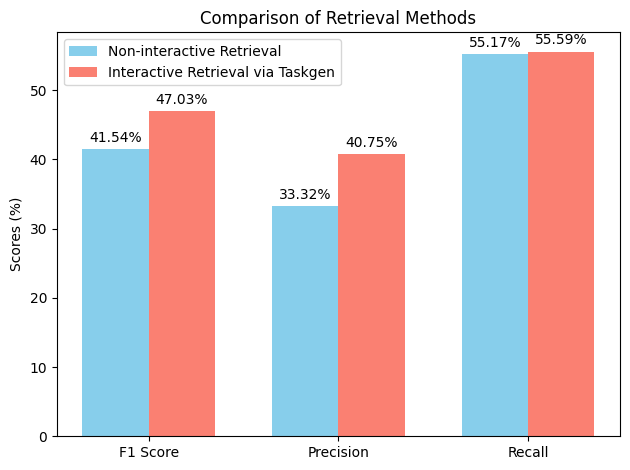}
\caption{Graphical representation of the benchmark results comparing F1 Score, Precision, and Recall for Non-Interactive versus Interactive Retrieval via TaskGen (for both k=10 used for retrieval).}
\label{fig:benchmark_comparison}
\end{figure}

\end{document}